\documentclass[lettersize,journal]{IEEEtran}

\usepackage[utf8]{inputenc} % allow utf-8 input
\usepackage[T1]{fontenc}    % use 8-bit T1 fonts
\usepackage{adjustbox}
\usepackage{graphicx}  
\usepackage{url}            % simple URL typesetting
 \usepackage{graphicx}

\usepackage{soul}
\usepackage{caption}
\usepackage[font=tiny]{subfig}
\usepackage{verbatim}
 \usepackage{multirow}
\usepackage[linesnumbered,ruled,boxed]{algorithm2e}
\usepackage{amssymb}
\usepackage{mathtools} % amsmath with fixes and additions
\usepackage{tikz} % nice language for creating drawings and diagrams
\usepackage{booktabs}       % professional-quality tables
\usepackage{amsfonts}       % blackboard math symbols
\usepackage{nicefrac}       % compact symbols for 1/2, etc.
\usepackage{microtype}      % microtypography
\usepackage{xcolor}         % colors
\usepackage{amsthm}
\theoremstyle{theorem} 
\newtheorem{theorem}{Theorem}
\newtheorem{corollary}{Corollary}
\newtheorem{lemma}{Lemma}
\theoremstyle{definition}   
\newtheorem{definition}{Definition}
\newtheorem{assumption}{Assumption}
\theoremstyle{remark}  
\newtheorem{remark}{Remark}

\def\BibTeX{{\rm B\kern-.05em{\sc i\kern-.025em b}\kern-.08em
    T\kern-.1667em\lower.7ex\hbox{E}\kern-.125emX}}
\usepackage{balance}
\begin{document}
\title{ Advocating for the Silent: Enhancing Federated
Generalization for Non-Participating Clients}
\author{Zheshun Wu, Zenglin Xu, \IEEEmembership{Senior Member, IEEE}, Dun Zeng, Qifan Wang, and Jie Liu, \IEEEmembership{Fellow, IEEE}
    \thanks{This work was supported in part by National Natural Science Foundation of China (No. 62350710797), and the Major Key Project of PCL (No. PCL2023A09). (\emph{Corresponding author: Jie Liu, Zenglin Xu.})}
\thanks{Zheshun Wu and Jie Liu are with the School of Computer Science and Technology, Harbin Institute of Technology Shenzhen, Shenzhen 518055, China (e-mail:
wuzhsh23@gmail.com; jieliu@hit.edu.cn).}
\thanks{Zenglin Xu is with the Fudan University, and also with the Shanghai Academy of Artificial Intelligence for Science  (e-mail: zenglin@gmail.com).}
\thanks{Dun Zeng is with the Department of Computer Science and Engineering, University of Electronic Science and Technology of China, Chengdu
611731, China (e-mail: zengdun@std.uestc.edu.cn).}
\thanks{Qifan Wang is with the FaceBook AI, Menlo Park, CA 94025 USA. (e-mail: wqfcr@fb.com).}}

\markboth{Journal of \LaTeX\ Class Files,~Vol.~18, No.~9, September~2020}%
{How to Use the IEEEtran \LaTeX \ Templates}

\maketitle

\begin{abstract}
Federated Learning (FL) has surged in prominence due to its capability of collaborative model training without direct data sharing. However, the vast disparity in local data distributions among clients, often termed the Non-Independent Identically Distributed (Non-IID) challenge, poses a significant hurdle to FL's generalization efficacy. The scenario becomes even more complex when not all clients participate in the training process, a common occurrence due to unstable network connections or limited computational capacities. This can greatly complicate the assessment of the trained models' generalization abilities. While a plethora of recent studies has centered on the generalization gap pertaining to unseen data from participating clients with diverse distributions, the distinction between the training distributions of participating clients and the testing distributions of non-participating ones has been largely overlooked.
In response, our paper unveils an information-theoretic generalization framework for FL. Specifically, it quantifies generalization errors by evaluating the information entropy of local distributions and discerning discrepancies across these distributions. Inspired by our deduced generalization bounds, we introduce a weighted aggregation approach and a duo of client selection strategies. These innovations are designed to strengthen FL’s ability to generalize and thus ensure that trained models perform better on non-participating clients by incorporating a more diverse range of client data distributions. Our extensive empirical evaluations reaffirm the potency of our proposed methods, aligning seamlessly with our theoretical construct.

\end{abstract}

\section{Introduction}\label{sec:intro}
\IEEEPARstart{F}{ederated} Learning (FL) offers a collaborative paradigm to train a shared global model across distributed clients, ensuring data privacy by eliminating the need for direct data transfers~\cite{zhu2021federated,DBLP:conf/aistats/McMahanMRHA17,10106044}. Therefore, FL provides a secure architecture to bolster confidentiality and effectively manage diverse sensitive data~\cite{DBLP:journals/comcom/CoelhoNVSN23,10.1145/3678181}, ranging from financial to healthcare information. Consequently, FL applications span various domains, encompassing finance~\cite{DBLP:conf/icdm/MaoWHYHY23}, healthcare~\cite{DBLP:journals/comcom/CoelhoNVSN23}, recommendation systems~\cite{10423793}, Internet-of-Things (IoT)~\cite{DBLP:journals/tnsm/WuWL23}, and more. However, the inherent heterogeneity among clients—often due to their distinct operational environments—generates Non-Independent and Identically Distributed (Non-IID) scenario~\cite{10304290,hu2023generalization,zhao2018federated}. This distinctness complicates the assessment of FL's generalization capabilities, setting it apart from traditional centralized learning~\cite{yuan2021we,DBLP:conf/icml/MohriSS19,huang2024federated}.

\begin{figure}[ht]
    \centering
    \includegraphics[width=3.4in]{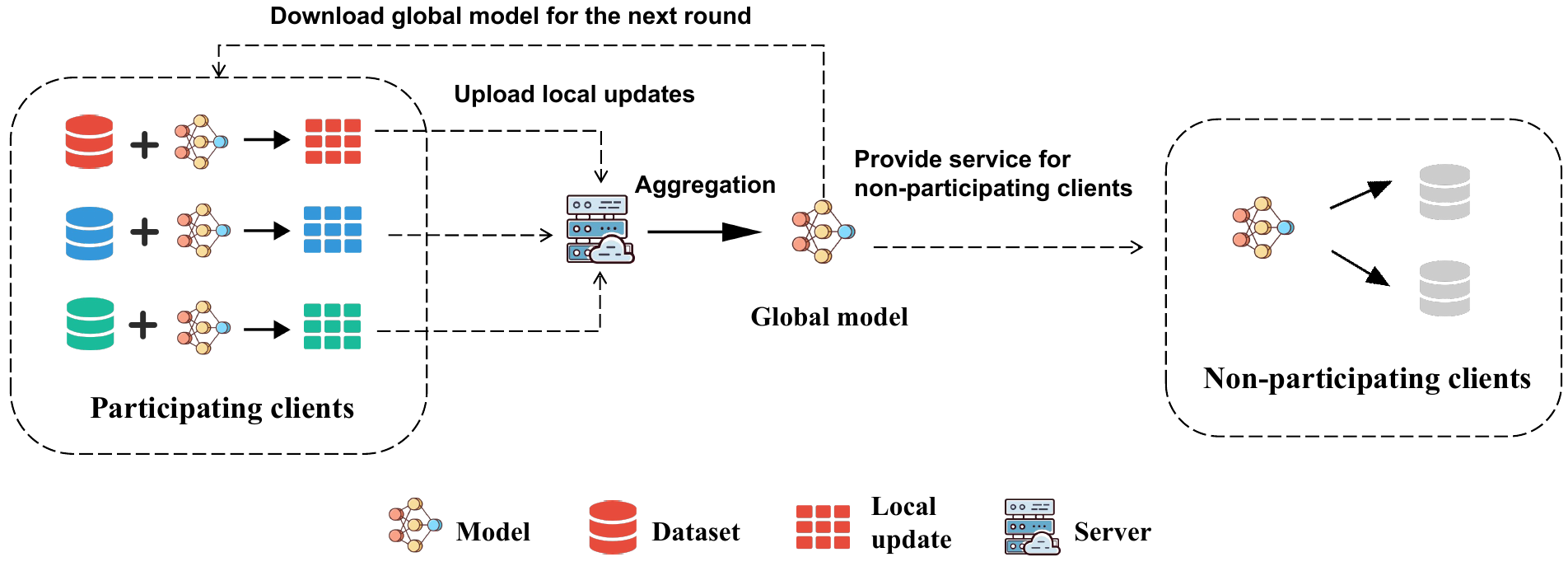}
    \caption{An illustration of the considered FL system with 3 participating clients and 2 non-participating clients. The participating clients perform local training and upload local updates to the server. The server aggregates these results and updates the global model. Then the global model is used for providing service for non-participating clients. }
    \label{fig:system}
\end{figure}

While prevailing research on FL generalization predominantly concentrates on actively participating clients~\cite{DBLP:conf/icml/MohriSS19,DBLP:conf/icml/QuLDLTL22,wei2022non}. it offers a limited view: it addresses the model's adaptability to observed local distributions without reconciling the divergence between observed data distributions in actively participating clients and unobserved data distributions from passive non-participating clients. 
In real-world settings, numerous clients might remain detached from the training process, due to unstable network connectivity or other constraints~\cite{hu2023generalization,lim2020federated}. For example, in cross-device federated edge learning, IoT devices, although reliant on robust deep learning models for edge inferences~\cite{DBLP:journals/corr/abs-2302-09019,DBLP:conf/icml/0005SLW0X22}, often abstain from FL owing to computational and communicative constraints~\cite{tak2020federated}. Figure~\ref{fig:system} depicts a FL system, where participating clients gather data to train a global model designed to provide services for non-participating clients. This landscape raises a pivotal question: \textbf{Can models, honed by active participants, cater effectively to passive clients that do not participate in the training process?}

At the heart of this question is the assurance of optimal performance for the model trained by FL on non-participating clients. For properly evaluating the generalization performance of models, some recent endeavors strive to demystify this by quantifying the performance disparity between models tested on active versus passive clients, proposing a dual-level framework to evaluate both out-of-sample and out-of-distribution generalization gaps, grounded in the premise that client data distributions stem from a meta-distribution~\cite{hu2023generalization,yuan2021we}. Yet, the approach by \cite{yuan2021we} primarily sketches the contours of the FL generalization gap, substantiating it with empirical evidence sans a comprehensive theoretical underpinning. Conversely, the approach by \cite{hu2023generalization} accentuates theoretical insights on meta-distribution-based errors but leaves practical algorithmic solutions largely unexplored, particularly those that might uplift the global model's adaptability for passive clients.

\textbf{Motivation.} Compared with traditional FL deploying a global model for actively participating clients and assuming an alignment between training and testing data distributions, our investigation spotlights the global model's out-of-distribution generalization capabilities when trained by participating clients. In essence, we seek to cultivate a model via FL across active clients that reliably serves even the passive ones. Specifically, a novel contribution of our paper is the introduction of the self-information weighted expected risk -- a metric to gauge model generalization. Our hypothesis is grounded in the conception that a model exhibiting proficiency with low-probability examples from training distributions might demonstrate adaptability to unfamiliar testing distributions. Such examples are anticipated to hold a greater significance in these unseen datasets. Operationalizing this concept within FL, our framework leverages the self-information of examples to craft a generalization boundary. Delving into the resulting information entropy-aware and distribution discrepancy-aware generalization disparities, a revelation emerges: data sources tend to manifest informational redundancy. This implies that certain data sources, characterized by diminished informational weight, can be seamlessly supplanted by others. Essentially, only a fraction of the client base holds substantial sway over FL's generalization. This redundancy is especially pronounced in the IoT scenario, where a myriad of edge devices—often operating in overlapping zones—partake in FL~\cite{10472111}. Consider drones, serving as FL clients, and amassing spatial data for model training. Drones constrained by limited flight spans might be rendered redundant due to their shared operational territories with other drones~\cite{wang2021federated}. Informed by these insights, our paper further proposes strategies to enhance the generalization capabilities of FL, enabling trained models to better serve non-participating clients.

\textbf{Contributions.} Our paper presents several key contributions.
\begin{itemize}
    \item We introduce a novel theoretical framework to scrutinize the generalization error in FL. Distinctly, our approach shines a light on the distribution discrepancy, an aspect largely glossed over in preceding research. This framework adeptly harnesses the information entropy of data sources, coupled with the distribution variances among them, to provide a more refined insight into the generalization potential of models.
    \item Drawing from our theoretical results, we devise a weighted aggregation approach alongside a duo of client selection methods. These are designed to amplify the generalization prowess of FL, so trained models can provide better service for non-participating clients.
    \item The empirical evaluations using three widely-referenced datasets underscore the efficacy of our methods, consistently eclipsing benchmarks, which is matched with our theoretical findings.
\end{itemize}

\section{Related Work}
Data heterogeneity is a major challenge in federated learning \cite{10295990,DBLP:conf/nips/ReisizadehFPJ20,ma2022state}.  Despite numerous studies investigating the generalization error in the presence of data heterogeneity \cite{DBLP:conf/icml/MohriSS19,DBLP:conf/icml/QuLDLTL22,caldarola2022improving}, most of these have  focused only on scenarios where a global model is trained on distributed data sources and tested on unseen data sampled from these sources. While~\cite{hu2023generalization,yuan2021we,wei2022non}   also consider this generalization problem in FL, they do not account for local distribution characteristics nor propose methods to enhance the out-of-distribution generalization performance of models. In essence, they cannot ensure that models trained via  FL will perform effectively for non-participating clients. In contrast, our work is motivated by the need to design algorithms to train a global model providing good generalization performance on passive clients (that do not participate in the training process) with unknown data distributions. Our framework takes into account the local distribution property  and provides methods for improving generalization performance in this setting. 

Additionally, there have been several studies that propose information-theoretic generalization analyses for FL \cite{DBLP:conf/spawc/YagliDP20, DBLP:conf/isit/BarnesDP22, sefidgaran2022ratedistortion,10374183}. For instance, \cite{DBLP:conf/spawc/YagliDP20} developed a framework for generalization bounds that also accounts for privacy leakage in FL. \cite{DBLP:conf/isit/BarnesDP22} presented generalization bounds for FL problems with Bregman divergence or Lipschitz continuous losses.~\cite{10374183} derives the algorithm-dependent generalization bound for federated edge learning. However, these works focus on bounding the generalization error via the mutual information between models and training samples, ignoring the information stored in data sources. And they only consider the in-distribution generalization error, with the assumption that the source distribution is identical to the target distribution. 

Besides, compared with Federated Domain Generalization (FedDG)~\cite{huang2024federated, DBLP:conf/cvpr/ZhangXYZ0W23, DBLP:conf/nips/NguyenTL22} dealing with the challenge of domain distribution shift under FL settings, our focus is solely on a specific collection of all the data sources in FL. Our objective is to train a global model from seen distributions of participating clients  and generalize it to other unseen distributions of non-participating clients. Additionally, it is commonly assumed that each domain contains equal information and cannot represent each other in FedDG. In other words, FedDG does not consider the inherent distribution correlation or information redundancy across data sources. However, we consider the presence of information redundancy among data sources and only a subset of clients contributes to the generalization of FL. Our focus is thus  to identify representative data sources. In terms of developed algorithms,   most DG studies concentrate on learning an invariant representation across different domains \cite{nguyen2022fedsr, DBLP:conf/aaai/LiSWZRLK022,muandet2013domain}. Conversely, our focus is to design proper weighting aggregation and client selection methods to mitigate the generalization error in FL.

\section{Theoretical Framework}\label{sec:the}
Similar to previous studies~\cite{hu2023generalization,yuan2021we}, we model each data source as a random variable with its corresponding distribution. In order to better evaluate the generalization performance of FL to ensure trained models perform well for non-participating clients, we focus on the information-theoretic generalization gap in FL. Furthermore, similar to~\cite{hu2023generalization, DBLP:conf/icml/MohriSS19}, we also utilize uniform convergence-based generalization analysis in this study. This technique, as discussed by~\cite{DBLP:conf/nips/ParkSE22}, is algorithm-independent, making it applicable to a wide range of federated optimization algorithms~\cite{DBLP:conf/aistats/McMahanMRHA17,li2020federated,10269141}. To conserve space, we have provided proofs for all our theorems in the appendix.
\subsection{Preliminaries}
%The key difference between the proposed theoretical framework and \cite{hu2023generalization} is that we focus on the self-information weighted expected risk based on the joint distribution of different data sources, in order to analyze the out-of-distribution generalization of FL better. 
Let the sample space  $\mathcal{Z}=\mathcal{X} \times \mathcal{Y}$ be the set of all the possible outcomes $z=(x,y)$ (e.g. image-label pairs) focused on in this paper, where $\mathcal{X}$ is the feature space and $\mathcal{Y}$ is the label space. Let $\mathcal{I}$ be the index set of all the possible clients. The total number of all possible clients in $\mathcal{I}$ is $\vert \mathcal{I} \vert=N$ and it is possibly infinite. We assume that only clients in the finite subset $\mathcal{I}_p$ of $\mathcal{I}$ participating in FL practically and the number of these clients is $\vert \mathcal{I}_p \vert=M$. Following \cite{hu2023generalization}, $N$ is commonly much larger than $M$ due to the unreliable network links. Additionally, we can only select a subset $\mathcal{I}_t$ of $\mathcal{I}_p$ for data collection and local model training in each round of FL. This approach offers the advantage of reducing both the computational load and communication burden on participating clients. Similarly, we denote the number of selected clients $|\mathcal{I}_t|$ as $K$. Therefore, we have $\mathcal{I}_t \subset \mathcal{I}_p \subset \mathcal{I}$ and $K\leq M\leq N $. 

In this paper, we assume that each participating client $i, \forall i \in \mathcal{I}_p $ is associated with a local data source $Z_i$, where $Z_i=(X_i,Y_i)$ is a discrete random variable with the probability mass function $P_{Z_i}$ and is with support on $\mathcal{Z}$. The sequence of participating data sources $Z_1,Z_2,...,Z_M$ is denoted as $Z^{\mathcal{I}_p}$ and the corresponding joint distribution is denoted as $P_{Z^{\mathcal{I}_p}}$ in the following. The local distribution $P_{Z_i}$ of each data source $Z_i, \forall i \in \mathcal{I}_p$ is different from each other, i.e., $ P_{Z_i} \neq P_{Z_j}, \forall i \neq j, i,j \in \mathcal{I}$, which is common in FL~\cite{zhu2021federated,hu2023generalization,zhao2018federated}. 
%The non-IID setting considered in this paper is as follows. The local distribution $P_{Z_i}$ of each data source $Z_i, \forall i \in \mathcal{I}$ is different from each other, i.e., $ P_{Z_i} \neq P_{Z_j}, \forall i \neq j, i,j \in \mathcal{I}$, which is common in FL scenario \cite{zhu2021federated}. And we assume that $Z^{\mathcal{I}_p}$ is independent of $Z^{\mathcal{I} \setminus \mathcal{I}_p}$. 

The local training set $ S_i=\{ s_i^j\}_{j=1}^{n_i}, \forall i \in \mathcal{I}_p$ stored on participating client $i$ is made of $n_i$ i.i.d. realizations from the local data source $Z_i \sim P_{Z_i}$. Referring to \cite{yuan2021we}, the objective of federated generalization is to train a global model on $\{ S_i\}_{i \in \mathcal{I}_p}$, such that all the possible clients in $\mathcal{I}$ will be provided satisfactory service by this global model trained by participating clients. Let $\mathcal{H}$ be a hypothesis class on $\mathcal{X}$. The loss function $\ell: \mathcal{Y} \times \mathcal{Y} \rightarrow \mathbb{R}^+$ is a non-negative function and we assume that $\ell$ is bounded by $b$ and Lipschitz continuous. For simplicity, we denote $\ell(h(x),y)$ as $\ell(h,z)$ in the following. 

To understand the proposed framework better, we first present some definitions as follows:
\begin{definition}
[Self-information weighted expected risk] The self-information of outcome $z_i \in \mathcal{Z}$ is denoted by $\log(\frac{1}{P(z_i)})$. Then, the self-information weighted expected risk is defined by
\begin{equation}\label{eq1}
    \begin{aligned}
\mathcal{L}_{Z_i}(h)&:=\mathbb{E}\Big[ \ell(h,Z_i)\log(\frac{1}{P(Z_i)}) \Big]\\
&=\sum_{z_i \in \mathcal{Z}}P_{Z_i}(z_i)\ell(h,z_i)\log(\frac{1}{P_{Z_i}(z_i)}),
    \end{aligned}
\end{equation}
where $ h $ is a specific model in  $\mathcal{H}$ and $\ell(h,z_i)$ is the loss function of model $h$ on sample $z_i$.
\end{definition}
The rationale behind the formulation of the risk outlined in~\eqref{eq1} can be attributed to the fact that target distributions are unknown in the training stage under the OOD setting. The proposed loss $\ell(h,z_i)\log(\frac{1}{P_{Z_i}(z_i)})$, is indicative of the requisite focus on outcomes $z_i \in \mathcal{Z}$  with lower probabilities in source distributions since they may have higher probabilities in unknown target distributions. Additionally, it is reasonable to assume that unknown data sources have maximum entropy distributions, signifying greater uncertainty, exemplified by a uniform distribution of discrete labels.  For example,  IoT devices may only collect data in specific areas, while the global model trained by these devices is expected to provide spatial-related services for all devices across the entire area~\cite{10472111}.  Hence, the self-information of each outcome is indispensable for measuring the expected risk in such a situation. Moreover, we know that applying FL in the healthcare field~\cite{10197230} should focus on silos within rare disease cases, which underscores the significance of rare yet informative samples, thereby supporting our viewpoint.
%To sum up, the self-information of each sample point should be taken into account when facing with the OOD challenge.

\begin{definition}
[Joint self-information weighted expected risk] The self-information of one particular combination of outcomes $z_1,z_2,..., z_{M}$ in the product space is denoted by $\log(\frac{1}{P_{Z^{\mathcal{I}_p}}(z_1,...,z_{{M}})})$. We use the term $\sum_{i \in \mathcal{I}_p}\alpha_i\ell(h,z_i), \sum_{i\in\mathcal{I}_p}\alpha_i=1, \alpha_i\geq 0,\forall i \in \mathcal{I}_p$ as the loss below. This term denotes the average loss of model $h$ predicated on $z_1,z_2,...,z_{M}$. Then, the joint self-information weighted expected risk on multiple data sources is defined by
\begin{equation}\label{eq2}
    \begin{aligned}
\mathcal{L}_{Z^{\mathcal{I}_p}}(h)&:=\mathbb{E}_{Z^{\mathcal{I}_p}}\Big[\sum_{i \in \mathcal{I}_p}\alpha_i\ell(h,z_i)\log(\frac{1}{P_{Z^{\mathcal{I}_p}}})\Big]\\
&=\sum_{z_1\in \mathcal{Z}}\sum_{z_2\in \mathcal{Z}}...\sum_{z_{M}\in \mathcal{Z}}P_{Z^{\mathcal{I}_p}}(z_1,z_2,...,z_{M}) \\
&\quad \sum_{i \in \mathcal{I}_p}\alpha_i\ell(h,z_i)\log(\frac{1}{P_{Z^{\mathcal{I}}}(z_1,z_2,...,z_{{M}})}),\\
    \end{aligned}
\end{equation}
where $ h $ is a specific model in hypothesis space $\mathcal{H}$. Similarly, the joint self-information $\log(\frac{1}{P_{Z^{\mathcal{I}}}(z_1,...,z_{M})})$ reflects the uncertainty of the event that outcomes $z_1, z_2,..., z_{M}$ are sampled from $Z_1, Z_2,..., Z_{M}$ 
 respectively. We also denote $\mathcal{L}_{Z^{\mathcal{I}_p}}(h)$ as $\sum_{\mathcal{Z}^{ \mathcal{I}_p}}P_{Z^{\mathcal{I}_p}}\sum_{i \in \mathcal{I}_p}\alpha_i\ell(h,z_i)\log(\frac{1}{P_{Z^{\mathcal{I}_p}}})$ in the following.
\end{definition}

Analogously, the motivation for defining the risk in Eq.~\eqref{eq2} is that distributions of non-participating clients may differ from distributions of participating clients significantly. We must account for the self-information of every possible combination of outcomes in order to ensure that one model performing well on participating clients will also do well on non-participating clients with distinct distributions. 

Moving forward, we introduce the general training objective in FL, as well as the self-information weighted semi-empirical risk which we propose in this paper, respectively. Let $S_i=\{ s_i^j\}_{j=1}^{n_i}$ be the local training set of $i$-th participating client based on i.i.d. realizations from examples of the data source $Z_i $, where $n_i$ denotes the size of $S_i$. $S$ is the whole training set over all the participating clients defined as $S=\cup_{i \in \mathcal{I}_p}S_i$. The empirical risk minimization (ERM) objective in federated learning~\cite{hu2023generalization} is formulated as follows:
\begin{equation}
     \min_{h\in\mathcal{H}} \big\{\mathcal{L}_{S}(h):=\sum_{i \in \mathcal{I}_p} \frac{ \alpha_i}{n_i}\sum_{j=1}^{n_i} \ell(h,s_{i}^{j})\big\}, \alpha_i \geq 0, \sum_{i \in \mathcal{I}_p}\alpha_i=1,
\end{equation}
where $s_i^j$ denotes the $j$-th training sample at $i$-th selected client and $\alpha_i$ is the weighting factor of client $i$. The empirical risk minimizer $\hat{h}$ is defined by $\hat{h}=\arg\inf_{h \in \mathcal{H}}\mathcal{L}_{S}(h)$. 
\begin{comment}
    \begin{definition}
    [Empirical risk in federated learning  \cite{hu2023generalization}]
  The empirical risk minimization (ERM) objective in federated learning is formulated as follows
    \begin{equation}\label{eq7}
     \mathcal{L}_{S}(h):=\sum_{i \in \mathcal{I}_p} \frac{ \alpha_i}{n_i}\sum_{j=1}^{n_i} \ell(h,s_{i}^{j}), \alpha_i \geq 0, \sum_{i \in \mathcal{I}_p}\alpha_i=1,
\end{equation}
where $s_i^j$ denotes the $j$-th training sample at $i$-th selected client and $\alpha_i$ is the weighting factor of client $i$.
\end{definition}
\end{comment}

 Motivated by the definition of semi-empirical risk $\frac{1}{|\mathcal{I}_p|}\sum_{i\in \mathcal{I}_p}\mathbb{E}_{Z \sim P_{Z_i}} [\ell(h,Z)]$ in \cite{hu2023generalization}, we further propose the self-information weighted semi-empirical risk as follows. The self-information weighted semi-empirical risk $\mathcal{L}_{Z_{\mathcal{I}_p}}(h)$ rooted on data source $\{Z_i\}_{i \in \mathcal{I}_p}$ is defined by,
 
\begin{equation}
    \mathcal{L}_{Z_{\mathcal{I}_p}}(h):= \sum_{i \in \mathcal{I}_p} \alpha_i\mathcal{L}_{Z_i}(h)=\sum_{i \in \mathcal{I}_p} \alpha_i\mathbb{E}[ \ell(h,Z_i)\log(\frac{1}{P(Z_i)}) ].
\end{equation}
The proposed self-information weighted semi-empirical risk measures the average information-theoretic performance of $h$ on the participating data source $\{Z_i\}_{i \in \mathcal{I}_p}$.

%The semi-empirical risk minimizer $\hat{h}^*$ is denoted as $\hat{h}^*=\arg \inf_{h \in \mathcal{H}}\mathcal{L}_{Z_{\mathcal{I}_p}}(h)$. For participating clients $\mathcal{I}_p$, the corresponding semi-excess risk is defined as $\mathcal{L}_{Z_{\mathcal{I}_p}}(\hat{h})-\mathcal{L}_{Z_{\mathcal{I}_p}}(\hat{h}^*)$ \cite{hu2023generalization}. The semi-excess risk represents how well the trained model $\hat{h}$ in practice can perform on the unseen data from $\{ Z_i\}_{i \in \mathcal{I}_p}$ from the view of information theory. 

%\section{federated generalization}
\subsection{Federated Generalization}

We now formally present the introduction of our proposed information-theoretic generalization framework in FL. We first define the information-theoretic  generalization gap in FL as follows,
\begin{definition}[Information-theoretic  generalization gap in federated learning]
    \begin{equation}\label{eq9}
\begin{aligned}
&Gen(\mathcal{I}_p,\hat{h})\\
&:=|\mathcal{L}_{Z^{\mathcal{I}_p}}(\hat{h})-\mathcal{L}_{S}(\hat{h})|\\
&=\Big\vert\mathbb{E}_{Z^{\mathcal{I}_p}}\big[\sum_{i \in \mathcal{I}_p}\alpha_i\ell(\hat{h},z_i)\log\frac{1}{P_{Z^{\mathcal{I}_p}}}\big]-\sum_{i \in \mathcal{I}_p} \frac{ \alpha_i}{n_i}\sum_{j=1}^{n_i} \ell(\hat{h},s_{i}^{j}) \Big\vert,
\end{aligned}
\end{equation}
where $ \alpha_i \geq 0, \sum_{i \in \mathcal{I}_p}\alpha_i=1$.
\end{definition}

The motivation for defining the gap in~\eqref{eq9} is that we want to know the performance gap between the empirical risk evaluated by  $\hat{h}$  and the joint self-information weighted expected risk evaluated by   $\hat{h}$. Based on our aforementioned analysis, this generalization gap can reflect how well the trained model $\hat{h}$ will perform on unknown data sources $Z^{\mathcal{I}}$.

%We will find that the upper bound of this generalization gap $|\mathcal{L}_{Z^{\mathcal{I}}}(\hat{h})-\mathcal{L}_{Z_{\mathcal{I}_p}}(\hat{h})|$ includes the semi-excess risk $|\mathcal{L}_{Z_{\mathcal{I}_p}}(\hat{h})-\mathcal{L}_{Z_{\mathcal{I}_p}}(\hat{h}^*)|$ defined above. 

%\subsection{Generalization gap in federated learning}

Furthermore, we decompose the original  generalization gap in~\eqref{eq9}  as follows,
\begin{equation}\label{eq10}
    \begin{aligned}
&\Big\vert\mathcal{L}_{Z^{\mathcal{I}_p}}(\hat{h})- \mathcal{L}_{S}(\hat{h})\Big\vert  
 \\
&\leq\underbrace{\Big\vert\mathcal{L}_{Z^{\mathcal{I}_p}}(\hat{h})-\mathcal{L}_{{Z_{\mathcal{I}_p}}}(\hat{h})\Big\vert}_{\mbox{distributed learning gap}}+\underbrace{\Big\vert\mathcal{L}_{{Z_{\mathcal{I}_p}}}(\hat{h})-\mathcal{L}_{S}(\hat{h})\Big\vert}_{\mbox{semi-generalization gap}},
    \end{aligned}
\end{equation}

We conduct our theoretical analysis using the following assumption.

\begin{assumption}[Limited Independence]
\label{assumption2}
 Among the participating data sources, each single data source $Z_i,\forall i \in \mathcal{I}_p$ is independent of the sequence of other participating data sources $Z^{ \mathcal{I}_p \setminus i} $, i.e., $P_{Z^{\mathcal{I}_p}}=P_{Z_i}P_{Z^{\mathcal{I}_p\setminus i } }, \forall i \in \mathcal{I}_p$. In addition, data sources $Z^{\mathcal{I}_t}$ constructed by selected clients are independent of data sources $Z^{\mathcal{I}_p \setminus \mathcal{I}_t}$, i.e., $P_{Z^{\mathcal{I}_p}}=P_{Z^{\mathcal{I}_t}}P_{Z^{\mathcal{I}_p \setminus \mathcal{I}_t}}$. 
\end{assumption}

 Rooted on the above assumption, we can derive two lemmas introduced in the appendix and further derive the below theorem about the information-theoretic  generalization gap in FL.
%In the following, we analyze terms in this decomposed generalization gap step by step and regard them as three distinct lemmas.

\begin{theorem}
     [Information entropy-aware  generalization gap in FL]
     \label{the:FedEntropy} Let $\mathcal{G}$ be a family of functions related to hypothesis space $\mathcal{H}: z\mapsto \ell(h,z):h \in \mathcal{H}$ with VC dimension $VC(\mathcal{G})$. For any $\delta \geq 0$, if $\ell$ is bounded by $b$, it follows that with probability at least $1-\delta$,
\begin{equation}\label{eq15}
    \begin{aligned}
&|\mathcal{L}_{Z^{\mathcal{I}_p}}(\hat{h})- \mathcal{L}_{S}(\hat{h})  |\leq bH(Z^{\mathcal{I}_p})-b\sum_{i\in\mathcal{I}_p}\alpha_i H(Z_i)\\
&\quad+\mathcal{E}+cb\sqrt{\frac{VC(\mathcal{G})}{\sum_{i=1}^{\vert\mathcal{I}_p \vert}n_i}} +b\sqrt{\frac{\log(1/\delta)}{2\sum_{i=1}^{\vert\mathcal{I}_p \vert}n_i}},
    \end{aligned}
\end{equation}
where $c$ is a constant.   $\mathcal{E} =\sum_{z \in \mathcal{Z}}\ell(h',z)$, where $h'=\sup_{h \in \mathcal{H}}  \vert\mathcal{L}_{{Z_{\mathcal{I}_p}}}(h)-\sum_{i\in \mathcal{I}_p}\alpha_i \mathbb{E}[\ell(h,z_i)] \vert$. 
\end{theorem}
\begin{remark}
Theorem~\ref{the:FedEntropy} asserts that increasing the weighted sum $\sum_{i\in \mathcal{I}_p}\alpha_i  H(Z_i)$ will reduce this generalization bound. This suggests that models trained by FL will exhibit enhanced performance on unknown data sources when a greater weighting factor is assigned to the data source with richer information. In addition,  model complexity $VC(\mathcal{G})$ and sample complexity $\sum_{i=1}^{\vert\mathcal{I}_p \vert}n_i$ also affect the generalization capacity of FL.  Notice that alternative metrics, like Rademacher complexity and covering number, can be employed to refine VC-dimension-based generalization bounds~\cite{hu2023generalization,9927349}, but delving into these is beyond the scope of this paper. Our emphasis lies in devising algorithms by leveraging insights from the proposed information-theoretic generalization framework.
\end{remark}

We then consider assigning the identical weighting factor $\alpha_i=\frac{1}{M}$  for each client.To explore the impact on each individual client resulting from the distributed learning performed on decentralized data sources, we examine the average information-theoretic distributed learning gap defined as $\frac{1}{M}\big\vert\mathcal{L}_{Z^{\mathcal{I}_p}}(\hat{h})-\mathcal{L}_{{Z_{\mathcal{I}_p}}}(\hat{h})\big\vert$ and show that its upper bound is related to the entropy rate of the stochastic process $\{Z_i\}_{i \in \mathcal{I}_p}$.
\begin{corollary}\label{entropyrate}
    Assume the entropy rate $H(\mathcal{Z})=\lim_{M \rightarrow \infty}\frac{1}{M}H(Z^{\mathcal{I}_p})$ of stochastic process $\{Z_i\}_{i \in \mathcal{I}_p}$  exists. Let the weighting factor $\alpha_i$ be $\frac{1}{M}$ for each client, we have,
        \begin{equation}
        \begin{aligned}
         & \lim_{M \rightarrow \infty}\frac{1}{M} \big\vert\mathcal{L}_{Z^{\mathcal{I}_p}}(\hat{h})-\mathcal{L}_{{Z_{\mathcal{I}_p}}}(\hat{h})\big\vert\leq bH(\mathcal{Z}),
        \end{aligned}
    \end{equation}
     \begin{remark}
        Corollary~\ref{entropyrate} indicates that if the entropy rate $H(\mathcal{Z})$ of  $\{Z_i\}_{i \in \mathcal{I}_p}$ exists,  the average information rate or uncertainty associated with the considered stochastic process $\{Z_i\}_{i \in \mathcal{I}_p}$ influences the generalization of FL.  
    \end{remark}
    \end{corollary}

In the following, we consider only leveraging a selected subset $\mathcal{I}_t \subset \mathcal{I}_p$ with the identical weighting factor $\alpha_i=\alpha_j=\frac{1}{|\mathcal{I}_t|}=\frac{1}{K},\forall i\neq j, i,j\in \mathcal{I}_t$ for each selected client to measure the generalization gap in FL. In other words, we turn to focus on the below generalization gap,
\begin{equation}
\begin{aligned}
&Gen(\mathcal{I}_t,\hat{h})\\
&:=\vert\mathcal{L}_{Z^{\mathcal{I}_p}}(\hat{h}_t)- \mathcal{L}_{S_t} (\hat{h}_t)\vert \\&=\Big\vert\mathbb{E}_{Z^{\mathcal{I}_p}}\frac{\sum_{i\in\mathcal{I}_p}\ell(\hat{h}_t,Z_i) \log\frac{1}{P(Z^{\mathcal{I}_p})}}{M}-\sum_{i \in \mathcal{I}_t} \frac{ 1}{n_i}\sum_{j=1}^{n_i} \frac{\ell(\hat{h}_t,s_{i}^{j})}{K}\Big\vert,
\end{aligned}
\end{equation}
where $\hat{h}_t=\arg \inf_{h \in \mathcal{H}}\mathcal{L}_{S_t}(h)$ and $S_t=\cup_{i \in \mathcal{I}_t}S_i$.

Similarly, we can decompose this gap as follows,
\begin{equation}\label{eq101}
    \begin{aligned}
&\Big\vert\mathcal{L}_{Z^{\mathcal{I}_p}}(\hat{h}_t)- \mathcal{L}_{S_t}(\hat{h}_t)\Big\vert  
 \\
&\leq\underbrace{\Big\vert\mathcal{L}_{Z^{\mathcal{I}_p}}(\hat{h}_t)-\mathcal{L}_{{Z_{\mathcal{I}_t}}}(\hat{h}_t)\Big\vert}_{\mbox{selection gap}}+\underbrace{\Big\vert\mathcal{L}_{{Z_{\mathcal{I}_t}}}(\hat{h}_t)-\mathcal{L}_{S_t}(\hat{h}_t)\Big\vert}_{\mbox{semi-generalization gap}},
    \end{aligned}
\end{equation}

Referring to the derivation presented in the proof of Theorem~\ref{the:FedEntropy} in the appendix, we first derive the upper bound of the proposed information-theoretic selection gap below.

\begin{theorem} [Correlation-aware selection gap in FL] \label{the:Fedcorre}  If $\ell$ is bounded by $b$, we have
 \begin{equation}\label{eq151}
    \begin{aligned}
 \vert\mathcal{L}_{Z^{\mathcal{I}_p}}(\hat{h}_t)-\mathcal{L}_{{Z_{\mathcal{I}_t}}}(\hat{h}_t)\vert & \leq b(3-\frac{2K}{M}-\frac{1}{K})H(Z^{\mathcal{I}_p} )\\
 &\quad+\frac{b}{K}I(Z^{\mathcal{I}_p};Z^{\mathcal{I}_p\setminus\mathcal{I}_t}) ,\end{aligned}
\end{equation}
where $I(Z^{\mathcal{I}_p};Z^{\mathcal{I}_p\setminus\mathcal{I}_t}):=\mathbb{E}_{Z^{\mathcal{I}_p}}[\log\frac{P_{Z^{\mathcal{I}_p}}}{P_{Z^{\mathcal{I}_p}}P_{Z^{\mathcal{I}_p\setminus\mathcal{I}_t}}}]  $ is the mutual information between  $Z^{\mathcal{I}_p}$ and $Z^{\mathcal{I}_p\setminus\mathcal{I}_t}$, measuring the correlation between these two data sources.
\begin{remark}
    Theorem~\ref{the:Fedcorre} demonstrates that a lower mutual information $I(Z^{\mathcal{I}_p};Z^{\mathcal{I}_p\setminus\mathcal{I}_t})$ results in a smaller selection gap. This implies that the performance of models trained by selected clients will be enhanced if the unselected data sources $Z^{\mathcal{I}_p\setminus\mathcal{I}_t}$ contain less information from participating data sources $Z^{\mathcal{I}_p}$. In other words, Theorem~\ref{the:Fedcorre} reveals that participating data sources contain redundant information and a subset $\mathcal{I}_t$ of $\mathcal{I}_p$ is adequate to represent the entirety.
\end{remark}
\end{theorem}

Based on this result, we can further establish another theorem that pertains to the information-theoretical generalization gap in FL under the considered client selection scenario.

\begin{theorem} [Distribution discrepancy-aware generalization gap in FL] \label{the:FedCross}  Let $\mathcal{G}$ be a family of functions related to hypothesis space $\mathcal{H}: z\mapsto \ell(h,z):h \in \mathcal{H}$ with VC dimension $VC(\mathcal{G})$. For any $\delta \geq 0$, if $\ell$ is bounded by $b$, it follows that with probability at least $1-\delta$,
 \begin{equation}\label{eq16}
    \begin{aligned}
& \vert\mathcal{L}_{Z^{\mathcal{I}_p}}(\hat{h}_t)- \mathcal{L}_{S_t} (\hat{h}_t)\vert  \\
& \leq b(3-\frac{2K}{M}-\frac{1}{K})H(Z^{\mathcal{I}_p} )+\frac{b}{K}\sum_{i \in \mathcal{I}_p\setminus\mathcal{I}_t}H(P_{Z_i},P_{Z_j})+\mathcal{E}_t\\
&\quad+cb\sqrt{\frac{VC(\mathcal{G})}{\sum_{i=1}^{\vert\mathcal{I}_t \vert}n_i}}+b\sqrt{\frac{\log(1/\delta)}{2\sum_{i=1}^{\vert\mathcal{I}_t \vert}n_i}}, \forall j\neq i,j \in \mathcal{I}
_p,   \end{aligned}
\end{equation}
where  $c$ is a constant. $\mathcal{E}_t =\sum_{z \in \mathcal{Z}}\ell(h_t',z)$, where $h_t':=\sup_{h \in \mathcal{H}}  \vert\mathcal{L}_{{Z_{\mathcal{I}_t}}}(h)-\sum_{i\in \mathcal{I}_t}\alpha_i \mathbb{E}[\ell(h,z_i)] \vert$.  
$H(P_{Z_i},P_{Z_j})=\mathbb{E}_{P_{Z_i}}[\log\frac{1}{P_{Z_j}}]$ is the cross entropy between  $P_{Z_i}$ and $P_{Z_j}$, measuring the dissimilarity between these two distributions.
\end{theorem}

\begin{remark}
Theorem~\ref{the:FedCross} indicates that lower dissimilarity $H(P_{Z_i},P_{Z_j})$ between unselected distributions $P_{Z_i}, i \in \mathcal{I}_p \setminus \mathcal{I}_t $ and  other participating distributions $P_{Z_j}, j \in \mathcal{I}_p$  can  reduce the generalization gap.  Notice that we do not need to compute the cross entropy $H(P_{Z_i},P_{Z_j})$ in practice, the derived bound only inspires us to design the client selection algorithms to enhance federated generalization.
\end{remark}

\section{Methods}
  This section focuses on introducing a weighting aggregation approach, along with two client selection methods in FL, that are rooted on the theoretical findings mentioned earlier. The objective of employing these methods is to enhance the generalization performance of FL. 

\subsection{Maximum Entropy Aggregation}
Inspired by Theorem \ref{the:FedEntropy}, it is easy to find that to minimize the information-theoretic generalization gap is actually to maximize the term as follows
\begin{equation}\label{eq19}
     \max_{ \{\alpha_i\}_{ i \in \mathcal{I}_p}}\sum_{i\in \mathcal{I}_p}\alpha_i  H(Z_i),
\end{equation}
where $\sum_{i \in \mathcal{I}_p} \alpha_i=1,\alpha_i \geq 0$, $H(Z_i)$ is the information entropy of data source $Z_i$.

The true entropy of data source in~\eqref{eq19} is  inaccessible in practice, so we can not assign $\alpha_i=1$ for $i=\arg\max_{\Tilde{i} \in \mathcal{I}_p} H(Z_{\Tilde{i}})$ directly. Therefore, we can design proper weighting factors of local gradients in federated aggregation to maximize this term.

\textbf{Empirical entropy-based weighting:}
Based on the above analysis, the weighting factor of local gradients or local models should be increased proportionally to the information entropy $H(Z_i)$ of the data source $Z_i$. This paper considers only the label distribution skew scenario for verifying the proposed empirical entropy-based weighting method. Therefore, we can design the aggregation weighting factor as follows:

\begin{equation}\label{eq20}
    \alpha_i =\frac{\exp{(\hat{H}_i})}{\sum_{i \in \mathcal{I}_p}\exp{(\hat{H}_i)}}.
\end{equation}

In this paper, the empirical entropy can be thus calculated via  $\hat{H}_i=-\sum_{y \in \mathcal{Y}} \frac{\sum_{j}\mathbb{I}_{y=y_j^i} }{n_i}\log\frac{\sum_{j}\mathbb{I}_{y=y_j^i}}{n_i}$, where $y_j^i$ is the label of $j$-th sample of local dataset $S_i$ of client $i$ and $\mathbb{I}$ denotes the indicator function. The proposed maximum entropy aggregation can be applied to other distribution shift scenarios if we can estimate the empirical entropy of data source~\cite{paninski2003estimation}. How to leverage this aggregation method to benefit the federated generalization for other scenarios is out of the scope of this paper. 

The detailed workflow of the proposed empirical entropy-based weighting method is introduced as follows. Before FL, each participating client  calculates the empirical entropy in Eq.~\eqref{eq20}  based on its local dataset and further uploads the empirical entropy to the server. The server can thus assign the aggregating weighting factors for all the clients rooted on the received empirical entropy prior to the first round of FL. 

In the following, we will discuss the effect of privacy computation, noisy data and computation-communication costs on our proposed method.
    
\textbf{Privacy Computation:} Privacy computation techniques can be integrated into the proposed methods: clients  upload the empirical entropy of local data sources via homomorphic encryption or secure multi-party computation with low computation overhead since the empirical entropy is a scalar value. 

\textbf{Noisy Data Challenges:} If local models are trained on severely noisy data, we can regard such challenge as a form of Byzantine attack~\cite{10506990} and address this issue using corresponding solutions. We can also identify outliers with extremely large entropy and expel them from FL. Additionally,  replacing entropy with a relative metric of calculating the similarity between the local label distribution and  uniform distribution can mitigate the effect from noisy data. Besides, the methods of Noisy Clients Detection and Noisy Robust Aggregation proposed in~\cite{10337775}  can be integrated with our own approaches to alleviate the influence of noisy data.

\textbf{Computation and Communication Costs:}
In this study, we assume that  data sources of participating clients are stationary during the whole FL process. Hence, clients only need to upload empirical entropy to the server for one time, which can reduce the communication cost.

\subsection{Gradient similarity-based client selection }
Below, we will outline the proposed client selection methods based on the theory presented. Before delving into the specifics of these methods, we first introduce an assumption that underpins the forthcoming discussion.
\begin{assumption}
    [Bounded dissimilarity] \label{Assumption} The dissimilarity between two local gradients is bounded by the divergence of corresponding local data distributions, i.e,
    \begin{equation}\label{eq21}
        \begin{aligned}
       \Vert \nabla F_i (\mathbf{w}) -\nabla F_j (\mathbf{w})\Vert^2 \leq \sigma_{\mathcal{K}}^2, \forall i,j \in \mathcal{K},
        \end{aligned}
    \end{equation}
    where $\nabla F_i (\mathbf{w})$ denotes the practical local gradient calculated by client $i$ without considering the self information of samples, i.e., $\nabla F_i (\mathbf{w})=\nabla\ell(\mathbf{w},Z_i)$, where $\mathbf{w}$ denotes the global model and $Z_i$ denotes the $i$-th data source. $\sigma_{\mathcal{K}}^2$ depends on the dissimilarity of data distributions $\{ P_{Z_k}\}_{k \in \mathcal{K}}$. In other words, a higher level of distribution similarity of $\{ P_{Z_k}\}_{k \in \mathcal{K}}$ will incur a lower level of $\sigma_{\mathcal{K}}^2$ \cite{9843892,li2020federated}. This is because that $\Vert \nabla \mathcal{L}_{P_{z_i}}(z,w)-\nabla \mathcal{L}_{P_{z_j}}(z,w)\Vert^2 =\Vert \int_{\mathcal{Z}}\nabla \ell(w,z)(dP_{z_i}-dP_{z_j})\Vert^2\leq c'KL(P_{z_i}\Vert P_{z_j})$, the last inequality holds follows from Lipschitz continuous loss and  Pinsker's inequality. Hence, $\sigma^2_{\mathcal{K}}$ is related to the term $\sup_{i,j \in \mathcal{K}}KL(P_{Z_i}\Vert P_{Z_j})$.   
    
\end{assumption}
Assumption~\ref{Assumption} and Theorem~\ref{the:FedCross} can help us to design client selection methods for improving the generalization performance of FL in the following.

We begin by introducing the general procedure of client selection methods. In the following, we also use  the notation $\mathbf{g}^i_t$ to denote the stochastic gradient calculated by the $i$-th client at the $t$-th round. The server repeats the following steps in each round of FL: 

\begin{itemize}
    \item\textbf{(Update the gradient table):} After updating the global model based on local gradients uploaded by clients $i,\forall i \in \mathcal{I}_t$  following the update rule of $\mathbf{w}_{t+1} \leftarrow \mathbf{w}_t-\sum_{i \in \mathcal{I}_t} \alpha_i\mathbf{g}^i_t $, the server updates and maintains a table that stores the latest local gradients uploaded by the clients selected in each round. More specifically, the server performs the following actions: $\mathbf{g}^i_s\leftarrow \mathbf{g}^i_t, \forall i \in \mathcal{I}_t$ and $\mathbf{g}^i_s\leftarrow \mathbf{g}^i_s , \forall i \in \mathcal{I}_p \setminus \mathcal{I}_t$ to maintain the table $\{\mathbf{g}^i_s \}_{i \in \mathcal{I}_p}$.
     \item\textbf{(Execute the selection algorithm):} The server applies the client selection algorithms described below, utilizing the local gradients stored in the gradient table $\{\mathbf{g}^i_s \}_{i \in \mathcal{I}_p}$ in order to determine the clients $\mathcal{I}_{t+1}$ to participate in the next round of FL.
\end{itemize}

Notice that the server can maintain a dynamic table containing the client IDs, their respective entropy values, and the latest local gradients. This table empowers the server to effectively handle dynamic clients and execute proposed algorithms. Hence, the proposed method is robust to client churn, such as clients frequently joining, leaving, and rejoining the system.

\subsubsection{Minimax gradient similarity-based client selection }

Recall the Theorem~\ref{the:FedCross}, to minimize the distribution discrepancy-aware generalization gap is essentially to maximize the term $ \sum_{i \in \mathcal{I}_t}\min_{j \in \mathcal{I}_p}H(P_{Z_i},P_{Z_j})$, which means that selecting clients with distributions $P_{Z_i},i \in \mathcal{I}_t $ that differ significantly from other participating distributions $P_{Z_j}, \forall j \in \mathcal{I}_p $ can improve the generalization performance of FL. Furthermore, in reference to the Assumption~\ref{Assumption}, this objective can be formulated as a tractable optimization problem, given by
    \begin{align} \label{eq23}
&\min_{\{ a_t^i\}_{i \in \mathcal{I}_p}} \sum_{i \in \mathcal{I}_p }a_t^i\max_{j \in \mathcal{I}_p, j \neq i}S(\nabla F_i (\mathbf{w}),\nabla F_j (\mathbf{w})), \\
 &\quad s.t.\quad a_t^i \in  \{0,1 \} , \forall i \in \mathcal{I}_p; \sum_{i \in \mathcal{I}_p}a_t^i=\vert \mathcal{I}_t\vert. \tag{11a}
\end{align}
where $S(\nabla F_i (\mathbf{w}),\nabla F_j (\mathbf{w}))$ represents the similarity metric between $\nabla F_i (\mathbf{w}) $ and $\nabla F_j (\mathbf{w})$ uploaded by client $i$ and client $j$ respectively, is evaluated using cosine similarity in this work. This choice is motivated by the fact that the distance between bounded local gradients can be bounded by the distribution discrepancy between the corresponding local distributions, i.e, $\Vert \nabla_{\mathbf{w}}\mathcal{L}_{P_{Z_i}}(z;\mathbf{w})-\nabla_{\mathbf{w}}\mathcal{L}_{P_{Z_j}}(z;\mathbf{w})\Vert^2 \leq c'KL(P_{Z_i} \Vert P_{Z_j}) \leq c'H(P_{Z_i},P_{Z_j})$, where $c'$ is a constant. Besides, the cosine similarity is related to the Euclidean distance: $ \Vert\nabla F_i(\mathbf{w})-\nabla F_j(\mathbf{w})\Vert^2=\Vert \nabla F_i(\mathbf{w})\Vert^2-2<\nabla F_i(\mathbf{w}),\nabla F_j(\mathbf{w})>+\Vert \nabla F_j(\mathbf{w})\Vert^2$ and it provides a more stable approach in FL~\cite{DBLP:journals/corr/abs-2303-00897}. We propose a feasible approximate method for solving this optimization problem in Algorithm~\ref{alg1}. Additionally, in Algorithm~\ref{alg1}, we define the concept of "similarity set" denoted as  $\mathbb{S}_i, \forall i \in \mathcal{I}_p$, which is a set of gradient similarities between the local gradient of each participating client and the gradients calculated by the other participating clients. We now provide the workflow of the proposed Minimax gradient similarity-based client selection in the following:
\begin{itemize}
    \item \textbf{(Constructing the similarity set):} First, we develop a "similarity set" $\mathbb{S}_i$ for each stored gradients $\mathbf{g}_s^i,  \forall i \in \mathcal{I}_p$. This "similarity set" $\mathbb{S}_i$ of $i$-th stored gradients $\mathbf{g}_s^i$ contains the cosine similarities $S_{i,j}=\frac{<\mathbf{g}_s^i,\mathbf{g}_s^j>}{\Vert \mathbf{g}_s^i\Vert \Vert \mathbf{g}_s^j\Vert}$ between $\mathbf{g}_s^i$ and all the other stored gradients $\mathbf{g}_s^j, \forall j\neq i, j \in \mathcal{I}_p$, i.e., $\mathbb{S}_i=\{S_{i,j} \}_{j \in \mathcal{I}_p,j \neq i}$. 
    \item \textbf{(Calculating the maximum similarity):} Next, the server  builds another set  $\mathbb{S}_{max}$ to store the maximum similarity $\mathop{\max}\limits_{S_{i,j}}\mathbb{S}_{i}$ in each  $\mathbb{S}_i, \forall i \in \mathcal{I}_p$ while this maximum similarity measures the degree of the similarity between one data source $i$ and other data sources. 
    \item \textbf{(Selecting clients with the smallest maximum similarity):} In the final step, the server determines the clients  whose  maximum similarity $\mathop{\max}\limits_{S_{i,j}}\mathbb{S}_{i}$ is the smallest in the maximum similarity set $\mathbb{S}_{max}$.  
\end{itemize}

The core idea behind the aforementioned operations is to identify data sources that are “distant” from other data sources, thereby approximately achieving the objective stated in~\eqref{eq23}.
\begin{algorithm}[h]
% \scriptsize
\footnotesize
 \SetAlgoLined
  \caption{Minimax Gradient Similarity-based Client Selection Policy}
  \label{alg1}
  \KwIn{set of local gradients $\{ \mathbf{g}_t^i\}_{i \in \mathcal{I}_p}$, the size of selected clients $\vert \mathcal{I}_t\vert$ in round $t$.}
Initialize collection of similarity sets $\{\mathbb{S}_i\}_{i \in \mathcal{I}_p}$, where $\mathbb{S}_i=\emptyset, \forall i \in \mathcal{I}_p $. Max similarity set 
 $\mathbb{S}_{max}=\emptyset$. Candidate set $\Tilde{\mathcal{I}_t}=\emptyset$ \;
\For{$i \in \mathcal{I}_p$}{
\For{$j \in \mathcal{I}_p, j \neq i$}{
$S_{i,j} \leftarrow S(\mathbf{g}_t^i,\mathbf{g}_t^j)$ // Similarity computation\;
$\mathbb{S}_i \leftarrow \mathbb{S}_i \cup \{S_{i,j}\} $ // Similarity set construction
}
$\mathbb{S}_{max} \leftarrow \mathbb{S}_{max} \cup \{ S_i^*=\mathop{\max}\limits_{S_{i,j}} \mathbb{S}_{i}\}$ \;// Max similarity set construction
}
\While{$\vert \Tilde{\mathcal{I}_t} \vert \leq \vert \mathcal{I}_t\vert$}{
$\Tilde{\mathcal{I}_t}  \leftarrow \Tilde{\mathcal{I}_t} \cup \Big\{ \mathop{\arg \min}\limits_{i} \mathbb{S}_{max}  \Big\}$\; 
// Minimax similarity selection\;
$S^*\leftarrow\mathop{\min}\limits_{S_{i}^*} \mathbb{S}_{max} $\;
$\mathbb{S}_{max} \leftarrow \mathbb{S}_{max} \setminus \{ S^*\} $\;
// Remove selected client's maximum similarity
}
\KwRet $\Tilde{\mathcal{I}_t}$
\end{algorithm}

\begin{algorithm}[h]
% \scriptsize
\footnotesize
 \KwIn{ set of local datasets $\{ S_i\}_{i \in \mathcal{I}_p}$, participating client set $\mathcal{I}_p$ with size $\vert \mathcal{I}_p\vert$, the size of selected client set $\vert \mathcal{I}_t\vert$, the initial model $\mathbf{w}_0$, the local learning rate $\eta$.}
     Before the beginning of FL, all the participating clients update local models via $\mathbf{w}_0$ and upload local gradients $\mathbf{g}^i_0$ to the server\;
  \For{round  $t \in [T]$ }{
  $\mathcal{I}_t \leftarrow$    Convex Hull($\{ \mathbf{g}^i_s\}_{i \in \mathcal{I}_p}$) \;
  or $\mathcal{I}_t \leftarrow$    MinimaxSim($\{ \mathbf{g}^i_s\}_{i \in \mathcal{I}_p},\vert \mathcal{I}_t\vert=K$) \;
  \For{client $i \in \mathcal{I}_t$ in parallel}{
  $\mathbf{w}^i_t \leftarrow LocalSolver(\mathbf{w}_t,S_i,\eta )$\;
  $\mathbf{g}^i_t\leftarrow \mathbf{w}_t-\mathbf{w}^i_t$\;}  $\mathbf{w}_{t+1} \leftarrow \mathbf{w}_t-\sum_{i \in \mathcal{I}_t} \alpha_i\mathbf{g}^i_t $;
  
  }
  \caption{Federated Generalization via Client Selection }
  \label{alg2}
\end{algorithm}

\begin{comment}
    \IncMargin{0.1em}
\begin{algorithm}[h]
 \KwIn{ participating client set $\mathcal{I}_p$ with size $\vert \mathcal{I}_p\vert$, the size of selected client set $\vert \mathcal{I}_t\vert$, the initial model $\mathbf{w}_0$, the local learning rate $\eta$.}
  \SetKwProg{myproc}{ServerProcedure}{}{}
  %\SetAlgoLined\DontPrintSemicolon
  \SetKwFunction{algo}{algo}\SetKwFunction{proc}{proc}
  \myproc{}{
     Before the beginning of FL, all the participating clients update local models via $\mathbf{w}_0$ and upload local gradients $\mathbf{g}^i_0$ to the server\;
  \For{round  $t \in [T]$ }{
  $\mathcal{I}_t \leftarrow$    Convex Hul\ell($\{ g^i_{t-1}\}_{i \in \mathcal{I}_p}$) or $\mathcal{I}_t \leftarrow$    MinimaxSim($\{ g^i_{t-1}\}_{i \in \mathcal{I}_p},\vert \mathcal{I}_t\vert$) \;
  \For{client $i \in \mathcal{I}_t$ in parallel}{     $\mathbf{g}^i_t \leftarrow $   LocalUpdate($\mathbf{w}_t$)}  $\mathbf{w}_{t+1} \leftarrow \mathbf{w}_t-\sum_{i \in \mathcal{I}_t} \alpha_i\mathbf{g}^i_t $
  
  }}
\SetKwProg{myalg}{LocalUpdate($\mathbf{w}$)}{}{}
  \myalg{}{
 $\mathbf{w}^i \leftarrow LocalSolver(\mathbf{w},S_i,\eta )$, $\mathbf{g}^i\leftarrow \mathbf{w}-\mathbf{w}^i$\;
\KwRet $\mathbf{g}^i$
}
  \caption{Federated Generalization via Client Selection }
  \label{alg2}
\end{algorithm}
\end{comment}

\subsubsection{Convex hull construction-based client selection}

In the following, we provide another client selection method to enhance federated generalization based on our theoretical findings. Notice that we can also minimize the generalization gap in Theorem~\ref{the:FedCross} via maximizing a term as follows,
\begin{equation}\label{eq24}
   \max_{\mathcal{I}_t} \sum_{i \in \mathcal{I}_t }\sum_{j \in \mathcal{I}_p}H(P_{Z_i},P_{Z_j}).
\end{equation}

The above objective suggests that the larger difference between distributions $P_{Z_i}, \forall i \in \mathcal{I}_t $  and other participating distributions $P_{Z_j},\forall j \in \mathcal{I}_p$, the smaller generalization gap in Theorem~\ref{the:FedCross}. Following the insights gained from our analysis and Assumption~\ref{Assumption}, a convex hull construction-based client selection policy is proposed to enhance the generalization of FL.  The key idea of this method is to identify the vertices of the convex hull of local gradients, and then select the corresponding clients whose gradients are located on the vertices of the constructed convex hull to participate in FL.

Prior to delving into the specifics of the client selection method based on convex hull construction, we will begin by providing the definition of the convex hull.  The convex hull of a set $C$, denoted as $\textbf{conv}C$, refers to the set of all convex combinations of points in $C$~\cite{boyd2004convex}:
\begin{equation}
\begin{aligned}
&\textbf{conv}C:=\\
&\{ \lambda_1x_1+...+ \lambda_nx_n: x_i \in C, \lambda_i\geq 0,i=1,...,n,\sum_{i=1}^n\lambda_i=1\}.\\
\end{aligned}
\end{equation}

\begin{figure}[ht]
    \centering
    \includegraphics[width=1.in]{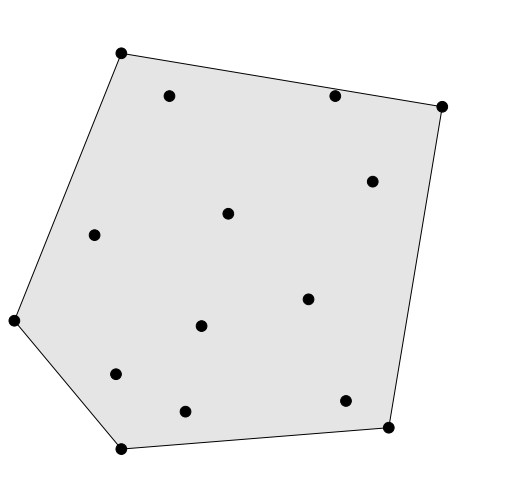}
    \caption{The convex hull of a point set in $\mathbb{R}^2$. The convex hull of a point set of $15$ points is the pentagon (shown shaded).}
    \label{fig:convexhullexapmle}
\end{figure}

Notice that the convex hull of a point set $C$ represents the smallest convex set that encompasses all the points in $C$. The definition of the convex hull is visually depicted in Figure~\ref{fig:convexhullexapmle}.

%The key idea of the proposed convex hull construction-based client selection is to identify the vertices of the convex hull formed by the stored gradients in the server’s table. Subsequently, the clients whose gradients are located at these vertices of the constructed convex hull are selected.

The formal work process of the proposed convex hull construction-based client selection  is outlined as follows. a) The server initiates the execution of the quickhull algorithm proposed in~\cite{barber1996quickhull} to construct the convex hull of the local gradients stored in the gradient table $\{\mathbf{g}^i_s \}_{i \in \mathcal{I}_p}$. b) Once  the vertices of the considered point set, which comprises 
 all the gradients stored in  $\{\mathbf{g}^i_s \}_{i \in \mathcal{I}_p}$ are identified, the server proceeds to select the corresponding clients whose gradients are located on these discovered vertices. These selected clients are then required to participate in the subsequent round of FL. 

Intuitively, the distances between points located on the vertices of the convex hull and other points tend to be larger. This observation served as inspiration for our client selection method. Potential gradients generated by unseen non-participating clients can be considered as "random points" occurring within or around a given point set. By utilizing the vertices of the convex hull, we can  more effectively "cover" these "random points", providing a geometric perspective to further explain our method.

%From an alternative interpretation, any points in a point set $C$ can be linearly represented by the convex hull $\textbf{conv}C$, which is the smallest convex set contains $C$. This implies that only the gradients located on the vertices of the constructed convex hull can linearly represent gradients of the entire set. In other words, the data sources that generate gradients located on the vertices of the convex hull can effectively represent all the data sources, thereby reducing the intrinsic information redundancy.

We outline the general procedure of two proposed client selection methods in Algorithm~\ref{alg2}. Then we provide some insights of reducing the communication and computation cost of these client selection methods.

\textbf{Communication and Computation cost:} To further reduce the communication and computation cost of proposed methods, we can employ the  event-triggered communication techniques~\cite{9669076,he2023asymptotic} widely used in distributed optimization to diminish the overhead caused by frequent communication of local gradients. Moreover, based on event-triggered communication, the server can utilize historical gradient similarities stored in another table to avoid redundant computations.

\section{Experiment}\label{sec:experiment}

In this section, we first evaluate the proposed methods on three common datasets in FL, in order to verify our theoretical results. Then we compare the proposed methods with more baselines on CIFAR-100 dataset to further validate the generalization performance of our methods.

\textbf{Experiment setting:} We fisrt consider three datasets commonly used in FL: i) image classification on EMNIST-10 and CIFAR-10 with a CNN model, ii) next character prediction on Shakespeare with a
RNN model. For the image classification task, we split the dataset into different clients using the Dirichlet distribution spitting method, and we compare the proposed weighting aggregation method and client selection methods with baselines provided below. For the Shakespeare task, each speaking role in each play is set as a local dataset~\cite{DBLP:journals/corr/abs-1812-01097}, and we only compare the proposed client selection methods with baselines on this dataset. We split all datasets into $100$ clients, and randomly select $40$ clients among them as  participating clients. The remaining $60$ clients  are considered non-participating clients.  We evaluate the global model's performance on two metrics: In Distribution (ID) performance and Out-of-Distribution (OOD) performance. The ID performance evaluates the global model on the local test set of selected clients, while the OOD performance evaluates the global model on a standard test set with a distribution equivalent to the total dataset of $100$ clients. All the selected clients perform $5$ local epochs before sending their updates. We use a batch size of $128$ and tune the local learning rate $\eta$ over a $\{ 0.1, 0.01,0.001\}$ grid in all experiments.

Then we introduce some baseline methods: a) \textbf{Random selection:} Participating clients are selected randomly with equal probability. b) \textbf{Maximum gradient similarity-based selection:} This baseline evaluates the minimax gradient similarity-based client selection method. The server selects clients with the most similar local gradients. c) \textbf{Interior selection:} This baseline evaluates the convex hull construction-based selection method. The server constructs the convex hull of local gradients and then selects clients with gradients in the interior randomly. d) \textbf{Full sampling:} All participating clients will be selected in each round. e) \textbf{Power-of-Choice selection:} The server selects clients with the largest loss values in the current round~\cite{DBLP:journals/corr/abs-2010-01243}. f) \textbf{Data size-based weighting:} The weighting factor of each client is proportional to the data size of its local dataset. g) \textbf{Equality weighting:} The weighting factor of each client in aggregation is set to $1/\vert \mathcal{I}_t \vert$.

\textbf{Experiment Results:} Table~\ref{tab:table1} reports the ID and OOD test accuracy of client selection methods on three datasets. The results in Table~\ref{tab:table1} indicate that, with respect to OOD test accuracy, the proposed client selection methods outperform random selection, Power-of-Choice selection, and other baselines.  Table~\ref{tab:table2} demonstrates that the proposed empirical entropy-based weighting method surpasses the data size-based weighting and equality weighting method on OOD test accuracy. This outcome aligns with our theoretical results and confirms that local models trained on distributions with greater information entropy contribute more significantly to federated generalization.

\begin{figure} [ht]
\vspace{-0.1in}
	\centering
         \hspace{-0.55in}
   \subfloat[EMNIST-10]{\includegraphics[width=0.385\linewidth]{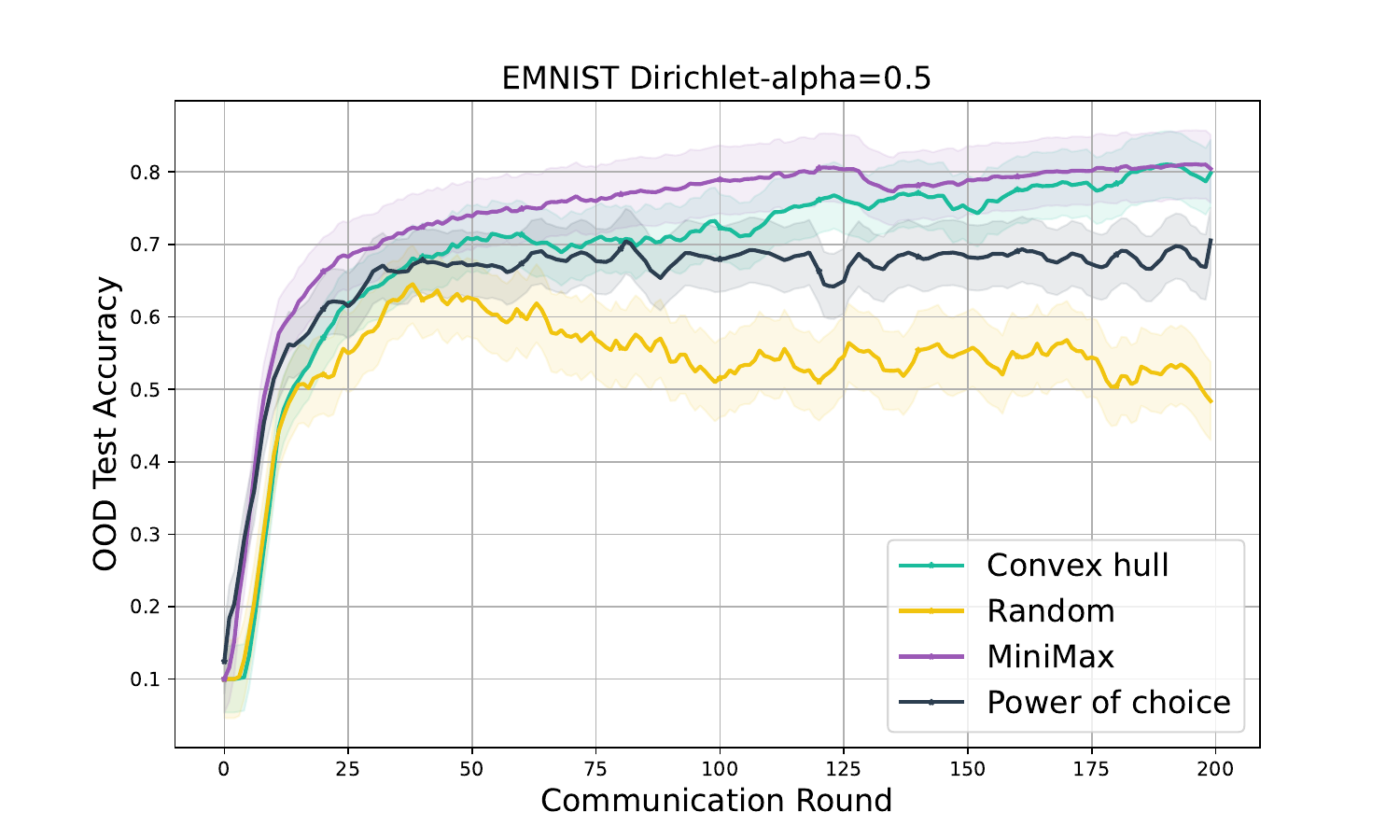} }
   \hspace{-0.23in}
   \subfloat[CIFAR-10]{\includegraphics[width=0.385\linewidth]{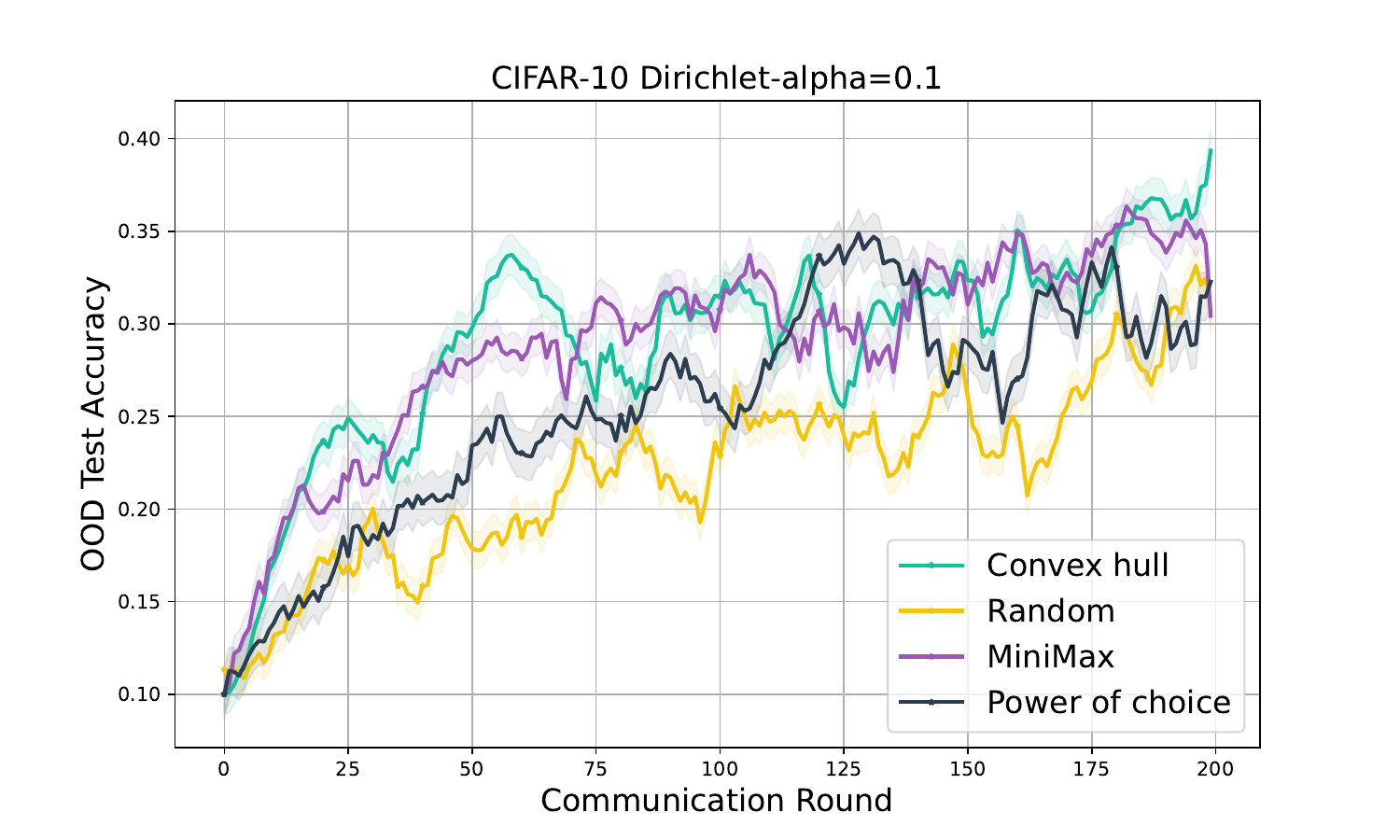} }
      \hspace{-0.23in}
      \subfloat[Shakespeare]{\includegraphics[width=0.385\linewidth]{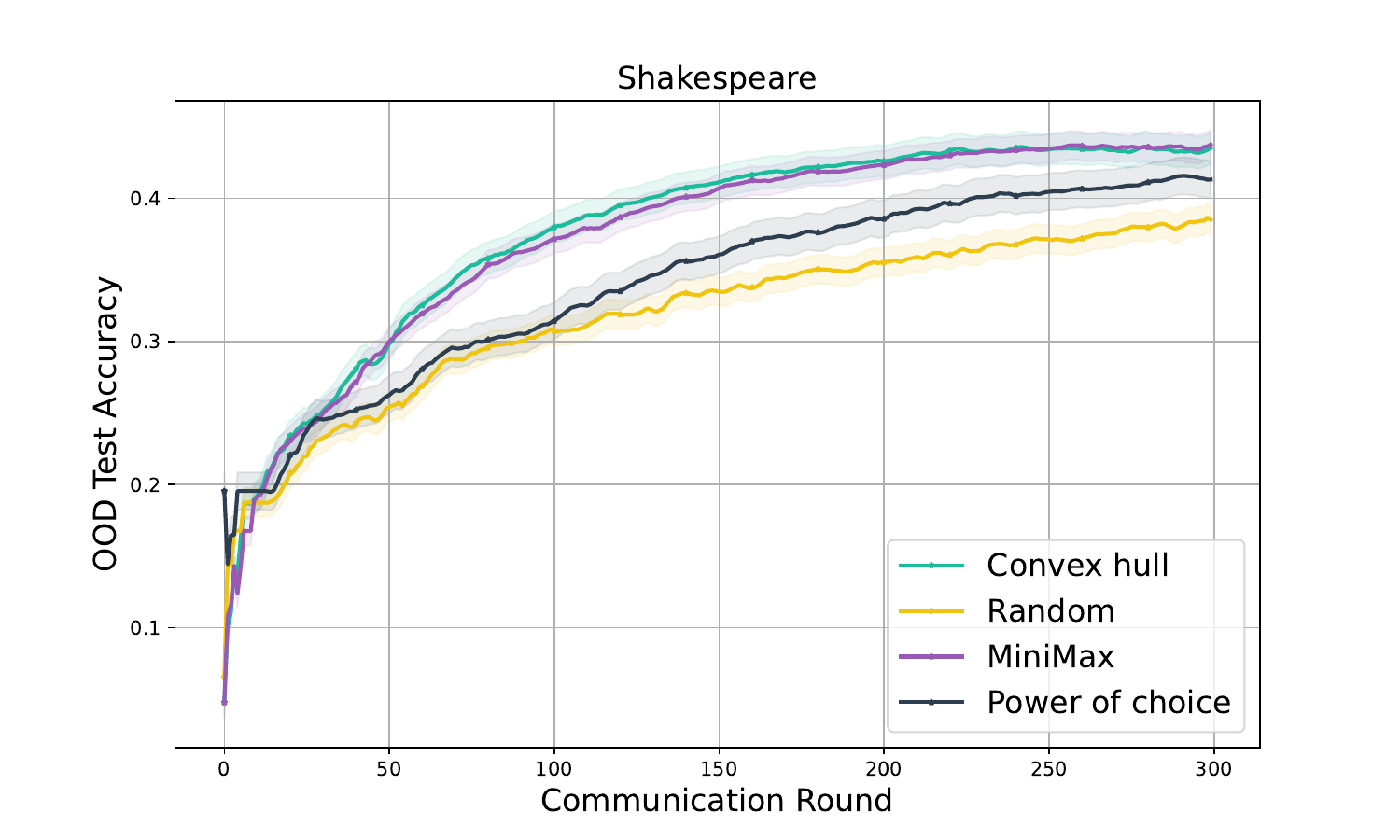} }
           \hspace{-0.63in}
	\caption{ The convergence analysis on OOD test accuracy of two proposed client selection methods compared with random selection and power-of-choice selection.  }
	\label{fig4} 
        \vspace{-0.2in}
\end{figure}

\begin{figure} [ht]
	\centering
     \hspace{-0.55in}
   \subfloat[EMNIST-10]{\includegraphics[width=0.385\linewidth]{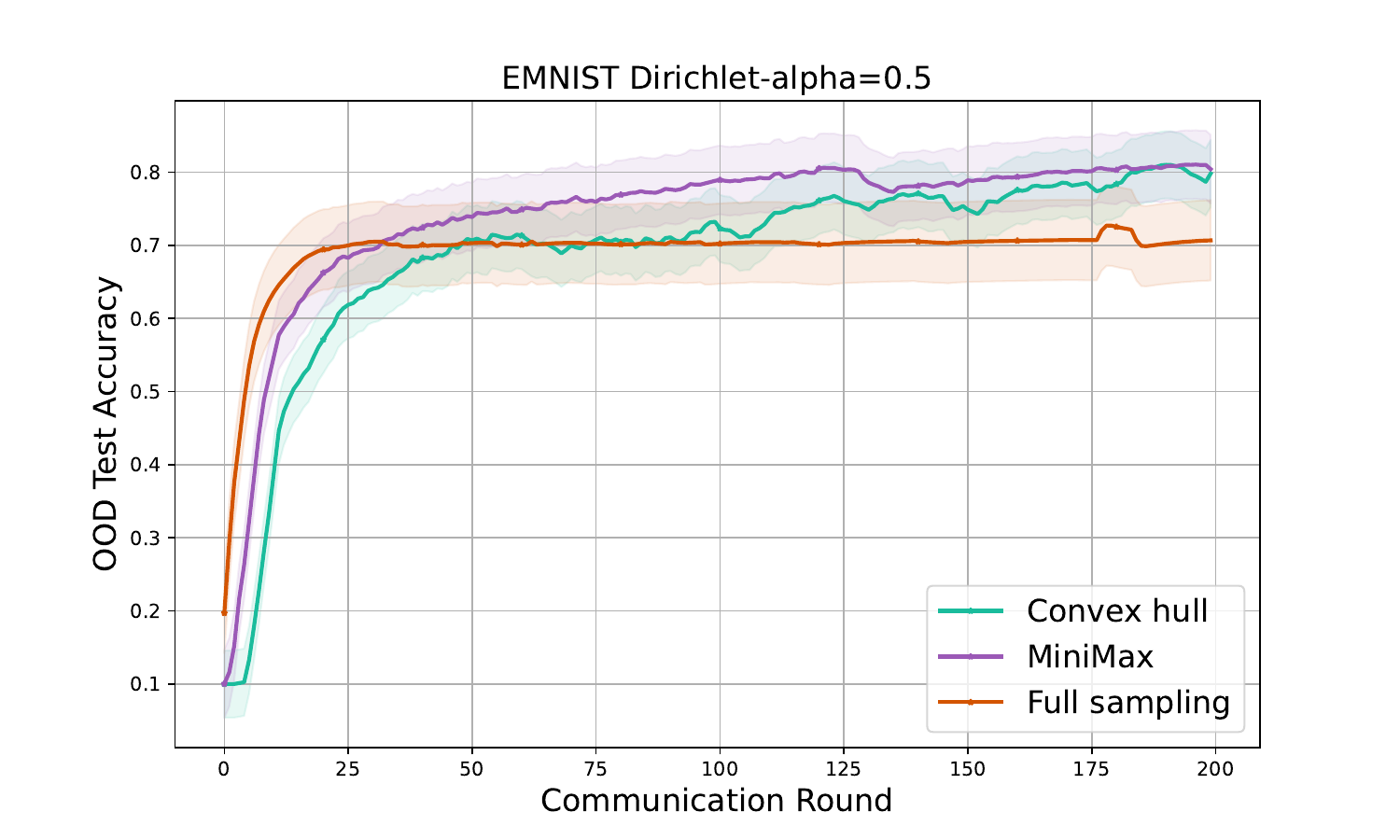} }
      \hspace{-0.23in}
   \subfloat[CIFAR-10]{\includegraphics[width=0.385\linewidth]{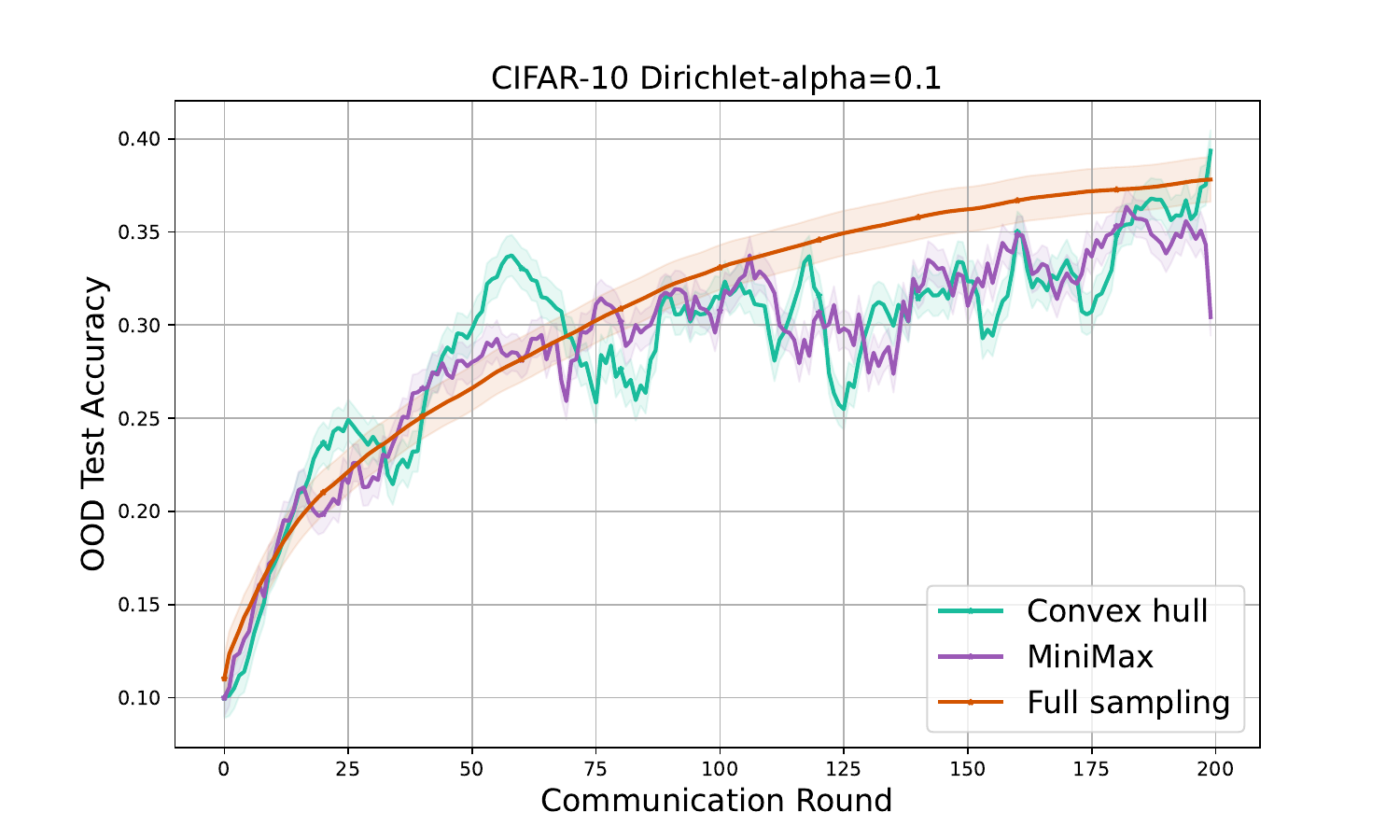} }
      \hspace{-0.23in}
      \subfloat[Shakespeare]{\includegraphics[width=0.385\linewidth]{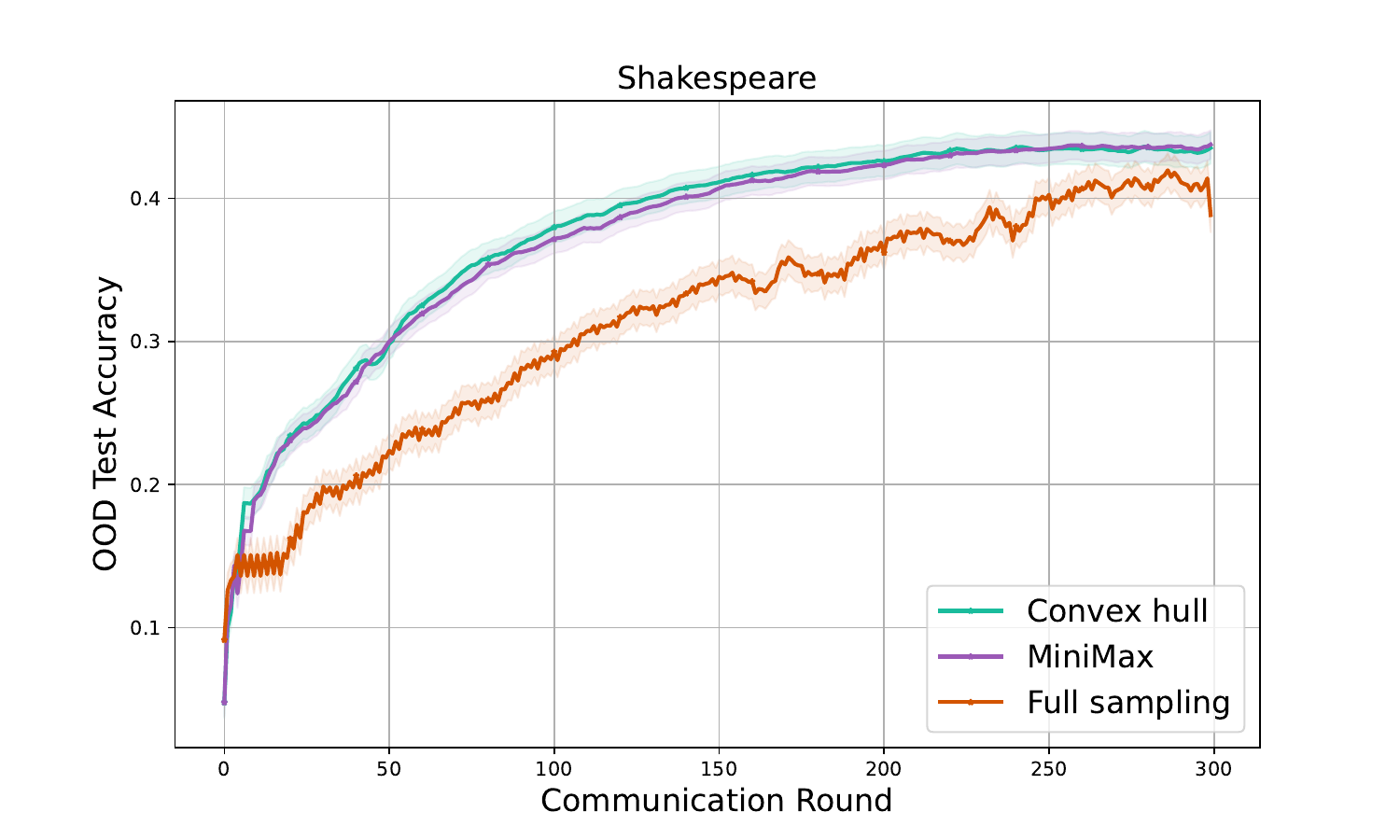} }
        \hspace{-0.63in}
	\caption{ The convergence analysis on OOD test accuracy of two proposed client selection methods in comparison with Full sampling.  }
	\label{fig5} 
    \vspace{-0.2in}
\end{figure}

\begin{figure} [ht]
	\centering
                \hspace{-0.55in}
   \subfloat[EMNIST-10]{\includegraphics[width=0.385\linewidth]{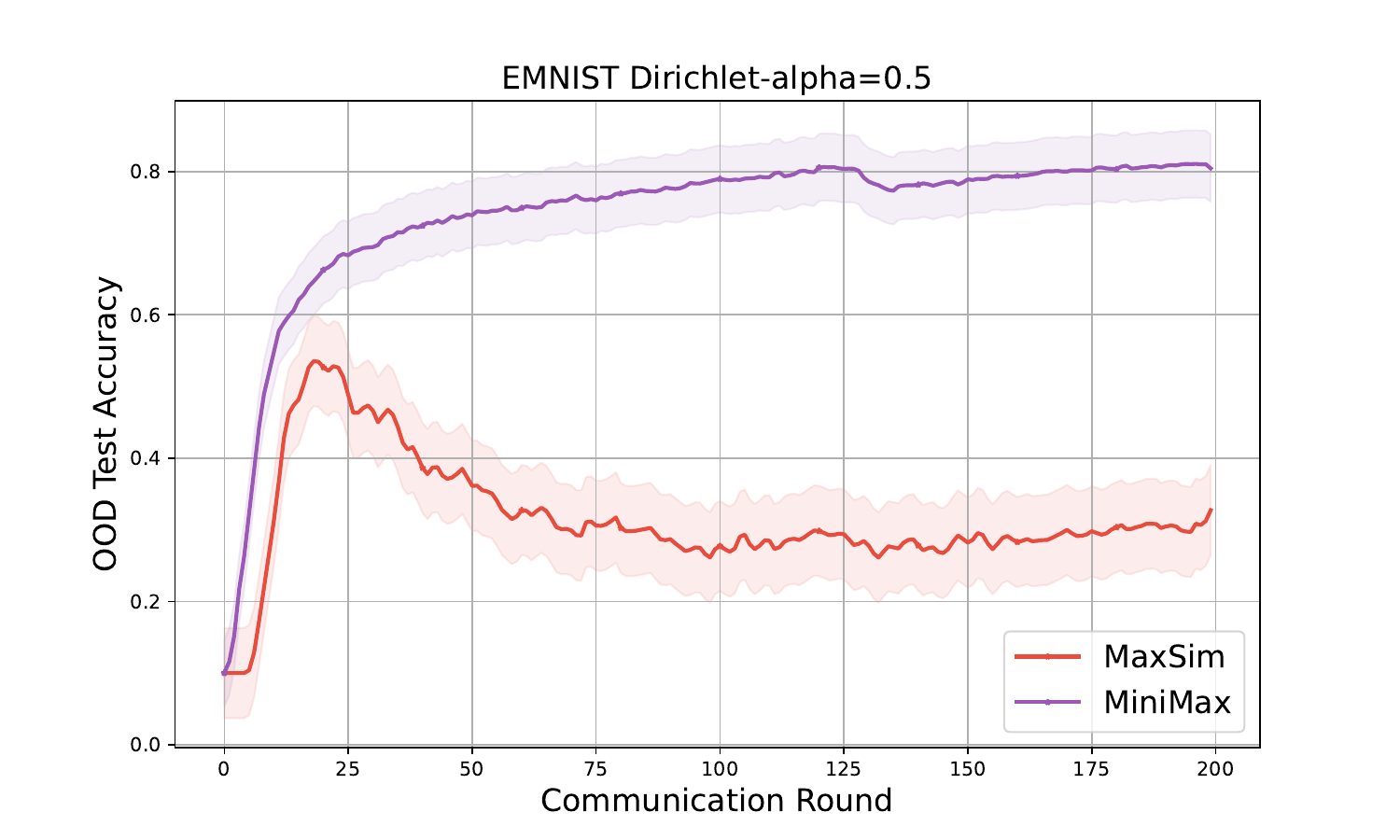} }
      \hspace{-0.23in}
   \subfloat[CIFAR-10]{\includegraphics[width=0.385\linewidth]{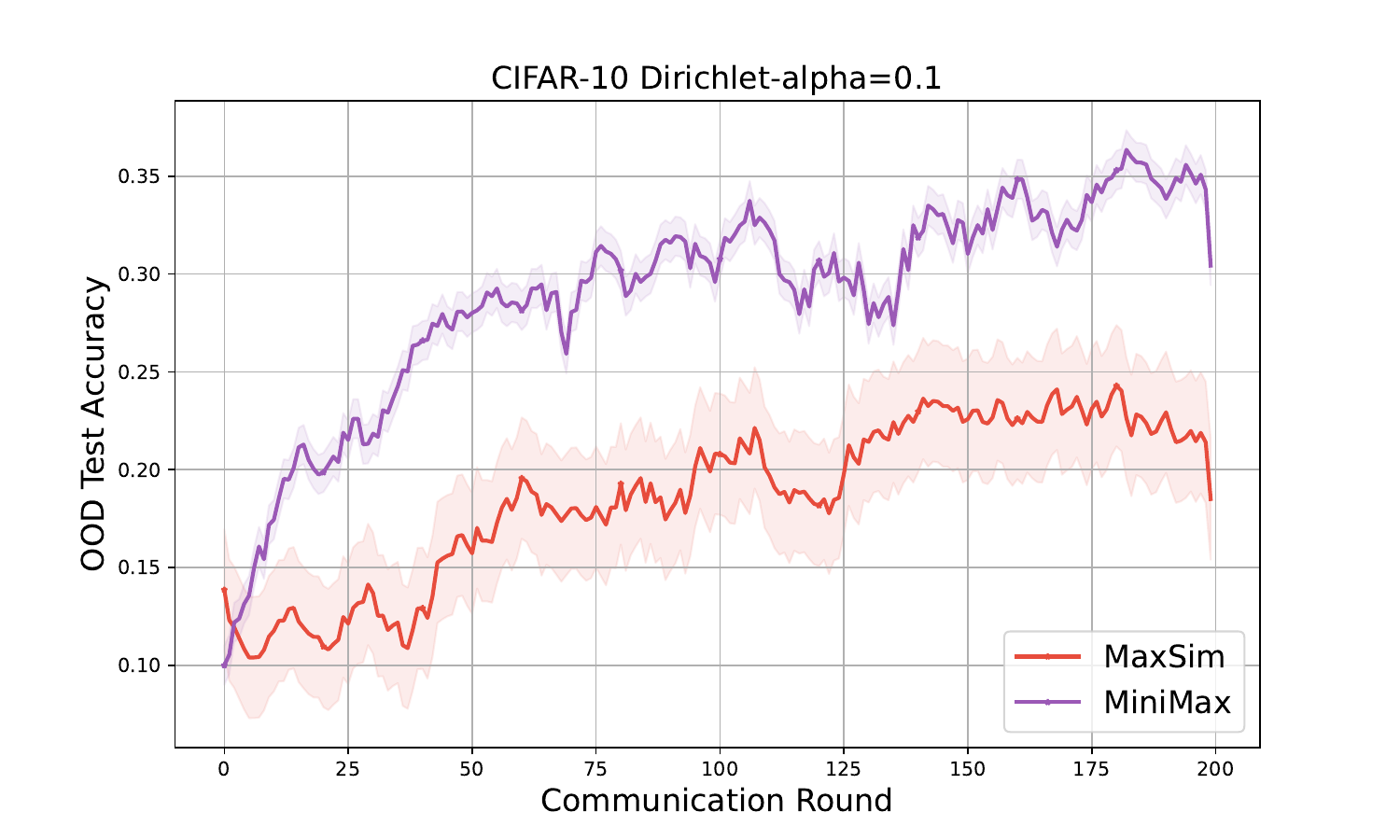} }
      \hspace{-0.23in}
      \subfloat[Shakespeare]{\includegraphics[width=0.385\linewidth]{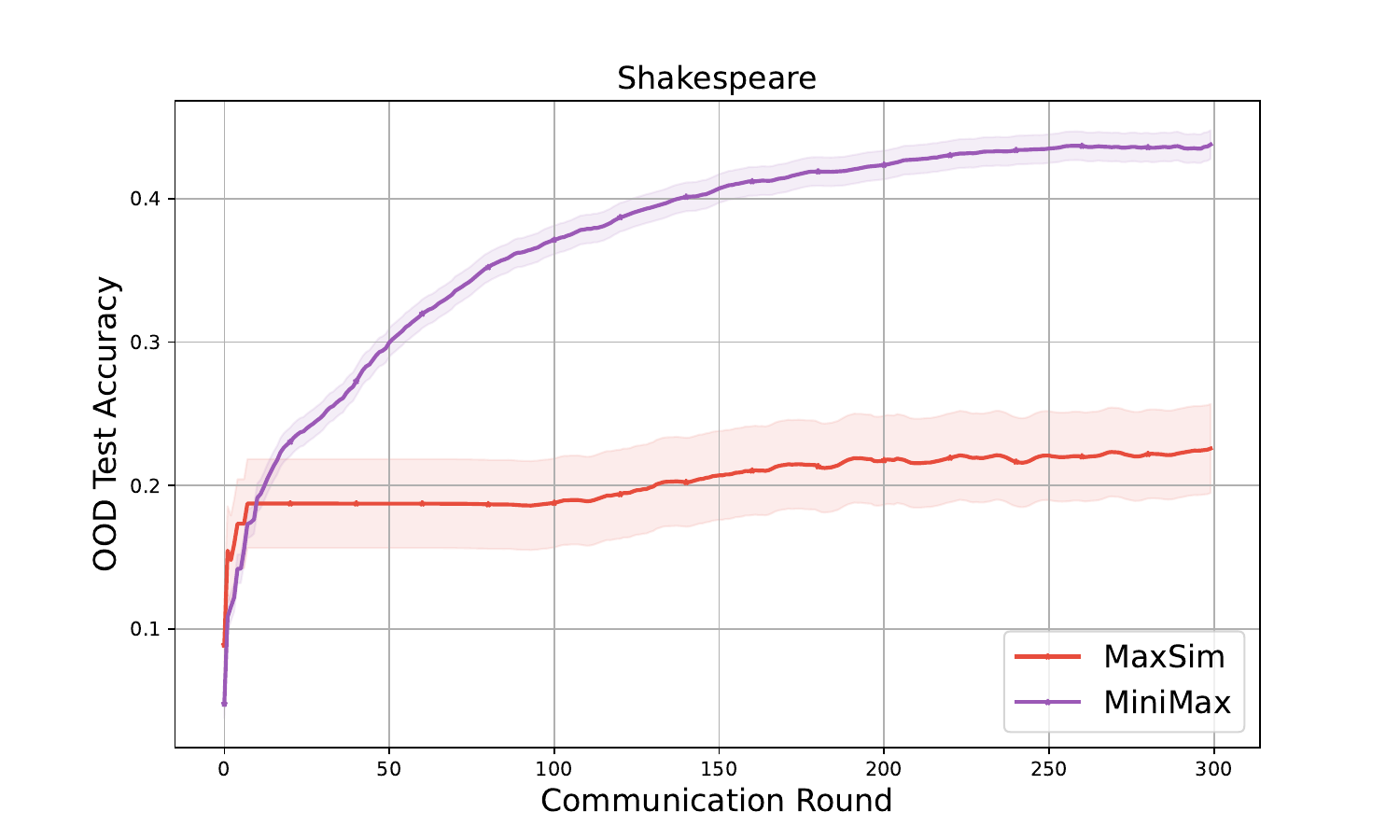} }
       \hspace{-0.63in}

             \hspace{-0.55in}
       \subfloat[EMNIST-10]{\includegraphics[width=0.385\linewidth]{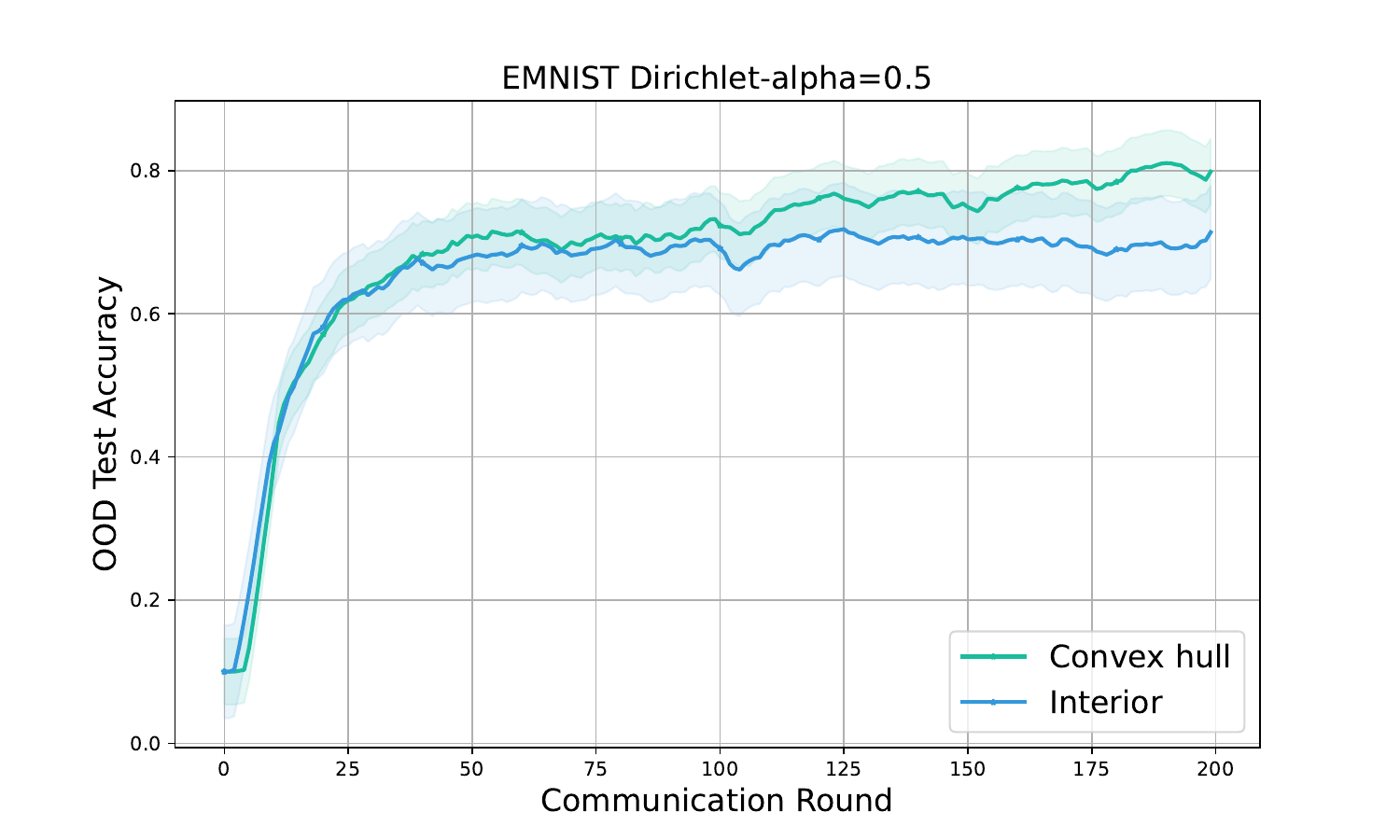} }
          \hspace{-0.23in}
   \subfloat[CIFAR-10]{\includegraphics[width=0.385\linewidth]{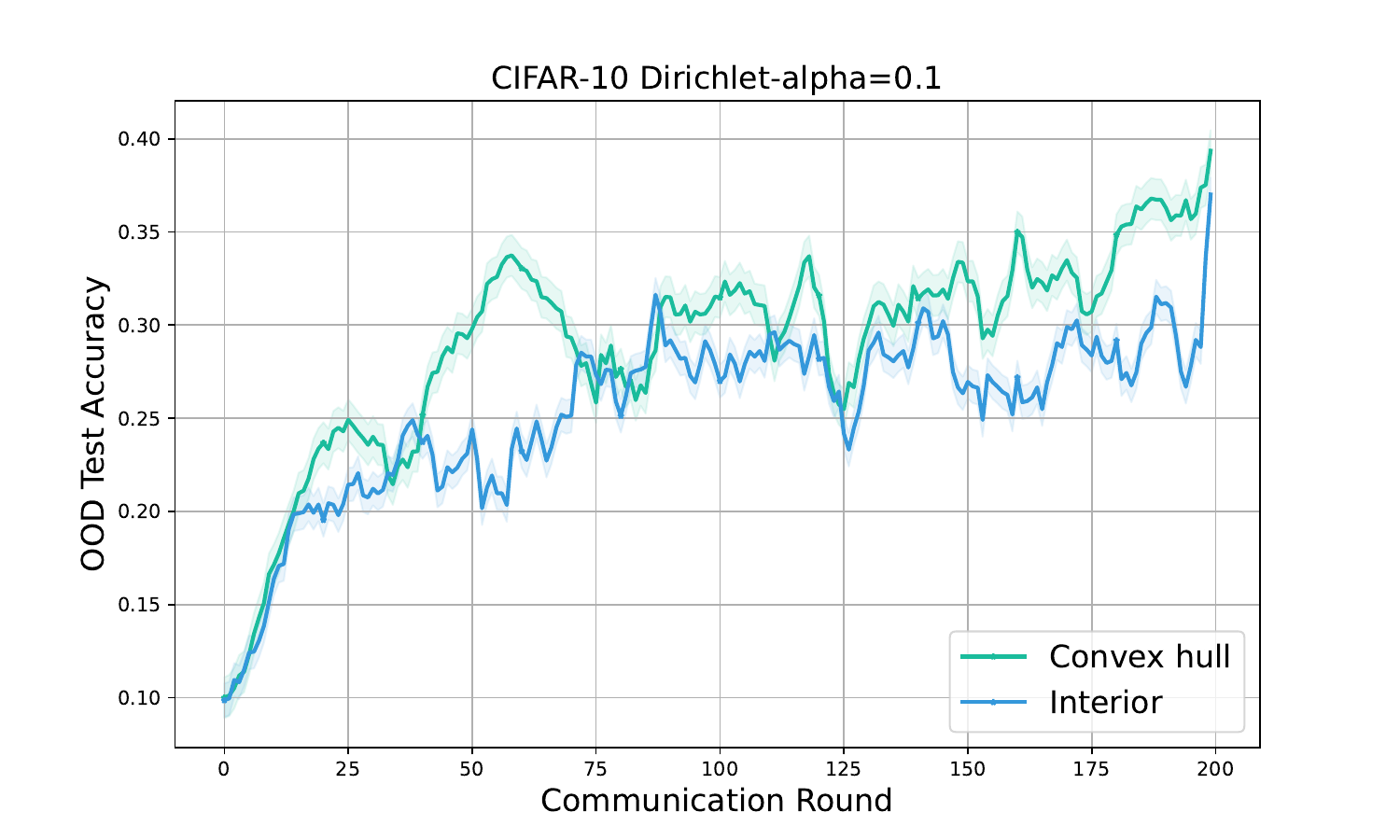} }
      \hspace{-0.23in}
      \subfloat[Shakespeare]{\includegraphics[width=0.385\linewidth]{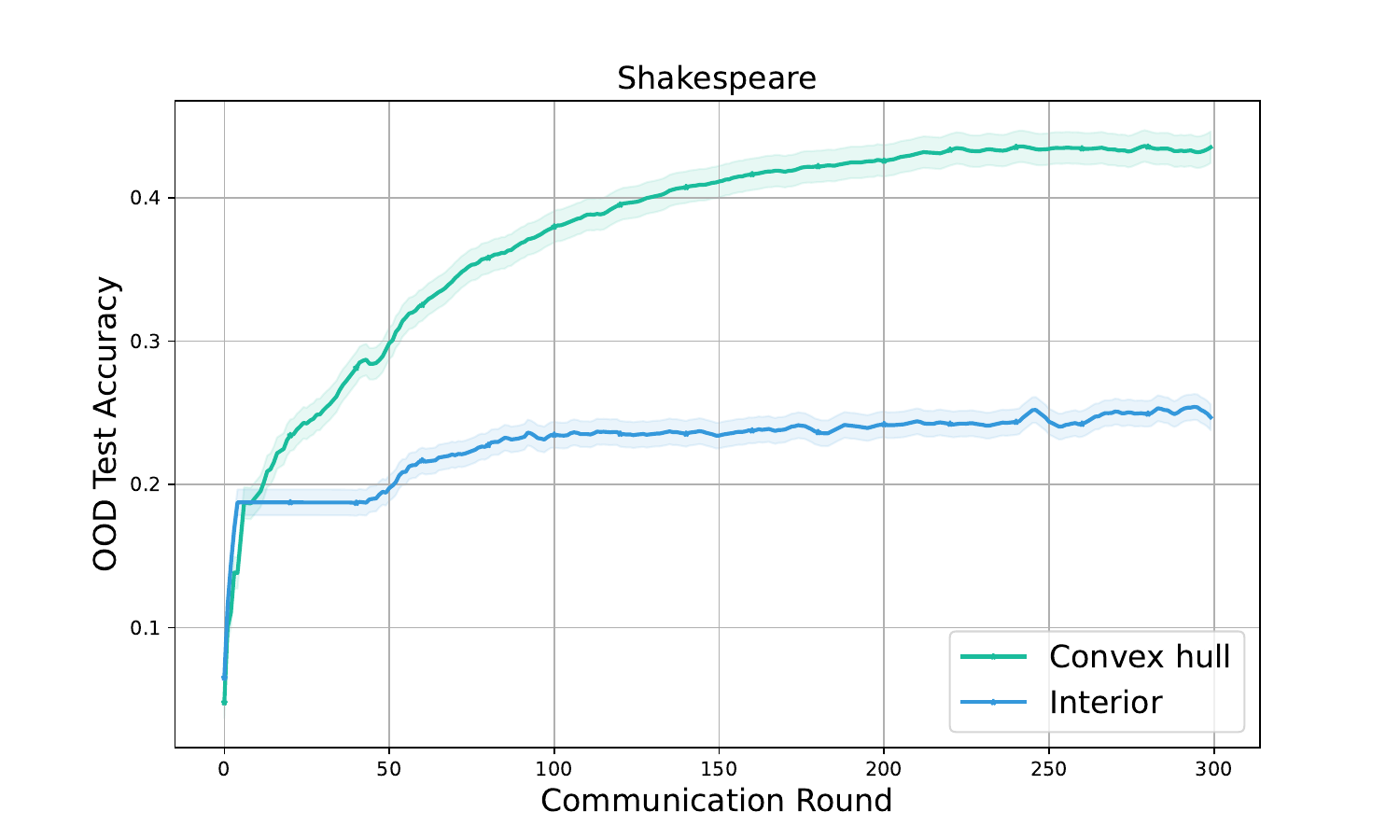} }
           \hspace{-0.63in}
	\caption{ The convergence analysis on OOD test accuracy of the ablation studies on two proposed client selection methods.}
	\label{fig6} 
\end{figure}

\begin{comment}
    
\begin{figure} [ht]
	\centering
	 \subfloat[Shakespeare]{\includegraphics[width=3.2in]{fig/SHAKESPEARE_Full_PoW_selection}}
  \subfloat[EMNIST-10]{\includegraphics[width=3.2in]{fig/EMNIST-10_Full_PoW_selection.pdf} }
	\caption{ The convergence analysis on OOD test accuracy of two proposed client selection methods compared with random selection and other proposed baselines on Shakespeare and EMNIST-10. }
	\label{fig4} 
\end{figure}

\begin{figure} [ht]
	\centering
	 \subfloat[CIFAR-10, comparison experiment with Full sampling]{\includegraphics[width=3.2in]{fig/CIFAR_FULL_selection}} 
  \subfloat[CIFAR-10, comparison experiment with Power-of-Choice selection]{\includegraphics[width=3.2in]{fig/CIFAR_PoW_selection} }
    \subfloat[CIFAR-10, comparison experiment with Random selection]{\includegraphics[width=3.2in]{fig/CIFAR_random_selection.pdf}} 
	\caption{ The convergence analysis on OOD test accuracy of two proposed client selection methods in comparison with Full sampling, Power-of-choice selection, and random selection respectively.}
	\label{fig5} 
\end{figure}

\begin{figure} [ht]
	\centering
     \subfloat[CIFAR-10, ablation study of Convex hull selection]{\includegraphics[width=3.2in]{fig/CIFAR_convexhull_ablation.pdf}} 
     \subfloat[CIFAR-10, ablation study of MinimaxSim selection]{\includegraphics[width=3.2in]{fig/CIFAR_minimax_ablation.pdf} } 
	\caption{ The convergence analysis on OOD test accuracy of the ablation studies on two proposed client selection methods  respectively.}
	\label{fig6} 
\end{figure}
\end{comment}

\begin{table*}[h]
\caption{Test accuracy (\%) $\pm$ std of client selection methods}
    \label{tab:table1}
\centering
\begin{adjustbox}{width=1\width}
\begin{tabular}{clcccccc}
\toprule
\multicolumn{2}{c}{\multirow{2}{*}{Method}} & \multicolumn{2}{c}{EMNIST-10}                                     & \multicolumn{2}{c}{CIFAR-10}                                      & \multicolumn{2}{c}{Shakespeare}                                   \\ \cmidrule(r){3-4} \cmidrule(r){5-6} \cmidrule(r){7-8}
\multicolumn{2}{c}{}                        & \multicolumn{1}{c}{In Distribution} & \multicolumn{1}{c}{Out-of-Distribution} & \multicolumn{1}{c}{In Distribution} & \multicolumn{1}{c}{Out-of-Distribution} & \multicolumn{1}{c}{In Distribution} & \multicolumn{1}{c}{Out-of-Distribution} \\ \midrule
\multicolumn{2}{c}{Convex Hull (ours)}              & 91.8±0.8                      & \multicolumn{1}{c}{\textbf{82.1±4.6}}      & 50.7±3.7                      & \multicolumn{1}{c}{\textbf{42.9±1.1}}      & 56.1±2.0                     &\textbf{43.6±1.0}                          \\ %\hline
\multicolumn{2}{c}{MiniMaxSim (ours)}              & 95.5±0.7                      & \multicolumn{1}{c}{\textbf{82.3±4.7}}      & 49.9±3.0                      & \multicolumn{1}{c}{\textbf{42.0±1.0}}      & 55.0±1.9                      &\textbf{43.5±0.8 }                         \\ %\hline
\multicolumn{2}{c}{Random Selection}                  & 95.9±0.9                      & \multicolumn{1}{c}{69.4±5.4}      & 49.9±2.6                      & \multicolumn{1}{c}{38.9±1.0}      & 45.2±0.6                      & 37.3±1.1                          \\ %\hline
\multicolumn{2}{c}{MaxSim}                  & 95.3±0.5                      & \multicolumn{1}{c}{59.8±6.3}      & 60.9±2.1                      & \multicolumn{1}{c}{32.2±3.1}      & 30.2±0.8                      & 22.2±1.4                          \\ %\hline
\multicolumn{2}{c}{Interior}                & 96.6±0.5                      & \multicolumn{1}{c}{73.8±6.5}      & 50.4±5.7                      & \multicolumn{1}{c}{40.8±0.9}      & 33.5±3.6                      & 25.4±0.8                          
  \\ %\hline
\multicolumn{2}{c}{Full Sampling}                &  98.7±0.3                      & \multicolumn{1}{c}{ 80.1±5.5}      & 61.4±6.8                      & \multicolumn{1}{c}{41.6±1.2}      & 53.7±2.3                      & 43.2±1.8  
 \\ %\hline
\multicolumn{2}{c}{Power-of-Choice}                &  97.3±0.6                      & \multicolumn{1}{c}{ 76.3±4.5}      & 60.3±6.7                      & \multicolumn{1}{c}{39.2±1.3}      & 56.7±2.6                      & 42.6±1.7  \\\bottomrule
\end{tabular}
\end{adjustbox}
\end{table*}

\begin{table}[h]
\caption{Test accuracy (\%) $\pm$ std of weighting aggregation methods }
    \label{tab:table2}
    \centering
\begin{adjustbox}{width=1\width}
\begin{tabular}{clcccc}
\toprule
\multicolumn{2}{c}{\multirow{2}{*}{Method}} & \multicolumn{2}{c}{EMNIST-10}       & \multicolumn{2}{c}{CIFAR-10} \\ %\cline{3-6} 
\cmidrule(r){3-4}
\cmidrule(r){5-6} 
\multicolumn{2}{c}{}                        & ID   & \multicolumn{1}{c}{OOD }    & ID           & OOD         \\ \midrule

\multicolumn{2}{c}{Entropy (ours)}                 & 96.7±0.2 & \multicolumn{1}{c}{\textbf{74.9±10.5}} & 56.5±0.7    & \textbf{35.7±1.0}      \\ %\hline
\multicolumn{2}{c}{Data size}                & 94.7±1.8  & \multicolumn{1}{c}{53.1±12.1} &53.3±3.2        & 33.2±0.3       \\ %\hline
\multicolumn{2}{c}{Equality}                & 96.2±0.6 & \multicolumn{1}{c}{70.7±13.2} & 60.8±3.9        & 34.9±0.9      \\ \bottomrule
\end{tabular}
\end{adjustbox}
\end{table}

\begin{figure} [ht]
\vspace{-0.1in}
	\centering
    \hspace{-0.35in}
   \subfloat[EMNIST-10]{\includegraphics[width=0.55\linewidth]{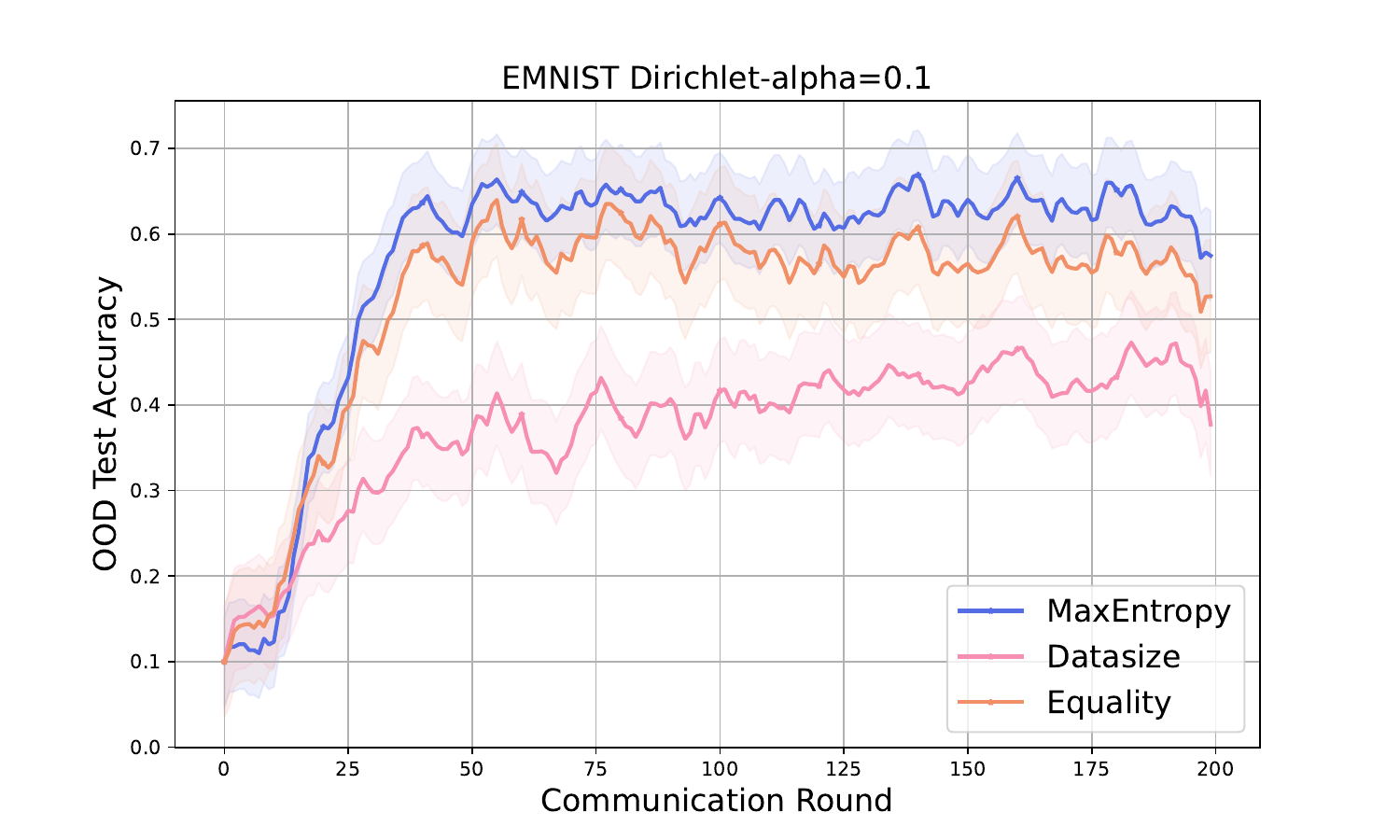} }
   \hspace{-0.2in}
   \subfloat[CIFAR-10]{\includegraphics[width=0.55\linewidth]{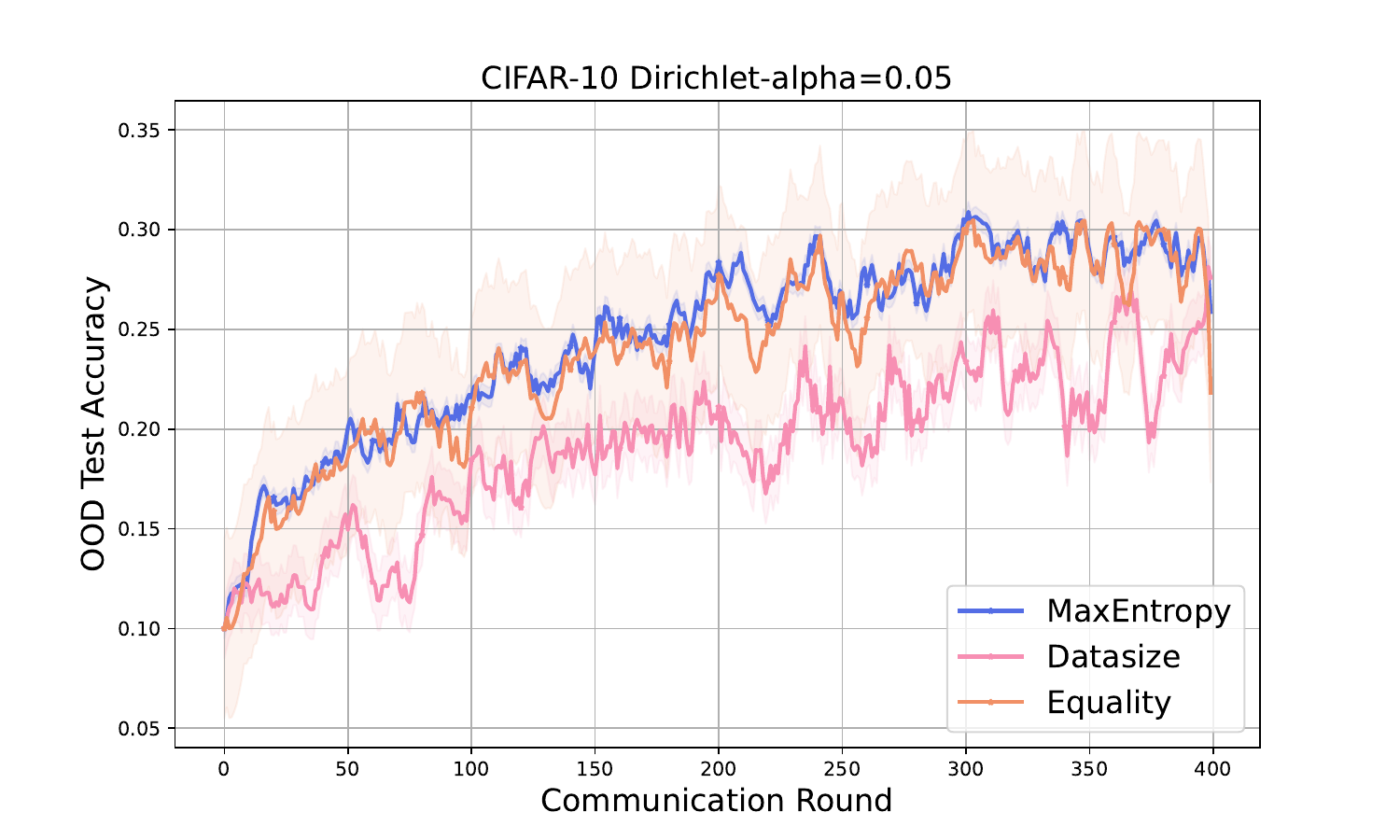} }
     \hspace{-0.5in}
	\caption{ The convergence analysis on OOD test accuracy of empirical entropy weighting method compared with other weighting aggregation methods on EMNIST-10 and CIFAR-10. }
	\label{fig3} 
            \vspace{-0.1in}
\end{figure}

Then we perform the convergence analysis on weighting aggregation methods for EMNIST-10 and CIFAR-10. The label distribution skew of FL is considered in this part. More specifically, we split the total training set into different clients via the Dirichlet distribution spitting. The splitting parameter $\alpha$ of the Dirichlet distribution is set as $0.1$ and $0.05$ for EMNIST-10 and CIFAR-10 respectively. The convergence behavior of the proposed empirical entropy-based weighting method compared with other baselines for EMNIST-10 and CIFAR-10 is presented in Figure~\ref{fig3}. For EMNIST-10, the proposed empirical entropy-based weighting method converges faster than the other two baselines and it also maintains the highest OOD test accuracy among these weighting methods after about $20$-th communication round. For CIFAR-10, both the proposed empirical entropy-based weighting method and equality weighting method converge faster than the data size-based weighting method while the proposed weighting method converges more stably than other baselines. The above results show that giving higher aggregation weights to local gradients trained on data sources with greater information entropy will improve the generalization performance of  models, which is matched with our theoretical basis in Theorem~\ref{the:FedEntropy}.

% \begin{figure} [ht]
% \vspace{-0.1in}
% \centering
%   \hspace{-0.35in}
%    \subfloat[Client selection method]{\includegraphics[width=0.56\linewidth]{Figure/NICO_selection.pdf} }
%    \hspace{-0.32in}
%    \subfloat[Weighting method]{\includegraphics[width=0.56\linewidth]{Figure/NICO_weighting.pdf} }
%       \hspace{-0.5in}
% 	\caption{The convergence analysis on OOD test accuracy of empirical entropy weighting method and two proposed client selection methods compared with baselines on the NICO dataset.}
% 	\label{fig:nico} 
% \end{figure}

We then compare the convergence behavior of different client selection methods, and conduct the ablation studies for our proposed client selection methods on three datasets. The experimental results are shown in Figure~\ref{fig4},~\ref{fig5} and~\ref{fig6} resepectively. For EMNIST-10 and CIFAR-10, we still split the total training set via the Dirichlet distribution spitting. The splitting parameter $\alpha$ of the Dirichlet distribution is set as $0.5$ and $0.1$ for EMNIST-10 and CIFAR-10 respectively in this part. For the Shakespeare dataset, each speaking role in each play is set as a local dataset. 

We first focus on the Shakespeare and EMNIST-10 datasets. We can find that two proposed client selection methods almost converge at the same rate and converge faster than full sampling, power-of-choice selection and random selection. It shows the superiority of our proposed methods and indicates that the empirical results are matched with the presented theoretical findings. It can be also found that the full sampling scheme converges fastest and the most stably on EMNIST-10 dataset. However, it achieves worse OOD test accuracy than the proposed methods since the randomness induced by the selection will even improve the out-of-distribution performance of the global model. From another perspective, the nature of the proposed methods is to "compress" the information from participating data sources, i.e., removing the redundant information from certain data sources making less contribution to generalization. For CIFAR-10, we notice that the proposed methods perform much better than random selection and achieve better OOD test accuracy than power-of-choice selection in most of rounds. However, the proposed methods perform worse than full sampling for stability. The reason why the full sampling scheme performs more stably for CIFAR-10 is that training a model performing well on unseen distributions for CIFAR-10 is the most difficult among three tasks. Consequently, more participating clients will generalize the global model to unseen data source better. 

Furthermore, we  carry out the ablation studies for two proposed client selection methods on three datasets. The results in Figure~\ref{fig6} show that two proposed client selection methods converge faster than their ablation baselines. It indicates that selecting clients with more dissimilar local gradients and selecting clients with local gradients in the convex hull not in the interior will both improve the out-of-distribution generalization performance of the global model. According to Assumption~\ref{Assumption} about the relationship between the gradient dissimilarity and the distribution discrepancy, we can get the following conclusion immediately: selecting clients with more diverse local distributions will enhance the generalization capacity of FL, so the trained model can provide better service for non-participating clients.

\begin{figure} [ht]
\vspace{-0.1in}
	\centering
     \hspace{-0.5in}
   \subfloat[Empirical entropy weighting method]
   {\includegraphics[width=0.38\linewidth]{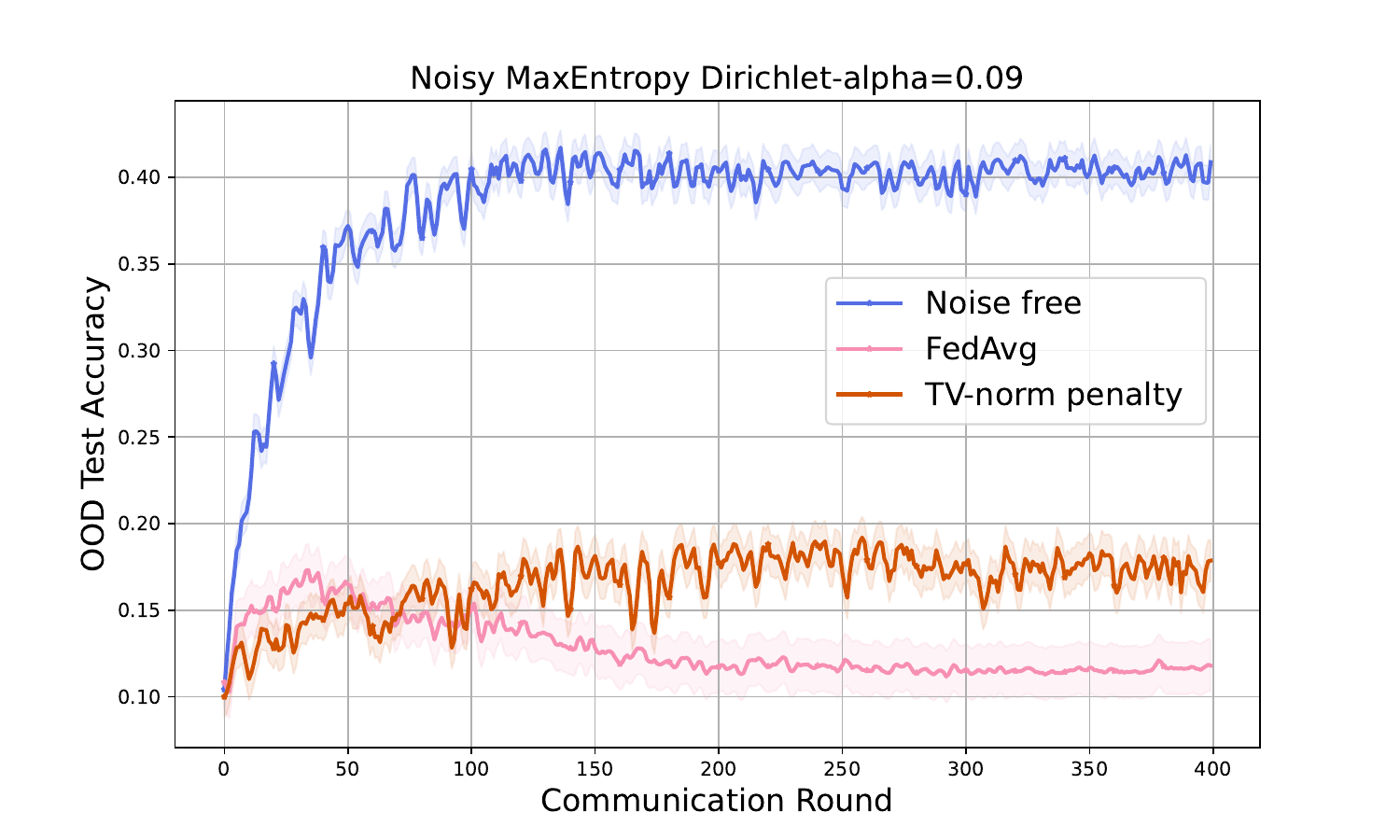} }
      \hspace{-0.2in}
    \subfloat[Convex-hull selection method ]{\includegraphics[width=0.38\linewidth]{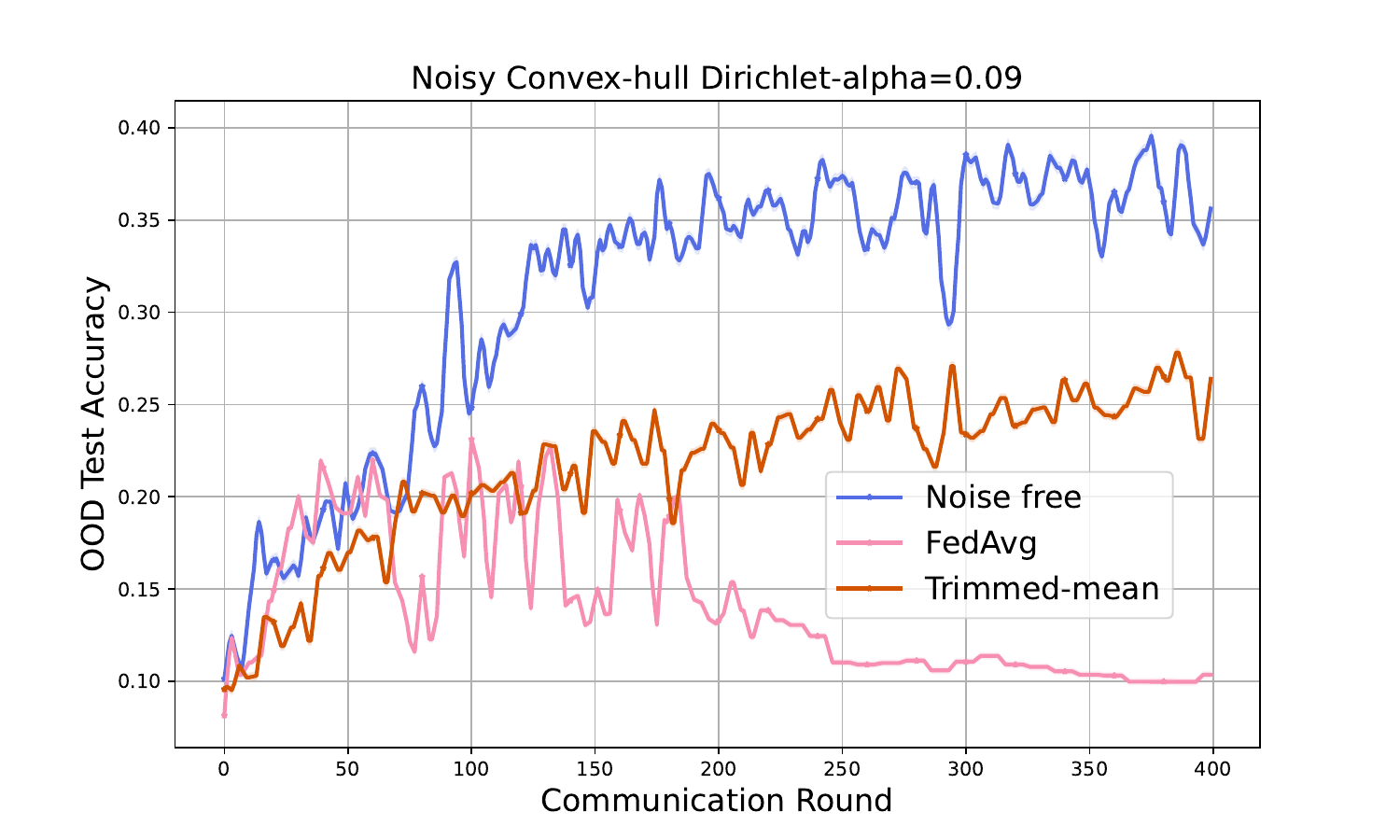} }
      \hspace{-0.2in}
       \subfloat[Minimax selection method ]{\includegraphics[width=0.38\linewidth]{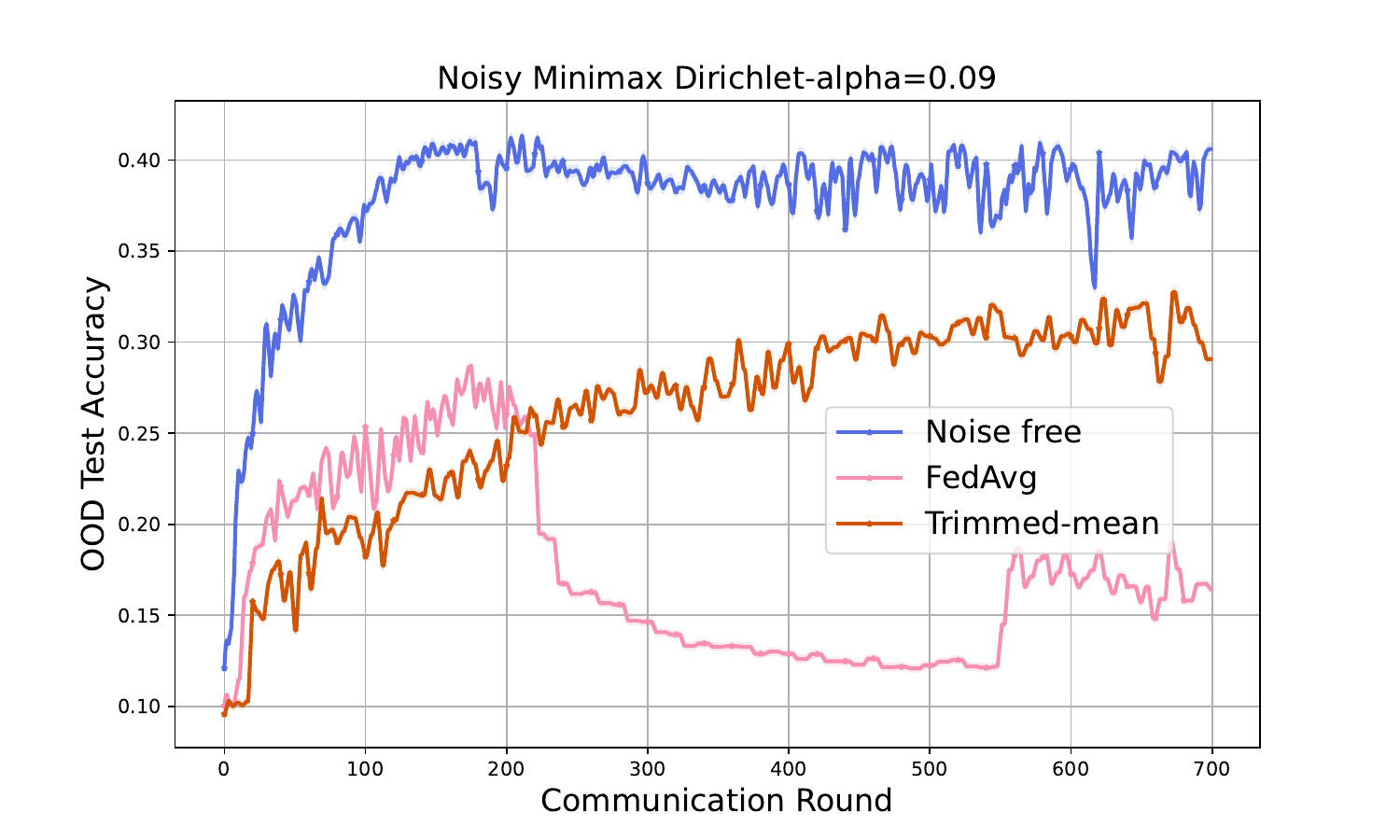} }
         \hspace{-0.55in}
	\caption{ The convergence analysis on OOD test accuracy of three proposed  methods combined with robust algorithms to address noisy label challenges on CIFAR-10. }
	\label{fig:noisy} 
            \vspace{-0.1in}
\end{figure}

\textbf{Additional experiments on verifying the capability of proposed methods combined with robust algorithms to address noisy label challenges:} In the following, we conduct some simple experiments to validate the effectiveness of proposed methods combined with robust algorithms to cope with noisy labels challenges. In this part, we assess the OOD performance of a CNN model on CIFAR-10. Similarly, we partition the dataset into $100$ clients using Dirichlet distribution splitting with $\alpha$ of $0.09$. We also select $40$ clients to join in FL, while the remaining ones act as non-participating clients. For the hyper-parameters of local training,  we utilize the batch size of $16$, the learning rate of $0.01$, and the local epoch of $3$.

Next, we introduce the noisy label setting considered in this part. Following~\cite{10337775}, we adopt the common Bernoulli distribution method to determine whether a local training set is affected by label noise. To inject label noise, we employ the symmetric noise method~\cite{song2022learning,DBLP:journals/corr/abs-2306-11650}, which manually introduces noise into datasets using a label transition matrix $T$, where $T_{i,j}:=p(\Tilde{y}=j|y=i)$. In particular, we utilize an equal probability distribution for flipping true labels into other labels, thereby employing a symmetric noise method.

We  integrate two prevalent Byzantine-robust algorithms into our proposed approaches to tackle the noisy label issues. Firstly, we combine two proposed client selection methods with Trimmed-mean~\cite{yin2018byzantine}, which removes outliers for each model parameter, and calculates the mean based on the remaining ones to update the global model. We combine the total variation (TV) norm penalty method~\cite{9054377}, which adds a TV norm-based proximal regularization term in each local objective,  with the proposed  empirical entropy-based weighting method. 

Figure~\ref{fig:noisy} illustrates the effectiveness of proposed approaches in combination with robust algorithms mentioned above in addressing label noise. Notice that the baseline denoted as "Noise free" represents performing proposed methods on datasets without noisy labels. The baseline called "FedAvg" indicates conducting presented methods using datasets affected by label noise without the use of robust algorithms to mitigate such degradation. In Figure~\ref{fig:noisy} (a), we observe that the proposed empirical entropy-based weighting method is significantly impacted by noisy labels. However, upon combining it with the TV-norm penalty method, the corresponding OOD test accuracy shows some improvement. In Figure~\ref{fig:noisy} (b) and (c), we find that two proposed client selection methods, when combined with the Trimmed-mean method, effectively tackle with noisy labels and achieve OOD test accuracy comparable to that of the noise-free setting in the later stages of FL. These  experiments  have demonstrated that proposed methods can effectively integrate robust algorithms to alleviate the impact of noisy labels on FL.

\textbf{Additional experiments on CIFAR-100 with ResNet-18:} To further assess the generalization performance of the proposed methods, we proceed with additional experiments, comparing our approaches with recent baselines using the CIFAR-100 dataset~\cite{krizhevsky2009learning} and a ResNet-18 model~\cite{he2016deep}. Initially, we partition the dataset into 100 clients using a Dirichlet distribution split with a parameter $\alpha$ set to 0.03. We then randomly choose 40 clients from this set to participate in FL. Each selected client conducts local training with a batch size of 16, runs for 3 local epochs, and uses a learning rate of 0.01. Our focus remains exclusively on the OOD performance of the global model.

\begin{figure} [ht]
\vspace{-0.1in}
\centering
  \hspace{-0.35in}
   \subfloat[Client selection method]
{\includegraphics[width=0.55\linewidth]{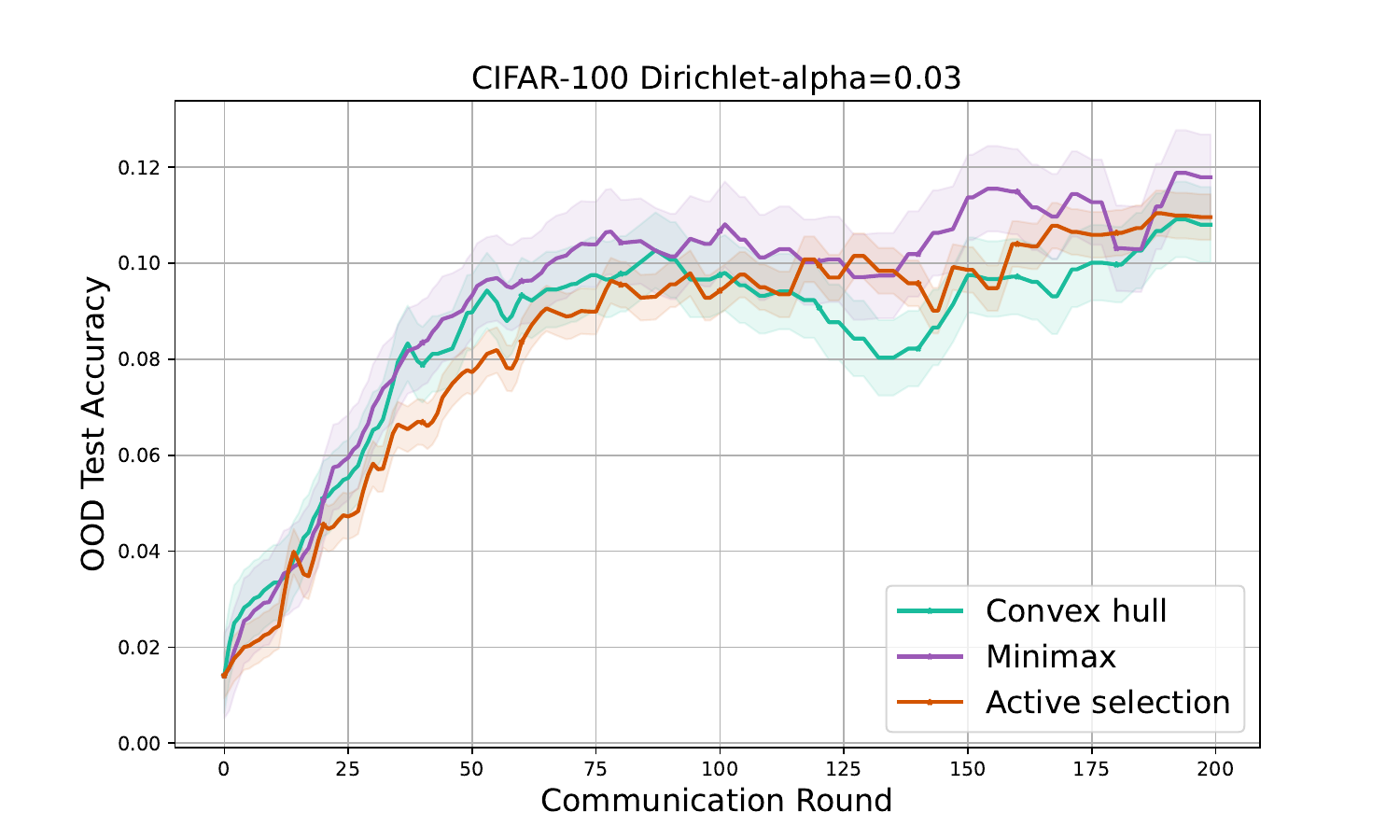} }
   \hspace{-0.2in}
   \subfloat[Weighting method]
   {\includegraphics[width=0.55\linewidth]{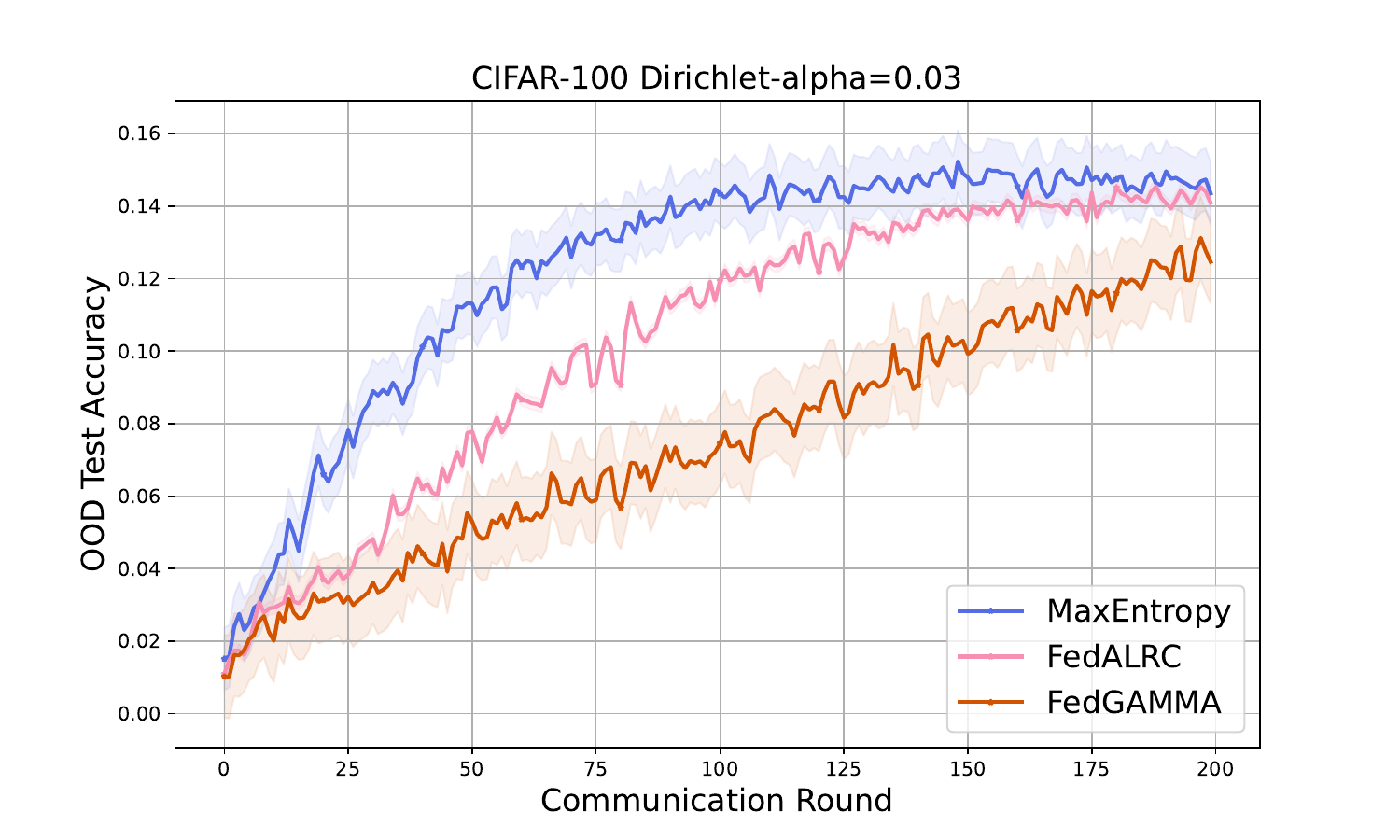} }
   \hspace{-0.5in}
	\caption{The convergence analysis on OOD test accuracy of empirical entropy weighting method and two proposed client selection methods compared with baselines on CIFAR-100.}
	\label{fig:baseline} 
    \vspace{-0.1in}

\end{figure}

We compare the proposed empirical entropy-based weighting method with two baselines, namely FedALRC~\cite{9927349} and FedGAMMA~\cite{10269141}. Additionally, we compare two presented client selection methods with active selection proposed in~\cite{10197242}. The experimental results are depicted in Figure~\ref{fig:baseline}.

In Figure~\ref{fig:baseline} (a), we employ a third-order moving average to present the results. We observe that the proposed convex hull construction-based client selection method performs worse than active selection in the later stage of FL. However, the proposed selection method based on minimax gradient similarity outperforms this baseline. This difference in performance can be attributed to the sensitivity of building convex hulls for complex models such as ResNes models, while using cosine similarity to measure gradient similarity demonstrates increased robustness in this situation. Figure~\ref{fig:baseline} (b) illustrates that the proposed empirical entropy-based weighting method outperforms other baseline methods. This superiority is due to the fact that these baseline methods only consider enhancing I.I.D. generalization performance and overlook potential non-participating clients. In contrast, our presented empirical entropy-based weighting method can improve the generalization capacity of models by identifying representative data sources.

\begin{figure} [ht]
\vspace{-0.1in}
\centering
  \hspace{-0.35in}
   \subfloat[Client selection method]{\includegraphics[width=0.55\linewidth]{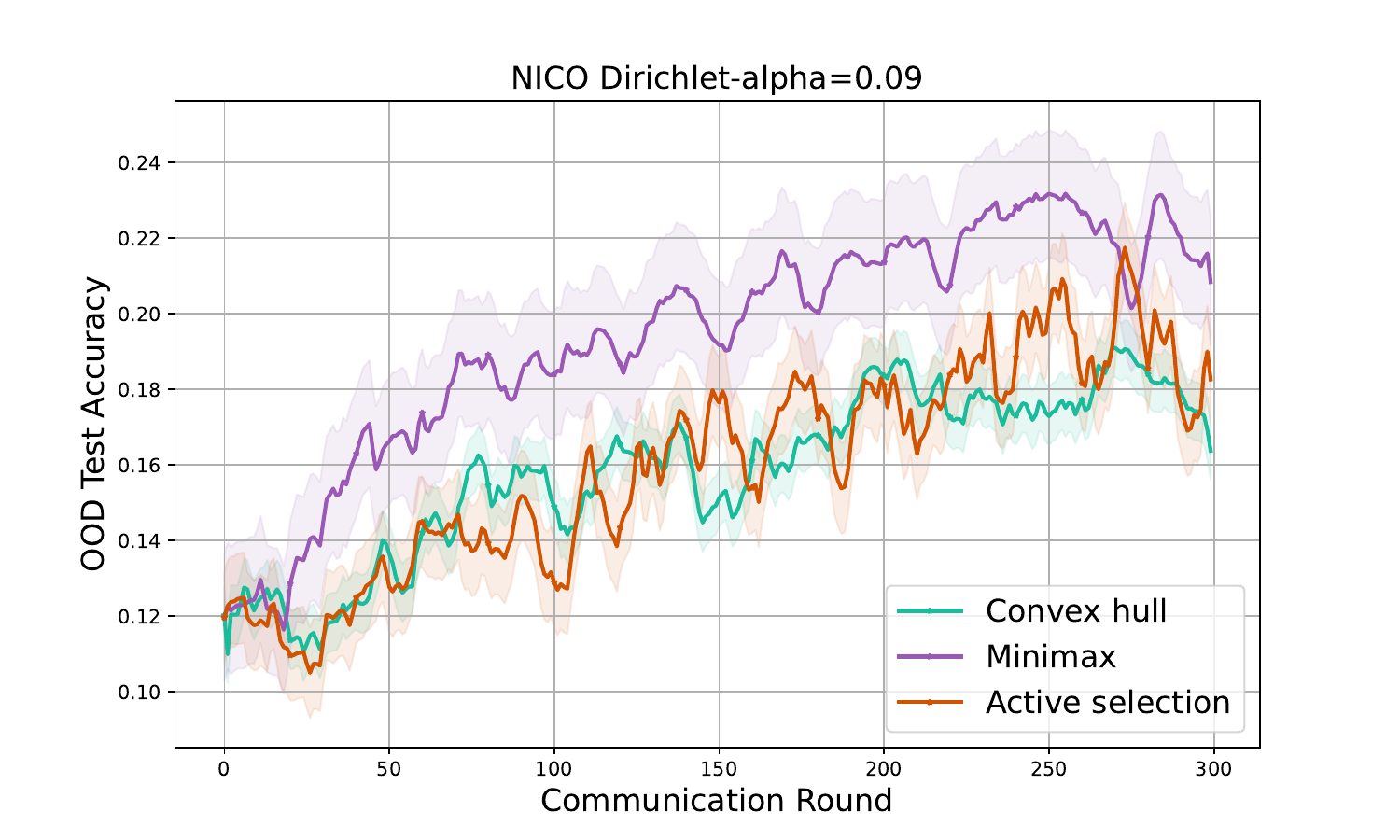} }
   \hspace{-0.2in}
   \subfloat[Weighting method]{\includegraphics[width=0.55\linewidth]{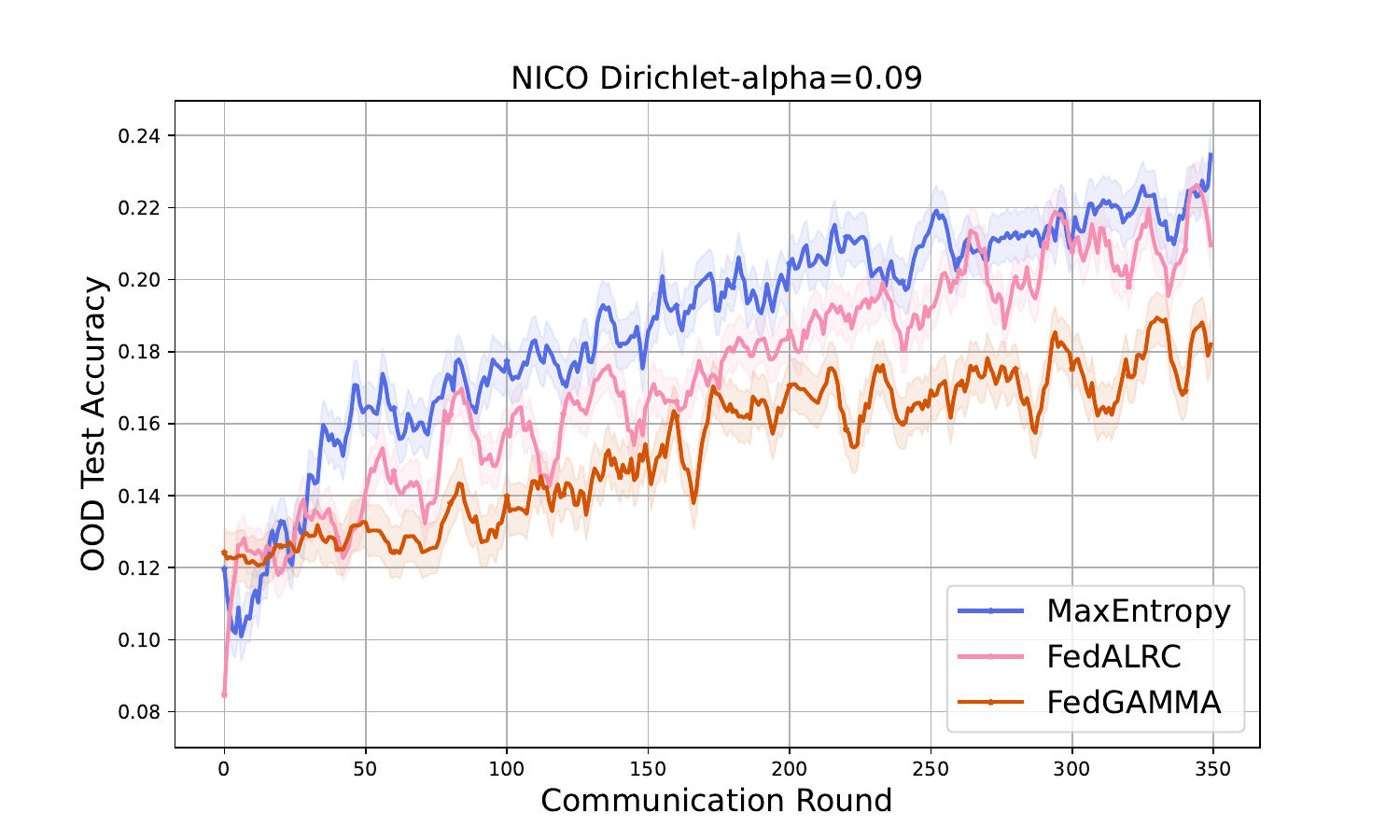} }
      \hspace{-0.5in}
	\caption{The convergence analysis on OOD test accuracy of empirical entropy weighting method and two proposed client selection methods compared with baselines on the NICO dataset.}
	\label{fig:nico} 
        \vspace{-0.1in}
\end{figure}

\textbf{Additional experiments on the NICO dataset with ResNet-18:} To further validate the generalizability of our proposed methods, we perform more experiments using a more realistic dataset. Specifically, we compared our approaches with previously mentioned baselines (FedALRC and FedGAMMA) using the Non-I.I.D. Image dataset with Contexts (NICO) dataset~\cite{he2021towards} with ResNet-18. The NICO dataset, which encompasses two superclasses of Animal and Vehicle, is tailored for Non-IID image classification. This dataset labels images with both primary concepts (e.g., dog) and the contexts (e.g., on grass) in which these visual concepts appear. The average number of images per context ranges from $83$ to $215$, while the average number of images per class approximates $1300$, akin to ImageNet. For this evaluation, we focus solely on the Animal superclass, which consists of $10$ classes. 

 We employ experimental settings akin to previous experiments conducted on CIFAR-100, setting $\alpha$ of the Dirichlet distribution to $0.09$. The experimental results are presented in Figure~\ref{fig:nico}. From Figure~\ref{fig:nico} (a), we find that the proposed Minimax client selection method still outperforms other methods. However, the method based on convex hull construction demonstrates similar or even inferior performance compared to active selection. This may be attributed to the bias introduced by using cosine similarity to construct the convex hull  when dealing with complex datasets like NICO. Figure~\ref{fig:nico} (b) illustrates that the proposed empirical entropy-based weighting method also surpasses other baselines due to its effective leveraging of the intrinsic category diversity within NICO, thereby enhancing the OOD performance of trained models.

\section{Conclusion}
This paper addresses the generalization issue in FL by exploring whether a global model trained by participating clients is capable of performing well for non-participating clients in the presence of heterogeneous data.  To capture the generalization gap in FL, we propose an information-theoretic generalization framework that takes into account both the information entropy of local distribution and the discrepancy between different distributions. Leveraging this framework, we are able to identify the generalization gap and further propose an empirical entropy-based weighting aggregation method,  as well as two gradient similarity-based client selection methods. These methods aim to enhance federated generalization for providing better service for non-participating clients through distribution diversification. Numerical results corroborate our theoretical findings, demonstrating that the proposed approaches can surpass baselines. In the future, our goal is to develop more comprehensive theories and approaches for tackling the challenge of balancing attention between rare examples and the overall distribution. It is imperative to ensure that the emphasis on rare cases of proposed methods does not compromise performance on common cases in practical scenarios.

\appendix

\section{Proof of theorems in federated generalization}\label{sec:proof}
In this section, we will present detailed proofs of theorems in the theoretical framework section.

\subsection{ Proof of Theorem~\ref{the:FedEntropy}}
\begin{proof}
    Let us recall the decomposed generalization gap in FL,
\begin{equation}
    \begin{aligned}
&\Big\vert\mathcal{L}_{Z^{\mathcal{I}_p}}(\hat{h})- \mathcal{L}_{S}(\hat{h})\Big\vert  
 \\
&\leq\underbrace{\Big\vert\mathcal{L}_{Z^{\mathcal{I}_p}}(\hat{h})-\mathcal{L}_{{Z_{\mathcal{I}_p}}}(\hat{h})\Big\vert}_{\mbox{distributed learning gap}}+\underbrace{\Big\vert\mathcal{L}_{{Z_{\mathcal{I}_p}}}(\hat{h})-\mathcal{L}_{S}(\hat{h})\Big\vert}_{\mbox{semi-generalization gap}}.
    \end{aligned}
\end{equation}

We regard the two terms in this decomposed generalization gap as two different lemmas and introduce the detailed proofs of these lemmas in the following.

\begin{lemma}
     [Distributed learning gap]\label{pargap}
     \begin{equation}
    \begin{aligned}
\Big\vert\mathcal{L}_{Z^{\mathcal{I}_p}}(\hat{h})-\mathcal{L}_{{Z_{\mathcal{I}_p}}}(\hat{h})\Big\vert&\leq  bH(Z^{\mathcal{I}_p})-b\sum_{i \in \mathcal{I}_p}\alpha_i H(Z_i).
    \end{aligned}
\end{equation}

\begin{proof}
   In the following, we show that the "distributed learning gap" term $Gen(\mathcal{I}_p;\hat{h}):=\vert \mathbb{E}_{\mathcal{Z}^{\mathcal{I}_p}}[\sum_{i \in \mathcal{I}_p}\alpha_i\ell(\hat{h},z_i)\log \frac{1}{P_{Z^{\mathcal{I}_p}}}]- \sum_{i \in \mathcal{I}_p} \alpha_i\mathbb{E}[ \ell(h,Z_i)\log(\frac{1}{P(Z_i)}) ]\vert$ can be bounded as,
    \begin{equation}
        \begin{aligned}
   & Gen(\mathcal{I}_p;\hat{h})=\Big\vert \mathbb{E}_{\mathcal{Z}^{\mathcal{I}_p}}\big[\sum_{i \in \mathcal{I}_p}\alpha_i\ell(\hat{h},z_i)\log \frac{1}{P_{Z^{\mathcal{I}_p}}}\big]\\
    &\quad- \sum_{i \in \mathcal{I}_p} \alpha_i\mathbb{E}\big[ \ell(h,Z_i)\log(\frac{1}{P(Z_i)}) \big]\Big\vert\\
    &=\Big\vert \sum_{\mathcal{Z}^{\mathcal{I}_p  }}P_{Z^{\mathcal{I}_p}}\sum_{i \in \mathcal{I}_p}\alpha_i\ell(\hat{h},z_i)\log \frac{1}{P_{Z^{\mathcal{I}_p}}}\\
   &\quad- \sum_{i \in \mathcal{I}_p} \alpha_i\sum_{z_i\in\mathcal{Z}} P_{Z_i}\ell(\hat{h},Z_i)\log\frac{1}{P_{Z_i}} \Big\vert\\
           &\leq \sum_{i \in \mathcal{I}_p}\alpha_i \big\vert\sum_{\mathcal{Z}^{\mathcal{I}_p  }}P_{Z^{\mathcal{I}_p}}\ell(\hat{h},z_i)\log \frac{1}{P_{Z^{\mathcal{I}_p}}}\\
           &\quad-\sum_{z_i \in \mathcal{Z}}P_{Z_i}\ell(\hat{h},z_i)\log \frac{1}{P_{Z_i}} \big\vert\\
              &=\sum_{i \in \mathcal{I}_p}\alpha_i \big\vert\sum_{\mathcal{Z}^{\mathcal{I}_p  }}P_{Z^{\mathcal{I}_p}}\ell(\hat{h},z_i)\log \frac{1}{P_{Z^{\mathcal{I}_p}}} \\
              &\quad-\sum_{z_i \in \mathcal{Z}}\sum_{\mathcal{Z}^{\mathcal{I}_p\setminus i  }}P_{Z_i}P_{Z^{\mathcal{I}_p \setminus i}}\ell(\hat{h},z_i)\log \frac{1}{P_{Z_i}} \big\vert.\\
        \end{aligned}
    \end{equation}

   Based on Assumption~\ref{assumption2}, we can  have,
     \begin{equation}
        \begin{aligned}
&Gen({I}_p;\hat{h}) \\    
              &\leq \sum_{i \in \mathcal{I}_p}\alpha_i \sum_{\mathcal{Z}^{\mathcal{I}_p  }}P_{Z^{\mathcal{I}_p}}\Big\vert\ell(\hat{h},z_i)\log \frac{1}{P_{Z^{\mathcal{I}_p}}} -\ell(\hat{h},z_i)\log \frac{1}{P_{Z_i}}\Big\vert\\
                 &\leq b\sum_{i \in \mathcal{I}_p}\alpha_i \sum_{\mathcal{Z}^{\mathcal{I}_p  }}P_{Z^{\mathcal{I}_p}}\Big\vert\log \frac{1}{P_{Z^{\mathcal{I}_p}}} -\log\frac{1}{P_{Z_i}}\Big\vert\\
                  & \stackrel{\text{(a)}}{=}b\sum_{i \in \mathcal{I}_p}\alpha_i \sum_{\mathcal{Z}^{\mathcal{I}_p  }}P_{Z^{\mathcal{I}_p}}\Big[\log \frac{1}{P_{Z^{\mathcal{I}_p}}} -\log\frac{1}{P_{Z_i}}\Big]\\
            &=b\sum_{i \in \mathcal{I}_p}\alpha_i \Big[\sum_{\mathcal{Z}^{\mathcal{I}_p  }}P_{Z^{\mathcal{I}_p}}\log \frac{1}{P_{Z^{\mathcal{I}_p}}} -\sum_{z_i \in \mathcal{Z}}P_{Z_i}\log \frac{1}{P_{Z_i}}\Big] \\
            &=bH(Z^{\mathcal{I}_p})-b\sum_{i \in \mathcal{I}_p}\alpha_i H(Z_i),
        \end{aligned}
    \end{equation}
where $(a)$ follows from $P_{Z^{\mathcal{I}_p}}=P_{Z_i}P_{Z^{\mathcal{I}_p\setminus i } }$ and $\log \frac{1}{P_{Z^{\mathcal{I}_p}}} -\log\frac{1}{P_{Z_i}}=\log \frac{1}{P_{Z^{\mathcal{I}_p \setminus i}}}+\log\frac{1}{P_{Z_i}}-\log\frac{1}{P_{Z_i}}\geq 0$.

\end{proof}
\end{lemma}

\begin{lemma}
    [Semi-generalization gap]\label{serisk} Let $\mathcal{G}$ be a family of functions related to hypothesis space $\mathcal{H}: z\mapsto \ell(h,z):h \in \mathcal{H}$ with VC dimension $VC(\mathcal{G})$. Distributed training sets $\{ S_i\}_{i=1}^{N}=\{ \{s_j^i\}_{j=1} ^{n_i}\}_{i=1}^{N}$ are constructed by i.i.d. realizations sampled from different data sources $\{Z_i\}_{i=1}^{N}$. If the loss function $\ell$ is bounded by $b$, for any $\delta \geq 0$, it follows that with probability at least $1-\delta$,
\begin{equation}
    \begin{aligned}
    \mathcal{L}_{{Z_{\mathcal{I}_p}}}(\hat{h})-\mathcal{L}_{S}(\hat{h}) &\leq \mathcal{E}+cb\sqrt{\frac{VC(\mathcal{G})}{\sum_{i=1}^{\vert\mathcal{I}_p \vert}n_i}} +b\sqrt{\frac{\log(1/\delta)}{2\sum_{i=1}^{\vert\mathcal{I}_p \vert}n_i}},
    \end{aligned}
\end{equation}
where $c$ is a constant, and $\mathcal{E}= \sum_{z \in \mathcal{Z}}\ell(h',z)$, where $ h'=\sup_{h \in \mathcal{H}} \vert\mathcal{L}_{{Z_{\mathcal{I}_p}}}(h)-\sum_{i\in \mathcal{I}_p}\alpha_i \mathbb{E}[\ell(h,z_i)]\vert$. The term $\mathcal{E}$ represents the gap between the self-information weighted expected risk and the vanilla expected risk without considering the self-information of outcomes.
\end{lemma}
Before starting the detailed proof of Lemma~\ref{serisk}, we introduce a theoretical result about the generalization bound for participating clients in IID federated learning in \cite{hu2023generalization} into our paper as a lemma.
    \begin{lemma}
        [Generalization bound for participating clients in IID FL]\label{IIDgen} \cite{hu2023generalization} Let $\mathcal{G}$ be a family of functions related to hypothesis space $\mathcal{H}: z\mapsto \ell(h,z):h \in \mathcal{H}$ with VC dimension $VC(\mathcal{G})$. Distributed training sets $\{ S_i\}_{i=1}^{N}=\{ \{s_j^i\}_{j=1} ^{n_i}\}_{i=1}^{N}$ are constructed by i.i.d. realizations sampled from different data sources $\{Z_i\}_{i=1}^{N}$. If the loss function $\ell$ is bounded by $b$, for any $\delta \geq 0$, it follows that with probability at least $1-\delta$,
        \begin{equation}
\begin{aligned}
  &  \sup_{h \in \mathcal{H}} \Big\vert \sum_{i=1}^{N}\alpha_i \mathbb{E}(\ell(h,Z_i))-\sum_{i=1}^{N}\frac{\alpha_i}{n_i}\sum_{j=1}^{n_i}\ell(h,s_j^i)\Big\vert \\
  &\leq cb\sqrt{\frac{VC(\mathcal{G})}{\sum_{i=1}^N n_i}}+b\sqrt{\frac{ln(1/\delta)}{2\sum_{i=1}^N n_i}},
\end{aligned}
\end{equation}
where $c$ is a constant.
    \end{lemma}

Based on this lemma, we can complete the proof of Lemma~\ref{serisk}. We then start the formal proof of the Lemma~\ref{serisk}
\begin{proof}
\begin{equation}
    \begin{aligned}
|\mathcal{L}_{{Z_{\mathcal{I}_p}}}(\hat{h})-\mathcal{L}_{S}(\hat{h})|
&=|\mathcal{L}_{{Z_{\mathcal{I}_p}}}(\hat{h})-\sum_{i\in \mathcal{I}_p}\alpha_i \mathbb{E}[\ell(\hat{h},z_i)]\\
&\quad+\sum_{i\in \mathcal{I}_p}\alpha_i \mathbb{E}[\ell(\hat{h},z_i)]-\mathcal{L}_{S}(\hat{h})|\\
&\leq \sup_{h \in \mathcal{H}} \Big \vert\mathcal{L}_{{Z_{\mathcal{I}_p}}}(h)-\sum_{i\in \mathcal{I}_p}\alpha_i \mathbb{E}[\ell(h,z_i)]\Big \vert\\
&\quad+  \sup_{h \in \mathcal{H}} \Big \vert\sum_{i\in \mathcal{I}_p}\alpha_i \mathbb{E}[\ell(h,z_i)]-\mathcal{L}_{S}(h)\Big \vert.\\
    \end{aligned}
\end{equation}

Rooted on the Lemma~\ref{IIDgen}, we can immediately have the below result with a probability of at least $1-\delta$,
\begin{equation}
    \begin{aligned}
|\mathcal{L}_{{Z_{\mathcal{I}_p}}}(\hat{h})-\mathcal{L}_{S}(\hat{h})|&\leq \sup_{h \in \mathcal{H}} \Big \vert\mathcal{L}_{{Z_{\mathcal{I}_p}}}(h)-\sum_{i\in \mathcal{I}_p}\alpha_i \mathbb{E}[\ell(h,z_i)]\Big \vert\\
&\quad+cb\sqrt{\frac{VC(\mathcal{G})}{\sum_{i=1}^{\vert\mathcal{I}_p \vert}n_i}} +b\sqrt{\frac{\log(1/\delta)}{2\sum_{i=1}^{\vert\mathcal{I}_p \vert}n_i}}.\\
    \end{aligned}
\end{equation}

For simplicity, we denote the term $cb\sqrt{\frac{VC(\mathcal{G})}{\sum_{i=1}^{\vert\mathcal{I}_p \vert}n_i}} +b\sqrt{\frac{\log(1/\delta)}{2\sum_{i=1}^{\vert\mathcal{I}_p \vert}n_i}}$ as $\mathcal{N}(\mathcal{G},\mathcal{I}_p)$ temporarily.

 Assume that $h' \in \mathcal{H}$ satisfying $\sup_{h \in \mathcal{H}}  \vert\mathcal{L}_{{Z_{\mathcal{I}_p}}}(h)-\sum_{i\in \mathcal{I}_p}\alpha_i \mathbb{E}[\ell(h,z_i)] \vert=\vert\mathcal{L}_{{Z_{\mathcal{I}_p}}}(h')-\sum_{i\in \mathcal{I}_p}\alpha_i \mathbb{E}[\ell(h',z_i)]\vert$, we have
\begin{equation}
    \begin{aligned}
&\mathcal{L}_{{Z_{\mathcal{I}_p}}}(\hat{h})-\mathcal{L}_{S}(\hat{h}) \\
& \leq \Big\vert\mathcal{L}_{{Z_{\mathcal{I}_p}}}(h')-\sum_{i\in \mathcal{I}_p}\alpha_i \mathbb{E}[\ell(h',z_i)]\Big\vert +\mathcal{N}(\mathcal{G},\mathcal{I}_p)\\
   & =\Big\vert \sum_{i\in \mathcal{I}_p} \alpha_i \sum_{z_i \in \mathcal{Z}}P_{Z_i} \ell(h',z_i) \\
   &\quad-\sum_{i\in \mathcal{I}_p} \alpha_i \sum_{z_i \in \mathcal{Z}}P_{Z_i} \ell(h',z_i) \log \frac{1}{P_{Z_i}}\Big\vert+ \mathcal{N}(\mathcal{G},\mathcal{I}_p)\\
   &\leq\sum_{i\in \mathcal{I}_p} \alpha_i \sum_{z_i \in \mathcal{Z}}P_{Z_i}\ell(h',z_i)  \big\vert \log P_{Z_i} +1\big\vert+\mathcal{N}(\mathcal{G},\mathcal{I}_p)\\
    & \stackrel{\text{(a)}}{\leq}\sum_{i\in \mathcal{I}_p} \alpha_i \sum_{z_i \in \mathcal{Z}}P_{Z_i}\ell(h',z_i)  \big( \log \frac{1}{P_{Z_i}}+1\big)+\mathcal{N}(\mathcal{G},\mathcal{I}_p)\\
   &\stackrel{\text{(b)}}{\leq} \sum_{i\in \mathcal{I}_p} \alpha_i \sum_{z_i \in \mathcal{Z}}\ell(h',z_i) + \mathcal{N}(\mathcal{G},\mathcal{I}_p)\\
    &= \sum_{z_i \in \mathcal{Z}}\ell(h',z_i)+\mathcal{N}(\mathcal{G},\mathcal{I}_p),\\
    \end{aligned}
\end{equation}
where $(a)$ holds since $0<P_{Z_i}(z_i)< 1$ and $(b)$ follows from the fundamental inequality $\log x\leq x-1, \forall x>0$.

%If we assume that $h' \in \mathcal{H}$ satisfying $\sup_{h \in \mathcal{H}} \Big \vert\mathcal{L}_{{Z_{\mathcal{I}_p}}}(h)-\sum_{i\in \mathcal{I}_p}\alpha_i \mathbb{E}[\ell(h,z_i)]\Big \vert=\mathcal{L}_{{Z_{\mathcal{I}_p}}}(h')-\sum_{i\in \mathcal{I}_p}\alpha_i \mathbb{E}[\ell(h',z_i)]$, we have $\mathcal{L}_{{Z_{\mathcal{I}_p}}}(\hat{h})-\mathcal{L}_{Z_{\mathcal{I}_p}}(\hat{h}^*) \leq b\sum_{i\in \mathcal{I}_p}  \sum_{z_i \in \mathcal{Z}}P_{Z_i}^2 +\mathcal{N}(\mathcal{G},\mathcal{I}_p)$.

To sum up, we eventually have
\begin{equation}
    \begin{aligned}
\mathcal{L}_{{Z_{\mathcal{I}_p}}}(\hat{h})-\mathcal{L}_{Z_{\mathcal{I}_p}}(\hat{h}^*)
    &\leq \mathcal{E}+cb\sqrt{\frac{VC(\mathcal{G})}{\sum_{i=1}^N n_i}}+b\sqrt{\frac{ln(1/\delta)}{2\sum_{i=1}^N n_i}},\\
    \end{aligned}
\end{equation}
where the term $\mathcal{E}$ represents  $\sum_{z_i \in \mathcal{Z}}\ell(h',z_i)$.
\end{proof}

On the basis of the proven three lemmas introduced above, we can find that Theorem~\ref{the:FedEntropy} is proven immediately.
\end{proof}

\subsection{Proof of Corollary~\ref{entropyrate}}
 \begin{proof}
   Following the proof of Lemma~\ref{pargap}, we can have,
        \begin{equation}
        \begin{aligned}
     &\frac{1}{M}  \Big
            \vert\mathbb{E}\big[\sum_{i \in \mathcal{I}_p}\frac{\ell(\hat{h},Z_i)}{M}\log \frac{1}{P_{Z^{\mathcal{I}_p}}}\big]-\frac{1}{M}\sum_{i \in \mathcal{I}_p}\mathbb{E}\big[\ell(\hat{h},Z_i)\log \frac{1}{P_{Z_i}} \big] \Big
            \vert\\
            &\leq \frac{b}{M}[H(Z^{\mathcal{I}_p})-\frac{1}{M}\sum_{i \in \mathcal{I}_p}H(Z_i)]\\
            &\leq\frac{ bH(Z^{\mathcal{I}})}{M}-\frac{ bH(Z^{\mathcal{I}_p})}{M^2}
        \end{aligned}
    \end{equation}
    Since the entropy rate $H(\mathcal{Z})=\lim_{M \rightarrow \infty}\frac{H(Z^{\mathcal{I}_p})}{M}$ exists,  we  have,
    \begin{equation}
        \begin{aligned}
              \lim_{M \rightarrow \infty}\frac{\big\vert\mathcal{L}_{Z^{\mathcal{I}_p}}(\hat{h})-\mathcal{L}_{{Z_{\mathcal{I}_p}}}(\hat{h})\big\vert}{M}  & \leq                 \lim_{M \rightarrow \infty}\{\frac{ bH(Z^{\mathcal{I}_p})}{M}-\frac{ bH(Z^{\mathcal{I}_p})}{M^2}\}\\
            &=bH(\mathcal{Z}).
        \end{aligned}
    \end{equation}
   \end{proof}

\subsection{Proof of Theorem~\ref{the:Fedcorre} and Theorem~\ref{the:FedCross}}
\begin{proof}
    Following the  proof of Theorem~\ref{the:FedEntropy}, we can also decompose and amplify the gap $\vert\mathcal{L}_{Z^{\mathcal{I}_p}}(\hat{h}_t)- \mathcal{L}_{S_t} (\hat{h}_t)\vert $ as,
    \begin{equation}
    \begin{aligned}
              &  \vert\mathcal{L}_{Z^{\mathcal{I}_p}}(\hat{h}_t)- \mathcal{L}_{S_t} (\hat{h}_t)\vert  \\
              &\leq \underbrace{\Big\vert\mathcal{L}_{Z^{\mathcal{I}_p}}(\hat{h}_t)-\mathcal{L}_{{Z_{\mathcal{I}_t}}}(\hat{h}_t)\Big\vert}_{\mbox{selection gap}}+\underbrace{\Big\vert\mathcal{L}_{{Z_{\mathcal{I}_t}}}(\hat{h}_t)-\mathcal{L}_{S_t}(\hat{h}_t)\Big\vert}_{\mbox{semi-generalization gap}}.
    \end{aligned}
    \end{equation}
    
We can use the theoretical results derived in Theorem~\ref{the:FedEntropy} to bound the semi-generalization gap defined above directly. Therefore, we mainly focus on the selection gap defined above. The selection gap can be decomposed and amplified as,
      \begin{equation}
      \begin{aligned}
  & \Big\vert\mathcal{L}_{Z^{\mathcal{I}_p}}(\hat{h}_t)-\mathcal{L}_{{Z_{\mathcal{I}_t}}}(\hat{h}_t)\Big\vert\\
  &\leq \underbrace{\Big\vert\mathcal{L}_{Z^{\mathcal{I}_p}}(\hat{h}_t)-\mathcal{L}_{Z^{\mathcal{I}_t}}(\hat{h}_t)\Big\vert}_{\mbox{distribution gap}}+\underbrace{\Big\vert\mathcal{L}_{Z^{\mathcal{I}_t}}(\hat{h}_t)-\mathcal{L}_{{Z_{\mathcal{I}_t}}}(\hat{h}_t)\Big\vert}_{\mbox{distributed learning gap}}
      \end{aligned}
  \end{equation}
\begin{comment}
      \begin{equation}
      \begin{aligned}
   \Big\vert\mathcal{L}_{Z^{\mathcal{I}_p}}(\hat{h}_t)-\mathcal{L}_{{Z_{\mathcal{I}_t}}}(\hat{h}_t)\Big\vert&=\Big\vert\sum_{\mathcal{Z}^{\mathcal{I}_p}}\frac{P_{Z^{\mathcal{I}_p}}}{M}\sum_{i \in \mathcal{I}}\ell(\hat{h}_t,z_i)\log \frac{1}{P_{Z^{\mathcal{I}}}}\\
&\quad-\sum_{i \in \mathcal{I}_t}\sum_{z_i \in \mathcal{Z}}\frac{P_{Z_i}}{K}\ell(\hat{h}_t,z_i)\log\frac{1}{P_{Z_i}}\Big\vert\\
      \end{aligned}
  \end{equation}
\end{comment}
  
Similar to the proof of Theorem~\ref{the:FedEntropy}, we can easily find that the "distributed learning gap" term $\vert\mathcal{L}_{Z^{\mathcal{I}_t}}(\hat{h}_t)-\mathcal{L}_{{Z_{\mathcal{I}_t}}}(\hat{h}_t)\vert$ can be bounded as,
\begin{equation}
  \begin{aligned}
        &\vert\mathcal{L}_{Z^{\mathcal{I}_t}}(\hat{h}_t)-\mathcal{L}_{{Z_{\mathcal{I}_t}}}(\hat{h}_t)\vert \leq bH(Z^{\mathcal{I}_t})-\frac{b}{K}\sum_{i \in \mathcal{I}_t} H(Z_i).
  \end{aligned}
\end{equation}

Then we turn to bound the "distribution gap" term $\vert\mathcal{L}_{Z^{\mathcal{I}_p}}(\hat{h}_t)-\mathcal{L}_{Z^{\mathcal{I}_t}}(\hat{h}_t)\vert$.
  \begin{equation}
      \begin{aligned}
        & \vert\mathcal{L}_{Z^{\mathcal{I}_p}}(\hat{h}_t)-\mathcal{L}_{Z^{\mathcal{I}_t}}(\hat{h}_t)\vert \\
        &= \Big\vert \sum_{\mathcal{Z}^{\mathcal{I}_p  }}P_{Z^{\mathcal{I}_p}}\frac{1}{M}\sum_{i \in \mathcal{I}_p}\ell(\hat{h},z_i)\log\Big( \frac{1}{P_{Z^{\mathcal{I}_p}}} \Big)\\
        &\quad- \sum_{\mathcal{Z}^{\mathcal{I}_t  }}P_{Z^{\mathcal{I}_t}}\sum_{i \in \mathcal{I}_t}\frac{1}{K}\ell(\hat{h},z_i)\log\Big( \frac{1}{P_{Z^{\mathcal{I}_t}}} \Big)\Big\vert\\
        &=   \Big\vert \sum_{\mathcal{Z}^{\mathcal{I}_p  }}P_{Z^{\mathcal{I}_p}}\frac{1}{M}\sum_{i \in \mathcal{I}_t}\ell(\hat{h},z_i)\log\Big( \frac{1}{P_{Z^{\mathcal{I}_p}}} \Big)\\
        &\quad+\sum_{\mathcal{Z}^{\mathcal{I}_p  }}P_{Z^{\mathcal{I}_p}}\frac{1}{M}\sum_{i \in \mathcal{I}_p\setminus \mathcal{I}_t}\ell(\hat{h},z_i)\log\Big( \frac{1}{P_{Z^{\mathcal{I}_p}}} \Big)\\
        &\quad- \sum_{\mathcal{Z}^{\mathcal{I}_t  }}P_{Z^{\mathcal{I}_t}}\sum_{i \in \mathcal{I}_t}\frac{1}{K}\ell(\hat{h},z_i)\log\Big( \frac{1}{P_{Z^{\mathcal{I}_t}}} \Big)\Big\vert\\
        &\leq \Big\vert \sum_{\mathcal{Z}^{\mathcal{I}_p  }}P_{Z^{\mathcal{I}_p}}\frac{1}{M}\sum_{i \in  \mathcal{I}_p\setminus \mathcal{I}_t}\ell(\hat{h},z_i)\log \frac{1}{P_{Z^{\mathcal{I}_p}}} \Big\vert\\
&\quad+\Big\vert\sum_{\mathcal{Z}^{\mathcal{I}_p}}P_{Z^{\mathcal{I}_p}}\frac{1}{M}\sum_{i \in \mathcal{I}_t}\ell(\hat{h}_t,z_i)\log \frac{1}{P_{Z^{\mathcal{I}_p}}}\\
         &\quad-\sum_{\mathcal{Z}^{\mathcal{I}_t}}\sum_{\mathcal{Z}^{\mathcal{I}_p\setminus\mathcal{I}_t}}P_{Z^{\mathcal{I}_t}}P_{Z^{\mathcal{I}_p\setminus\mathcal{I}_t}}\frac{1}{K}\sum_{i \in \mathcal{I}_t}\ell(\hat{h}_t,z_i)\log \frac{1}{P_{Z^{\mathcal{I}_t}}}\Big\vert\\
        &\leq \frac{b(M-K)}{M}H(Z^{\mathcal{I}_p})\\
        &\quad+\Big\vert\sum_{\mathcal{Z}^{\mathcal{I}_p}}P_{Z^{\mathcal{I}_p}}\frac{1}{M}\sum_{i \in \mathcal{I}_t}\ell(\hat{h}_t,z_i)\log \frac{1}{P_{Z^{\mathcal{I}_p}}}\\
         &\quad-\sum_{\mathcal{Z}^{\mathcal{I}_t}}\sum_{\mathcal{Z}^{\mathcal{I}_p\setminus\mathcal{I}_t}}P_{Z^{\mathcal{I}_t}}P_{Z^{\mathcal{I}_p\setminus\mathcal{I}_t}}\frac{1}{K}\sum_{i \in \mathcal{I}_t}\ell(\hat{h}_t,z_i)\log \frac{1}{P_{Z^{\mathcal{I}_t}}}\Big\vert.\\
      \end{aligned}
  \end{equation}

 Based on Assumption~\ref{assumption2}, we can further have,
 \begin{equation}
     \begin{aligned}
      & \vert\mathcal{L}_{Z^{\mathcal{I}_p}}(\hat{h}_t)-\mathcal{L}_{Z^{\mathcal{I}_t}}(\hat{h}_t)\vert \\
         &\leq \frac{b(M-K)}{N}H(Z^{\mathcal{I}_p})+\sum_{\mathcal{Z}^{\mathcal{I}_p}}P_{Z^{\mathcal{I}_p}}\Big\vert \frac{1}{M}\sum_{i \in \mathcal{I}_t}\ell(\hat{h}_t,z_i)\log \frac{1}{P_{Z^{\mathcal{I}_p}}}\\
         &\quad-\frac{1}{K}\sum_{i \in \mathcal{I}_t}\ell(\hat{h}_t,z_i)\log \frac{1}{P_{Z^{\mathcal{I}_t}}}\Big\vert\\
        &=\frac{b(M-K)}{M}H(Z^{\mathcal{I}_p})\\
        &\quad+\sum_{\mathcal{Z}^{\mathcal{I}_p}}P_{Z^{\mathcal{I}_p}}\Big\vert [\frac{1}{M}-\frac{1}{K}]\sum_{i \in \mathcal{I}_t}\ell(\hat{h}_t,z_i)\log \frac{1}{P_{Z^{\mathcal{I}_p}}}\Big\vert\\
&\quad+\sum_{\mathcal{Z}^{\mathcal{I}_p}}P_{Z^{\mathcal{I}_p}}\Big\vert\frac{1}{K}\sum_{i \in \mathcal{I}_t}\ell(\hat{h}_t,z_i)[\log \frac{1}{P_{Z^{\mathcal{I}_p}}}-\log \frac{1}{P_{Z^{\mathcal{I}_t}}}]\Big\vert\\
              &\stackrel{\text{(a)}}{=}  \frac{b(M-K)}{K}H(Z^{\mathcal{I}_p})\\
              &\quad+b\sum_{\mathcal{Z}^{\mathcal{I}_p}}P_{Z^{\mathcal{I}_p}}\sum_{i \in \mathcal{I}_t} \frac{M-K}{MK}\log \frac{1}{P_{Z^{\mathcal{I}_p}}}\\
        &\quad+b\sum_{\mathcal{Z}^{\mathcal{I}_p}}P_{Z^{\mathcal{I}_p}}\frac{1}{K}\sum_{i \in \mathcal{I}_t}\big[\log \frac{1}{P_{Z^{\mathcal{I}_p}}}-\log \frac{1}{P_{Z^{\mathcal{I}_t}}}\big]\\
                     &= \frac{b(3M-2K)}{M}H(Z^{\mathcal{I}_p})-bH(Z^{\mathcal{I}_t}),
     \end{aligned}
 \end{equation}
where $(a)$ follows the fact that $P_{Z^{\mathcal{I}_p}}=P_{Z^{\mathcal{I}_t}}P_{Z^{\mathcal{I}_p\setminus \mathcal{I}_t } }$ and $\log \frac{1}{P_{Z^{\mathcal{I}_p}}} -\log \frac{1}{P_{Z^{\mathcal{I}_t}}}=\log \frac{1}{P_{Z^{\mathcal{I}_p\setminus \mathcal{I}_t } }}+\log \frac{1}{P_{Z^{\mathcal{I}_t}}}-\log \frac{1}{P_{Z^{\mathcal{I}_t}}}\geq 0$.
  
\begin{comment}
      According to the above analysis and Lemma~\ref{pargap}, we can bound the  term $T_2=\Big\vert\sum_{\mathcal{Z}^{\mathcal{I}_p}}\frac{P_{Z^{\mathcal{I}_p}}}{M}\sum_{i \in \mathcal{I}_p}\ell(\hat{h}_t,z_i)\log \frac{1}{P_{Z^{\mathcal{I}_p}}}-\sum_{i \in \mathcal{I}_t}\sum_{z_i \in \mathcal{Z}}\frac{P_{Z_i}}{K}\ell(\hat{h}_t,z_i)\log \frac{1}{P_{Z_i}}\Big\vert $:
  \begin{equation}
      \begin{aligned}
      T_2    &\leq \frac{2b(M-K)}{M}H(Z^{\mathcal{I}_P}) +bH(Z^{\mathcal{I}_P})-\frac{b}{K}\sum_{i \in \mathcal{I}_t}H(Z_i).
      \end{aligned}
  \end{equation}
\end{comment}
  
 \begin{comment}
      To sum up, we can derive the final participation gap in this scenario as follows,
    \begin{equation}
      \begin{aligned}
   & \Big\vert\mathcal{L}_{Z^{\mathcal{I}}}(\hat{h}_t)-\mathcal{L}_{{Z_{\mathcal{I}_t}}}(\hat{h}_t)\Big\vert\\
   &\leq \frac{b(3N-2M)}{N}H(Z^{\mathcal{I}})+ \frac{2b(M-K)}{M}H(Z^{\mathcal{I}_p} )\\
   &\quad-\frac{b}{K}\sum_{i \in \mathcal{I}_t}H(Z_i)\\
    &=\frac{b(3N-2M)}{N}H(Z^{\mathcal{I}})+ \frac{b(2MK-2K^2-M)}{MK}H(Z^{\mathcal{I}_p} )\\
    &\quad+\frac{b}{K}\Big[H(Z^{\mathcal{I}_p} )-\sum_{i \in \mathcal{I}_t}H(Z_i)\Big]\\
        &=b(3-\frac{2M}{N})H(Z^{\mathcal{I}})+ b(2-\frac{2K}{M}-\frac{1}{K})H(Z^{\mathcal{I}_p} )\\
        &\quad+\frac{b}{K}\Big[H(Z^{\mathcal{I}_p} )-\sum_{i \in \mathcal{I}_t}H(Z_i)\Big].\\
      \end{aligned}
  \end{equation}
 \end{comment}

To sum up, we can derive the selection gap as follows,
    \begin{equation}
      \begin{aligned}
&\Big\vert\mathcal{L}_{Z^{\mathcal{I}_p}}(\hat{h}_t)-\mathcal{L}_{{Z_{\mathcal{I}_t}}}(\hat{h}_t)\Big\vert\\
&\leq \frac{b(3M-2K)}{M}H(Z^{\mathcal{I}_p})-bH(Z^{\mathcal{I}_t})\\
&\quad+bH(Z^{\mathcal{I}_t})-\frac{b}{K}\sum_{i \in \mathcal{I}_t} H(Z_i)\\
&=b(3-\frac{2K}{M}-\frac{1}{K})H(Z^{\mathcal{I}_p})+\frac{b}{K}[H(Z^{\mathcal{I}_p} )-\sum_{i \in \mathcal{I}_t}H(Z_i)].
      \end{aligned}
  \end{equation}
  
The term $\frac{b}{K}\Big[H(Z^{\mathcal{I}_p} )-\sum_{i \in \mathcal{I}_t}H(Z_i)\Big]$ can be further derived as follows,
    \begin{equation}
        \begin{aligned}
   \frac{b}{K}\Big[H(Z^{\mathcal{I}_p} )-\sum_{i \in \mathcal{I}_t}H(Z_i)\Big]  &\stackrel{\text{(a)}}{\leq} \frac{b}{K}\big[H(Z^{\mathcal{I}_p})-H(Z^{\mathcal{I}_t})\big]\\
   &\stackrel{\text{(b)}}{\leq} \frac{b}{K}\big[H(Z^{\mathcal{I}_p})-H(Z^{\mathcal{I}_t}| Z^{\mathcal{I}_p\setminus\mathcal{I}_t})\big]\\
     &\stackrel{\text{(c)}}{=}\frac{b}{K}\big[H(Z^{\mathcal{I}_p})-H(Z^{\mathcal{I}_p}| Z^{\mathcal{I}_p\setminus\mathcal{I}_t})\big]\\
             &=\frac{b}{K}I(Z^{\mathcal{I}_p};Z^{\mathcal{I}_p\setminus\mathcal{I}_t}),\\
        \end{aligned}
    \end{equation}
   where $(a)$ follows from the chain rule property of entropy and the $(b)$ follows from conditioning reduces entropy. $(c)$ makes use of the fact $H(X,Y|Y)=H(X|Y)$, where $X$ and $Y$ are two random variables.  Based on the above results, we can derive the result in Theorem~\ref{the:Fedcorre}.

Notice that the mutual information $I(Z^{\mathcal{I}_p};Z^{\mathcal{I}_p \setminus\mathcal{I}_t })$ can be rewritten by the form of KL-divergence, i.e., $I(Z^{\mathcal{I}_p};Z^{\mathcal{I}_p \setminus\mathcal{I}_t })=KL( P_{Z^{\mathcal{I}_p}}\Vert P_{Z^{\mathcal{I}_p}}P_{ Z^{\mathcal{I}_p \setminus \mathcal{I}_t}  })$, we thus have,

\begin{equation}
    \begin{aligned}
       \frac{b}{K}\Big[H(Z^{\mathcal{I}_p} )-\sum_{i \in \mathcal{I}_t}H(Z_i)\Big]  &\leq \frac{b}{K} KL( P_{Z^{\mathcal{I}_p}}\Vert P_{Z^{\mathcal{I}_p}}P_{ Z^{\mathcal{I}_p \setminus \mathcal{I}_t}  })  \\
       &=  \frac{b}{K}\sum_{ \mathcal{Z}^{ \mathcal{I}_p}}P_{Z^{\mathcal{I}_p}} \log \frac{P_{Z^{\mathcal{I}_p}}}{P_{Z^{\mathcal{I}_p}}P_{Z^{\mathcal{I}_p \setminus \mathcal{I}_t}}}\\
       &=  \frac{b}{K}\sum_{ \mathcal{Z}^{ \mathcal{I}_p \setminus  \mathcal{I}_t}}P_{Z^{\mathcal{I}_p \setminus \mathcal{I}_t}} \log \frac{1}{P_{Z^{\mathcal{I}_p \setminus \mathcal{I}_t}}}\\
       &=\frac{b}{K}H(Z^{\mathcal{I}_p\setminus\mathcal{I}_t})\\
                    & \stackrel{\text{(a)}}{\leq}\frac{b}{K}\sum_{i \in \mathcal{I}_p\setminus\mathcal{I}_t}H(Z_i)\\
                        &\stackrel{\text{(b)}}{\leq} \frac{b}{K}\sum_{i \in \mathcal{I}_p\setminus\mathcal{I}_t}H(P_{Z_i},P_{Z_j}),
    \end{aligned}
\end{equation}
where $(a)$ follows from the chain rule property of entropy and $(b)$ follows the fact that the cross entropy $H(p,q)$ satisfies $H(p,q)=H(p)+KL(p||q)>H(p)$, where $p$ and $q$ are two probability distributions.

  Rooted on the results obtained in Theorem~\ref{the:FedEntropy}, we can similarly complete the proof of Theorem~\ref{the:FedCross}.
\begin{comment}
     \begin{equation}
    \begin{aligned}
&\mathcal{L}_{Z^{\mathcal{I}}}(\hat{h}_t^*)- \mathcal{L}_{Z_{\mathcal{I}_t}} (\hat{h}_t^*) \\
&\leq L\Vert \hat{h}^*_t-h^*_t\Vert H(Z^{\mathcal{I}})+b(3-\frac{2M}{N})H(Z^{\mathcal{I}})\\
&\quad+ b(2-\frac{2K}{M}-\frac{1}{K})H(Z^{\mathcal{I}_p} )+\frac{b}{K}\sum_{i \in \mathcal{I}_p\setminus\mathcal{I}_t}H(P_{Z_i},P_{Z_j})+\mathcal{E}_t\\
&\quad+cb\sqrt{\frac{VC(\mathcal{G})}{\sum_{i=1}^{\vert\mathcal{I}_t \vert}n_i}}+b\sqrt{\frac{\log(1/\delta)}{2\sum_{i=1}^{\vert\mathcal{I}_t \vert}n_i}}.
    \end{aligned}
\end{equation}
where  $c$ is a constant. $h^*:=\sup_{h \in \mathcal{H}}L\Vert\hat{h}^*-h\Vert H(Z^{\mathcal{I}})$ and $L$ is the Lipschitz constant.  $\mathcal{E}_t =2b\sum_{i\in \mathcal{I}_t}  \sum_{z_i \in \mathcal{Z}}P_{Z_i}^2$. 
$H(P_{Z_i},P_{Z_j})$ is the cross entropy between distributions $P_{Z_i}$ and $P_{Z_j}$.
\end{comment}
\end{proof}

\bibliographystyle{ieeetr}
\bibliography{IEEEtran}

\begin{thebibliography}{10}

\bibitem{zhu2021federated}
H.~Zhu, J.~Xu, S.~Liu, and Y.~Jin, ``Federated learning on non-iid data: A survey,'' {\em Neurocomputing}, vol.~465, pp.~371--390, 2021.

\bibitem{DBLP:conf/aistats/McMahanMRHA17}
B.~McMahan, E.~Moore, D.~Ramage, S.~Hampson, and B.~A. y~Arcas, ``Communication-efficient learning of deep networks from decentralized data,'' in {\em Proceedings of the 20th International Conference on Artificial Intelligence and Statistics, {AISTATS} 2017, 20-22 April 2017, Fort Lauderdale, FL, {USA}} (A.~Singh and X.~J. Zhu, eds.), vol.~54 of {\em Proceedings of Machine Learning Research}, pp.~1273--1282, {PMLR}, 2017.

\bibitem{10106044}
Y.~Yan, X.~Tong, and S.~Wang, ``Clustered federated learning in heterogeneous environment,'' {\em IEEE Transactions on Neural Networks and Learning Systems}, pp.~1--14, 2023.

\bibitem{DBLP:journals/comcom/CoelhoNVSN23}
K.~K. Coelho, M.~Nogueira, A.~B. Vieira, E.~F. Silva, and J.~A.~M. Nacif, ``A survey on federated learning for security and privacy in healthcare applications,'' {\em Comput. Commun.}, vol.~207, pp.~113--127, 2023.

\bibitem{10.1145/3678181}
Y.~Zhang, D.~Zeng, J.~Luo, X.~Fu, G.~Chen, Z.~Xu, and I.~King, ``A survey of trustworthy federated learning: Issues, solutions, and challenges,'' {\em ACM Trans. Intell. Syst. Technol.}, jul 2024.

\bibitem{DBLP:conf/icdm/MaoWHYHY23}
Q.~Mao, S.~Wan, D.~Hu, J.~Yan, J.~Hu, and X.~Yang, ``Leveraging federated learning for unsecured loan risk assessment on decentralized finance lending platforms,'' in {\em {ICDM} (Workshops)}, pp.~663--670, {IEEE}, 2023.

\bibitem{10423793}
Z.~Sun, Y.~Xu, Y.~Liu, W.~He, L.~Kong, F.~Wu, Y.~Jiang, and L.~Cui, ``A survey on federated recommendation systems,'' {\em IEEE Transactions on Neural Networks and Learning Systems}, pp.~1--15, 2024.

\bibitem{DBLP:journals/tnsm/WuWL23}
Z.~Wu, X.~Wu, and Y.~Long, ``Joint scheduling and robust aggregation for federated localization over unreliable wireless {D2D} networks,'' {\em {IEEE} Trans. Netw. Serv. Manag.}, vol.~20, no.~3, pp.~3359--3379, 2023.

\bibitem{10304290}
Y.~Wang, Q.~Shi, and T.-H. Chang, ``Why batch normalization damage federated learning on non-iid data?,'' {\em IEEE Transactions on Neural Networks and Learning Systems}, pp.~1--15, 2023.

\bibitem{hu2023generalization}
X.~Hu, S.~Li, and Y.~Liu, ``Generalization bounds for federated learning: Fast rates, unparticipating clients and unbounded losses,'' in {\em International Conference on Learning Representations}, 2023.

\bibitem{zhao2018federated}
Y.~Zhao, M.~Li, L.~Lai, N.~Suda, D.~Civin, and V.~Chandra, ``Federated learning with non-iid data,'' {\em arXiv preprint arXiv:1806.00582}, 2018.

\bibitem{yuan2021we}
H.~Yuan, W.~Morningstar, L.~Ning, and K.~Singhal, ``What do we mean by generalization in federated learning?,'' {\em arXiv preprint arXiv:2110.14216}, 2021.

\bibitem{DBLP:conf/icml/MohriSS19}
M.~Mohri, G.~Sivek, and A.~T. Suresh, ``Agnostic federated learning,'' in {\em Proceedings of the 36th International Conference on Machine Learning, {ICML} 2019, 9-15 June 2019, Long Beach, California, {USA}} (K.~Chaudhuri and R.~Salakhutdinov, eds.), vol.~97 of {\em Proceedings of Machine Learning Research}, pp.~4615--4625, {PMLR}, 2019.

\bibitem{huang2024federated}
W.~Huang, M.~Ye, Z.~Shi, G.~Wan, H.~Li, B.~Du, and Q.~Yang, ``Federated learning for generalization, robustness, fairness: A survey and benchmark,'' {\em IEEE Transactions on Pattern Analysis and Machine Intelligence}, 2024.

\bibitem{DBLP:conf/icml/QuLDLTL22}
Z.~Qu, X.~Li, R.~Duan, Y.~Liu, B.~Tang, and Z.~Lu, ``Generalized federated learning via sharpness aware minimization,'' in {\em International Conference on Machine Learning, {ICML} 2022, 17-23 July 2022, Baltimore, Maryland, {USA}} (K.~Chaudhuri, S.~Jegelka, L.~Song, C.~Szepesv{\'{a}}ri, G.~Niu, and S.~Sabato, eds.), vol.~162 of {\em Proceedings of Machine Learning Research}, pp.~18250--18280, {PMLR}, 2022.

\bibitem{wei2022non}
B.~Wei, J.~Li, Y.~Liu, and W.~Wang, ``Non-iid federated learning with sharper risk bound,'' {\em IEEE Transactions on Neural Networks and Learning Systems}, 2022.

\bibitem{lim2020federated}
W.~Y.~B. Lim, N.~C. Luong, D.~T. Hoang, Y.~Jiao, Y.-C. Liang, Q.~Yang, D.~Niyato, and C.~Miao, ``Federated learning in mobile edge networks: A comprehensive survey,'' {\em IEEE Communications Surveys \& Tutorials}, vol.~22, no.~3, pp.~2031--2063, 2020.

\bibitem{DBLP:journals/corr/abs-2302-09019}
M.~Wang, Y.~Pan, X.~Yang, G.~Li, and Z.~Xu, ``Tensor networks meet neural networks: {A} survey,'' {\em CoRR}, vol.~abs/2302.09019, 2023.

\bibitem{DBLP:conf/icml/0005SLW0X22}
Y.~Pan, Z.~Su, A.~Liu, J.~Wang, N.~Li, and Z.~Xu, ``A unified weight initialization paradigm for tensorial convolutional neural networks,'' in {\em {ICML}}, vol.~162 of {\em Proceedings of Machine Learning Research}, pp.~17238--17257, {PMLR}, 2022.

\bibitem{tak2020federated}
A.~Tak and S.~Cherkaoui, ``Federated edge learning: Design issues and challenges,'' {\em IEEE Network}, vol.~35, no.~2, pp.~252--258, 2020.

\bibitem{10472111}
Z.~Wu, Z.~Xu, D.~Zeng, J.~Li, and J.~Liu, ``Topology learning for heterogeneous decentralized federated learning over unreliable d2d networks,'' {\em IEEE Transactions on Vehicular Technology}, pp.~1--6, 2024.

\bibitem{wang2021federated}
Y.~Wang, Z.~Su, T.~H. Luan, R.~Li, and K.~Zhang, ``Federated learning with fair incentives and robust aggregation for uav-aided crowdsensing,'' {\em IEEE Transactions on Network Science and Engineering}, vol.~9, no.~5, pp.~3179--3196, 2021.

\bibitem{10295990}
W.~Huang, M.~Ye, Z.~Shi, and B.~Du, ``Generalizable heterogeneous federated cross-correlation and instance similarity learning,'' {\em IEEE Transactions on Pattern Analysis and Machine Intelligence}, vol.~46, no.~2, pp.~712--728, 2024.

\bibitem{DBLP:conf/nips/ReisizadehFPJ20}
A.~Reisizadeh, F.~Farnia, R.~Pedarsani, and A.~Jadbabaie, ``Robust federated learning: The case of affine distribution shifts,'' in {\em Advances in Neural Information Processing Systems 33: Annual Conference on Neural Information Processing Systems 2020, NeurIPS 2020, December 6-12, 2020, virtual} (H.~Larochelle, M.~Ranzato, R.~Hadsell, M.~Balcan, and H.~Lin, eds.), 2020.

\bibitem{ma2022state}
X.~Ma, J.~Zhu, Z.~Lin, S.~Chen, and Y.~Qin, ``A state-of-the-art survey on solving non-iid data in federated learning,'' {\em Future Generation Computer Systems}, vol.~135, pp.~244--258, 2022.

\bibitem{caldarola2022improving}
D.~Caldarola, B.~Caputo, and M.~Ciccone, ``Improving generalization in federated learning by seeking flat minima,'' in {\em Computer Vision--ECCV 2022: 17th European Conference, Tel Aviv, Israel, October 23--27, 2022, Proceedings, Part XXIII}, pp.~654--672, Springer, 2022.

\bibitem{DBLP:conf/spawc/YagliDP20}
S.~Yagli, A.~Dytso, and H.~V. Poor, ``Information-theoretic bounds on the generalization error and privacy leakage in federated learning,'' in {\em 21st {IEEE} International Workshop on Signal Processing Advances in Wireless Communications, {SPAWC} 2020, Atlanta, GA, USA, May 26-29, 2020}, pp.~1--5, {IEEE}, 2020.

\bibitem{DBLP:conf/isit/BarnesDP22}
L.~P. Barnes, A.~Dytso, and H.~V. Poor, ``Improved information theoretic generalization bounds for distributed and federated learning,'' in {\em {IEEE} International Symposium on Information Theory, {ISIT} 2022, Espoo, Finland, June 26 - July 1, 2022}, pp.~1465--1470, {IEEE}, 2022.

\bibitem{sefidgaran2022ratedistortion}
M.~Sefidgaran, R.~Chor, and A.~Zaidi, ``Rate-distortion theoretic bounds on generalization error for distributed learning,'' in {\em Advances in Neural Information Processing Systems} (A.~H. Oh, A.~Agarwal, D.~Belgrave, and K.~Cho, eds.), 2022.

\bibitem{10374183}
Z.~Wu, Z.~Xu, H.~Yu, and J.~Liu, ``Information-theoretic generalization analysis for topology-aware heterogeneous federated edge learning over noisy channels,'' {\em IEEE Signal Processing Letters}, pp.~1--5, 2023.

\bibitem{DBLP:conf/cvpr/ZhangXYZ0W23}
R.~Zhang, Q.~Xu, J.~Yao, Y.~Zhang, Q.~Tian, and Y.~Wang, ``Federated domain generalization with generalization adjustment,'' in {\em {CVPR}}, pp.~3954--3963, {IEEE}, 2023.

\bibitem{DBLP:conf/nips/NguyenTL22}
A.~T. Nguyen, P.~H.~S. Torr, and S.~N. Lim, ``Fedsr: {A} simple and effective domain generalization method for federated learning,'' in {\em NeurIPS}, 2022.

\bibitem{nguyen2022fedsr}
A.~T. Nguyen, P.~Torr, and S.-N. Lim, ``Fedsr: A simple and effective domain generalization method for federated learning,'' in {\em Advances in Neural Information Processing Systems}, 2022.

\bibitem{DBLP:conf/aaai/LiSWZRLK022}
B.~Li, Y.~Shen, Y.~Wang, W.~Zhu, C.~Reed, D.~Li, K.~Keutzer, and H.~Zhao, ``Invariant information bottleneck for domain generalization,'' in {\em Thirty-Sixth {AAAI} Conference on Artificial Intelligence, {AAAI} 2022, Thirty-Fourth Conference on Innovative Applications of Artificial Intelligence, {IAAI} 2022, The Twelveth Symposium on Educational Advances in Artificial Intelligence, {EAAI} 2022 Virtual Event, February 22 - March 1, 2022}, pp.~7399--7407, {AAAI} Press, 2022.

\bibitem{muandet2013domain}
K.~Muandet, D.~Balduzzi, and B.~Sch{\"o}lkopf, ``Domain generalization via invariant feature representation,'' in {\em International conference on machine learning}, pp.~10--18, PMLR, 2013.

\bibitem{DBLP:conf/nips/ParkSE22}
S.~Park, U.~Simsekli, and M.~A. Erdogdu, ``Generalization bounds for stochastic gradient descent via localized {\textdollar}{\textbackslash}varepsilon{\textdollar}-covers,'' in {\em NeurIPS}, 2022.

\bibitem{li2020federated}
T.~Li, A.~K. Sahu, M.~Zaheer, M.~Sanjabi, A.~Talwalkar, and V.~Smith, ``Federated optimization in heterogeneous networks,'' {\em Proceedings of Machine learning and systems}, vol.~2, pp.~429--450, 2020.

\bibitem{10269141}
R.~Dai, X.~Yang, Y.~Sun, L.~Shen, X.~Tian, M.~Wang, and Y.~Zhang, ``Fedgamma: Federated learning with global sharpness-aware minimization,'' {\em IEEE Transactions on Neural Networks and Learning Systems}, pp.~1--14, 2023.

\bibitem{10197230}
Y.~Chen, W.~Lu, X.~Qin, J.~Wang, and X.~Xie, ``Metafed: Federated learning among federations with cyclic knowledge distillation for personalized healthcare,'' {\em IEEE Transactions on Neural Networks and Learning Systems}, pp.~1--12, 2023.

\bibitem{9927349}
B.~Wei, J.~Li, Y.~Liu, and W.~Wang, ``Non-iid federated learning with sharper risk bound,'' {\em IEEE Transactions on Neural Networks and Learning Systems}, vol.~35, no.~5, pp.~6906--6917, 2024.

\bibitem{paninski2003estimation}
L.~Paninski, ``Estimation of entropy and mutual information,'' {\em Neural computation}, vol.~15, no.~6, pp.~1191--1253, 2003.

\bibitem{10506990}
Z.~Luan, W.~Li, M.~Liu, and B.~Chen, ``Robust federated learning: Maximum correntropy aggregation against byzantine attacks,'' {\em IEEE Transactions on Neural Networks and Learning Systems}, pp.~1--14, 2024.

\bibitem{10337775}
K.~Tam, L.~Li, B.~Han, C.~Xu, and H.~Fu, ``Federated noisy client learning,'' {\em IEEE Transactions on Neural Networks and Learning Systems}, pp.~1--14, 2023.

\bibitem{9843892}
Y.~Zou, Z.~Wang, X.~Chen, H.~Zhou, and Y.~Zhou, ``Knowledge-guided learning for transceiver design in over-the-air federated learning,'' {\em IEEE Transactions on Wireless Communications}, vol.~22, no.~1, pp.~270--285, 2023.

\bibitem{DBLP:journals/corr/abs-2303-00897}
D.~Zeng, X.~Hu, S.~Liu, Y.~Yu, Q.~Wang, and Z.~Xu, ``Stochastic clustered federated learning,'' {\em CoRR}, vol.~abs/2303.00897, 2023.

\bibitem{boyd2004convex}
S.~P. Boyd and L.~Vandenberghe, {\em Convex optimization}.
\newblock Cambridge university press, 2004.

\bibitem{barber1996quickhull}
C.~B. Barber, D.~P. Dobkin, and H.~Huhdanpaa, ``The quickhull algorithm for convex hulls,'' {\em ACM Transactions on Mathematical Software (TOMS)}, vol.~22, no.~4, pp.~469--483, 1996.

\bibitem{9669076}
M.~Xiong, B.~Zhang, D.~W.~C. Ho, D.~Yuan, and S.~Xu, ``Event-triggered distributed stochastic mirror descent for convex optimization,'' {\em IEEE Transactions on Neural Networks and Learning Systems}, vol.~34, no.~9, pp.~6480--6491, 2023.

\bibitem{he2023asymptotic}
X.~He, X.~Yi, Y.~Zhao, K.~H. Johansson, and V.~Gupta, ``Asymptotic analysis of federated learning under event-triggered communication,'' {\em IEEE Transactions on Signal Processing}, vol.~71, pp.~2654--2667, 2023.

\bibitem{DBLP:journals/corr/abs-1812-01097}
S.~Caldas, P.~Wu, T.~Li, J.~Kone{\v{c}}n{\'y}, H.~B. McMahan, V.~Smith, and A.~Talwalkar, ``{LEAF:} {A} benchmark for federated settings,'' {\em CoRR}, vol.~abs/1812.01097, 2018.

\bibitem{DBLP:journals/corr/abs-2010-01243}
Y.~J. Cho, J.~Wang, and G.~Joshi, ``Client selection in federated learning: Convergence analysis and power-of-choice selection strategies,'' {\em CoRR}, vol.~abs/2010.01243, 2020.

\bibitem{song2022learning}
H.~Song, M.~Kim, D.~Park, Y.~Shin, and J.-G. Lee, ``Learning from noisy labels with deep neural networks: A survey,'' {\em IEEE transactions on neural networks and learning systems}, vol.~34, no.~11, pp.~8135--8153, 2022.

\bibitem{DBLP:journals/corr/abs-2306-11650}
S.~Liang, J.~Huang, D.~Zeng, J.~Hong, J.~Zhou, and Z.~Xu, ``Fednoisy: Federated noisy label learning benchmark,'' {\em CoRR}, vol.~abs/2306.11650, 2023.

\bibitem{yin2018byzantine}
D.~Yin, Y.~Chen, R.~Kannan, and P.~Bartlett, ``Byzantine-robust distributed learning: Towards optimal statistical rates,'' in {\em International conference on machine learning}, pp.~5650--5659, Pmlr, 2018.

\bibitem{9054377}
J.~Peng and Q.~Ling, ``Byzantine-robust decentralized stochastic optimization,'' in {\em ICASSP 2020 - 2020 IEEE International Conference on Acoustics, Speech and Signal Processing (ICASSP)}, pp.~5935--5939, 2020.

\bibitem{krizhevsky2009learning}
A.~Krizhevsky, G.~Hinton, {\em et~al.}, ``Learning multiple layers of features from tiny images,'' 2009.

\bibitem{he2016deep}
K.~He, X.~Zhang, S.~Ren, and J.~Sun, ``Deep residual learning for image recognition,'' in {\em Proceedings of the IEEE conference on computer vision and pattern recognition}, pp.~770--778, 2016.

\bibitem{10197242}
H.~Huang, W.~Shi, Y.~Feng, C.~Niu, G.~Cheng, J.~Huang, and Z.~Liu, ``Active client selection for clustered federated learning,'' {\em IEEE Transactions on Neural Networks and Learning Systems}, pp.~1--15, 2023.

\bibitem{he2021towards}
Y.~He, Z.~Shen, and P.~Cui, ``Towards non-iid image classification: A dataset and baselines,'' {\em Pattern Recognition}, vol.~110, p.~107383, 2021.

\end{thebibliography}

\begin{IEEEbiography}[{\includegraphics[width=1in,height=1.25in,clip,keepaspectratio]{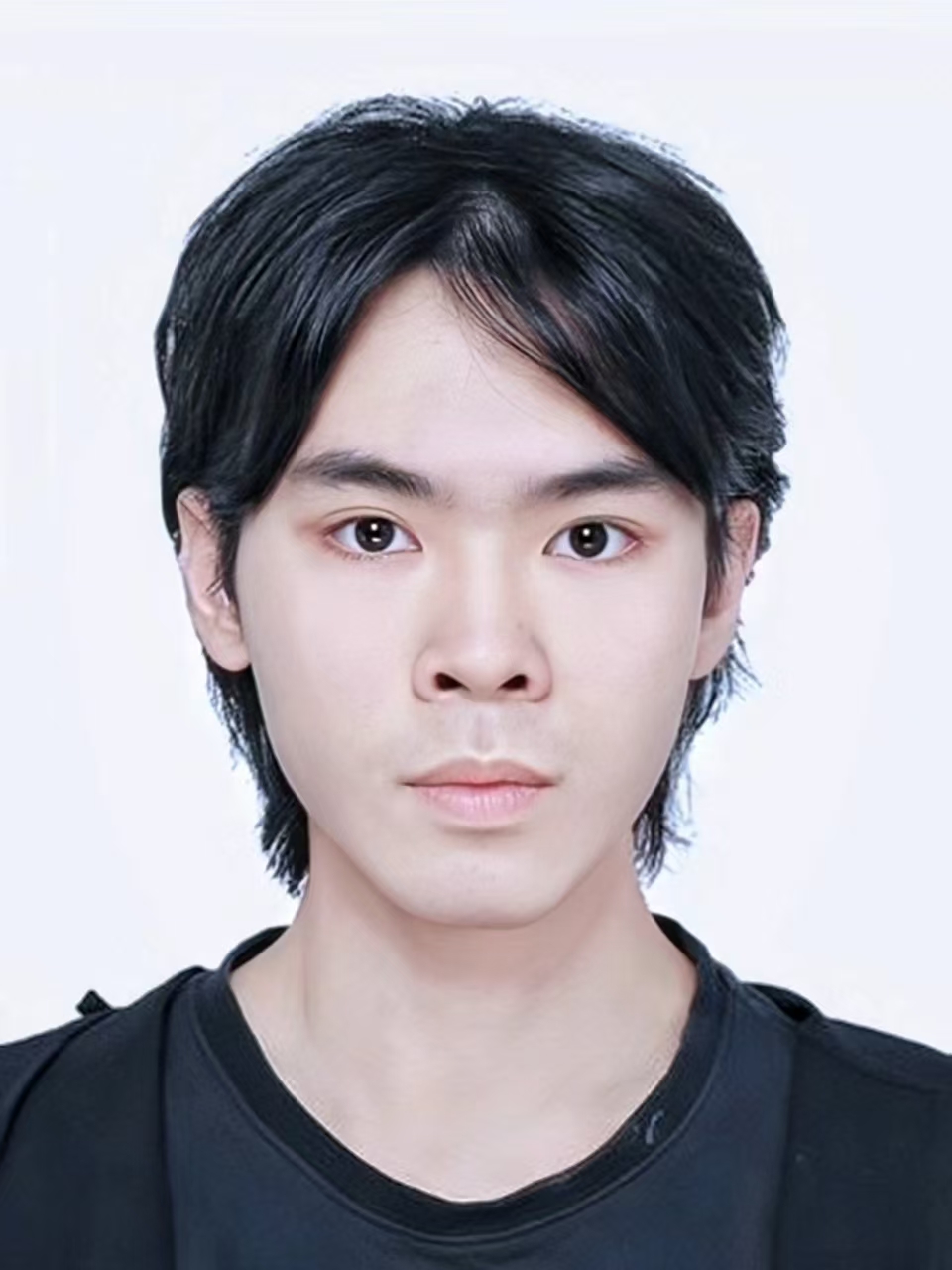}}]{\small Zheshun Wu} \small is a Ph.D. student at the School of Computer Science
and Technology, Harbin Institute of Technology, Shenzhen, China. He received his
Bachelor’s degree from South China University of Technology, China, and a Master’s
degree from Sun Yat-sen University, China.
His research interests include federated learning and wireless communications.
\end{IEEEbiography}

\vspace{-39pt}
\begin{IEEEbiography}[{\includegraphics[width=1in,height=1.25in,clip,keepaspectratio]{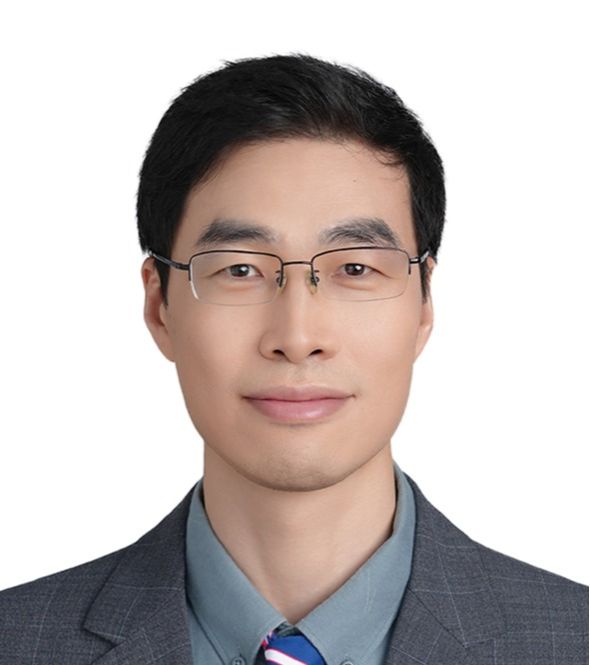}}]{\small Zenglin Xu} \small (Senior Member, IEEE) is currently a full professor in Fudan University.
He received the Ph.D. degree in computer science and engineering from the The Chinese University of Hong Kong. 
He has been working at Michigan State University, Cluster of Excellence at Saarland University and Max Planck Institute for Informatics, Purdue University, and University of Electronic Science \& Technology of China. 
His research interests include machine learning and its applications in computer vision, natural language processing, and health informatics. 
%He currently serves as an associate editor Neural Networks and Neurocomputing. He is the recipient of the outstanding student paper honorable mention of AAAI 2015, the best student paper runner up of ACML 2016, and the 2016 young researcher award from APNNS.
\end{IEEEbiography}

\vspace{-39pt}
\begin{IEEEbiography}[{\includegraphics[width=1in,height=1.25in,clip,keepaspectratio]{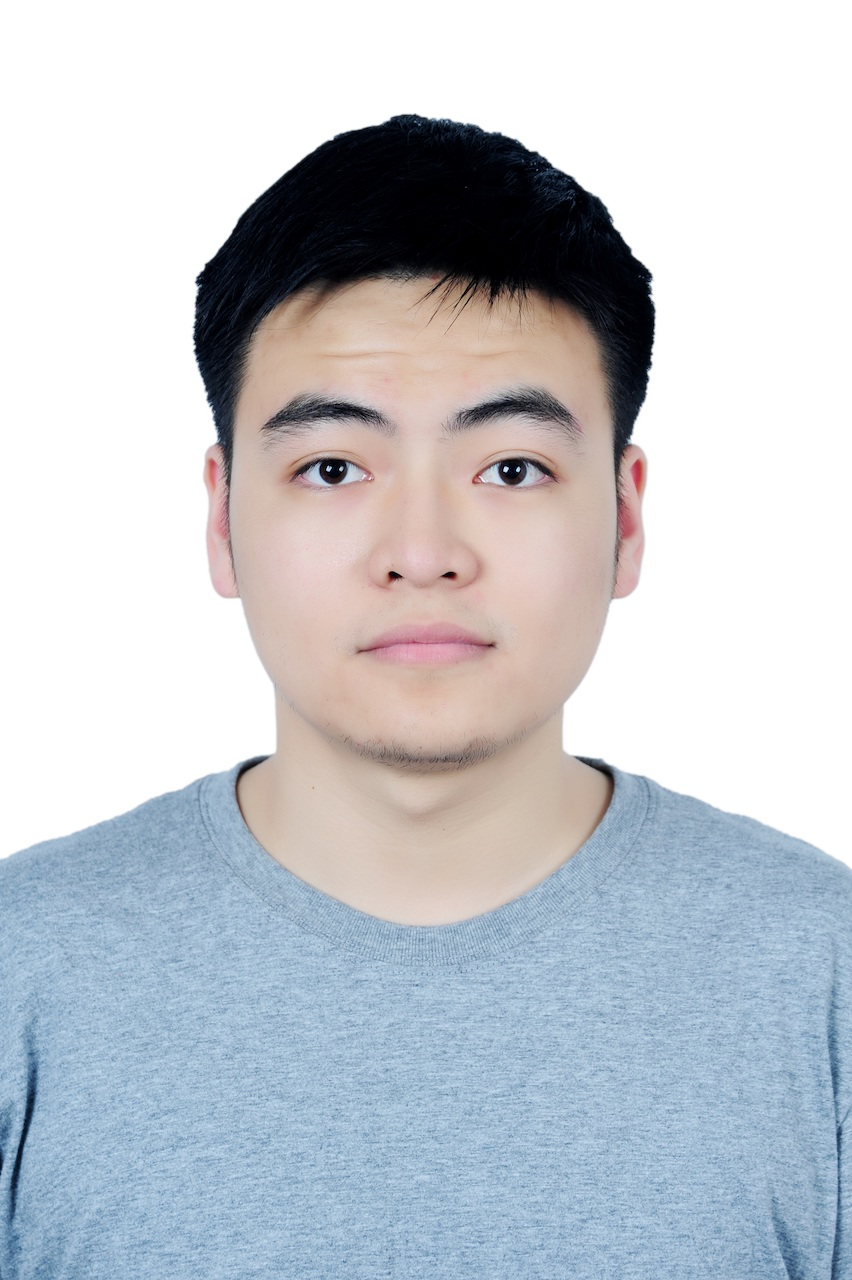}}]{\small Dun Zeng} \small is a Ph.D. candidate at the School of Computer Science and Engineering, University of Electronic Science and Technology of China, focusing on federated learning and distributed optimization.
\end{IEEEbiography}

\vspace{-39pt}

\begin{IEEEbiography}[{\includegraphics[width=1in,height=1.25in,clip,keepaspectratio]{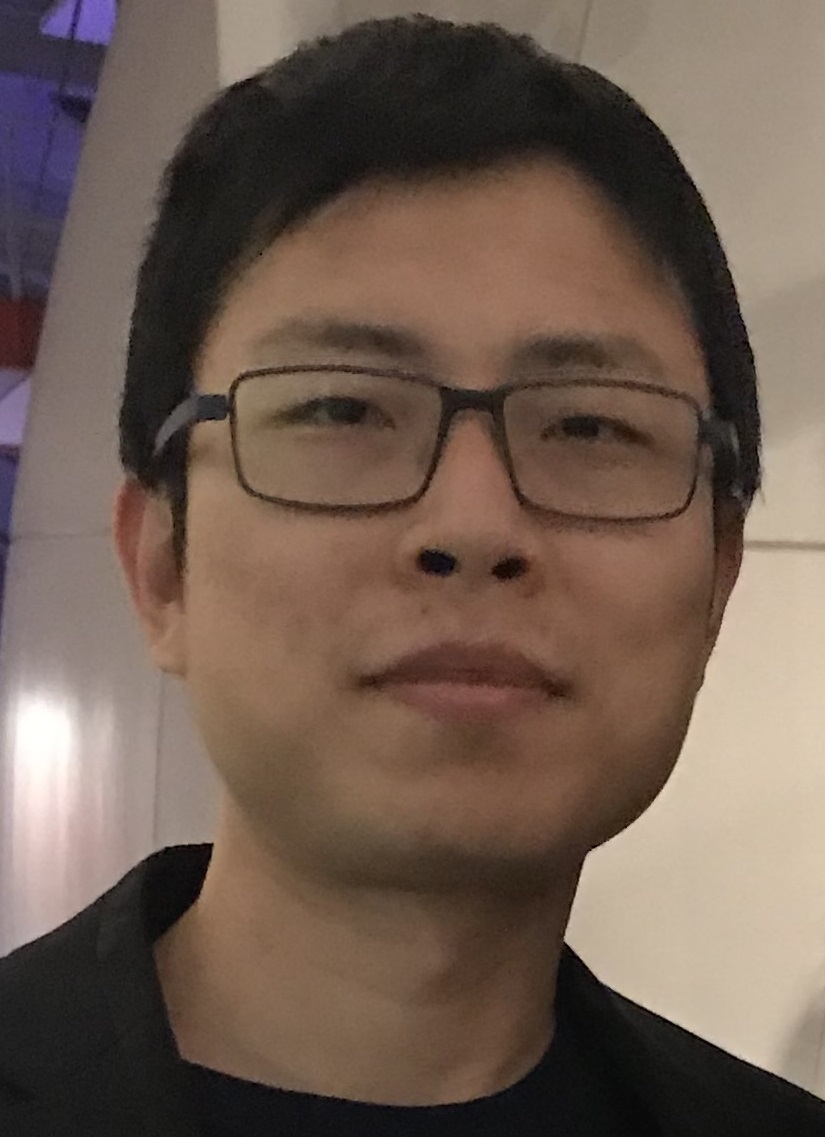}}]{\small Qifan Wang} \small received the BS and MS degrees from
Tsinghua University, and the PhD degree from Purdue
University. He is the research scientist with Meta AI, leading a team building innovative Deep Learning
and Natural Language Processing models for Recommendation System. Before joining Meta, He worked with Google Research, and Intel Labs before joining Meta. His research interests include deep learning, natural language processing, information retrieval, data mining, and computer vision. He has co-authored more than 100 publications in top-tier conferences and journals, including NeurIPS, SIGKDD, WWW, SIGIR, AAAI, IJCAI, ACL,
EMNLP, WSDM, CIKM, ECCV, TPAMI, TKDE, and TOIS. He also serve as area chairs, program committee members, editorial board members, and reviewers for academic conferences and journals.
\end{IEEEbiography}

\vspace{-39pt}
\begin{IEEEbiography}[{\includegraphics[width=1in,height=1.25in,clip,keepaspectratio]{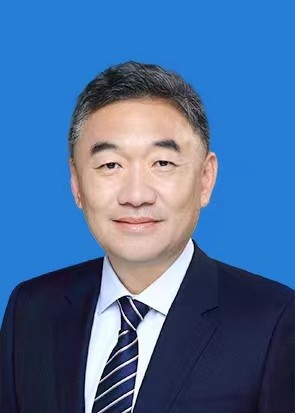}}]
{\small Jie Liu}
\small (Fellow, IEEE)  is a Chair Professor at Harbin Institute of Technology Shenzhen (HIT Shenzhen), China and the Dean of its AI Research Institute. 
Before joining HIT, he spent 18 years at Xerox PARC and Microsoft. 
He was a Principal Research Manager at Microsoft Research, Redmond and a partner of the company. 
His research interests are Cyber-Physical Systems, AI for IoT, and energy-efficient computing. 
He received IEEE TCCPS Distinguished Leadership Award and 7 Best Paper Awards from top conferences. 
He is an IEEE Fellow and an ACM Distinguished Scientist, and founding Chair of ACM SIGBED China.
\end{IEEEbiography}

\end{document}